\documentclass[10pt]{article} 
\usepackage[accepted]{tmlr}

\usepackage{amsmath,amsfonts,bm}
\usepackage{amsthm}

\def\eqref#1{equation~\ref{#1}}

\def\1{\bm{1}}

\newtheorem{assumption}{Assumption}[section]
\newtheorem{theorem}{Theorem}[section] 
\newcommand{\EE}{\mathbb{E}}
\newcommand{\T}{\mathcal{T}}

\newcommand{\D}{\mathcal{D}}

\DeclareMathAlphabet{\mathsfit}{\encodingdefault}{\sfdefault}{m}{sl}
\SetMathAlphabet{\mathsfit}{bold}{\encodingdefault}{\sfdefault}{bx}{n}

\usepackage{placeins}   
\usepackage{afterpage}  

\usepackage[utf8]{inputenc} 
\usepackage{hyperref}       
\usepackage{url}            
\usepackage{booktabs}       
\usepackage{amsfonts}       
\usepackage{nicefrac}       
\usepackage{microtype}      
\usepackage{lipsum}
\usepackage{graphicx}
\graphicspath{ {./images/} }

\usepackage{subcaption}

\usepackage{algorithm}
\usepackage{algorithmic}
\usepackage{xcolor}
\usepackage{pifont} 

\definecolor{lightpurple}{rgb}{0.725,0.45,0.725} 
\definecolor{darkpurple}{rgb}{0.325,0.0,0.325}   

\usepackage{newfloat}
\usepackage{listings}
\usepackage{dsfont}
\providecommand{\mathbbm}[1]{\mathds{#1}}

\usepackage{amssymb}
\usepackage{amsmath}
\usepackage{amsthm}
\usepackage{refcount}
\usepackage{bm}
\usepackage{tabularx}
\usepackage{enumitem}
\usepackage{wrapfig}
\usepackage{thmtools}
\usepackage{thm-restate}

\usepackage{booktabs,tabularx,array,makecell,adjustbox}
\newcolumntype{L}[1]{>{\raggedright\arraybackslash}p{#1}}

\newtheorem{definition}[theorem]{Definition}
\newtheorem{corollary}[theorem]{Corollary}

\newtheorem{lemma}[theorem]{Lemma}

\usepackage{scalerel}
\usepackage{mathtools}
\usepackage{calc}

\usepackage{caption}
\definecolor{forestgreen}{RGB}{34,139,34}

\newcommand{\commentsh}[1]{}
\newcommand{\commentl}[1]{}
\newcommand{\emma}[1]{}
\newcommand{\e}[1]{}
\newcommand{\eb}[1]{}
\newcommand{\commentj}[1]{}
\newcommand{\commentjmiao}[1]{}

\title{Repairing Reward Functions with Feedback to\\Mitigate Reward Hacking}

\author{\name Stephane Hatgis-Kessell \email stephhk@stanford.edu \\
      \addr Computer Science Department\\
      Stanford University
      \AND
      \name Logan Mondal Bhamidipaty \email \\
      \addr School of Informatics\\
      The University of Edinburgh
      \AND
      \name Emma Brunskill \email \\
      \addr Computer Science Department\\
      Stanford University}

\def\month{11}  
\def\year{2026} 
\def\openreview{\url{https://openreview.net/forum?id=GmwH8FBthd}} 

\begin{document}
\raggedbottom
\maketitle
\makeatletter
\let\AND\@undefined
\makeatother

\begin{abstract}
Human-designed reward functions for reinforcement learning (RL) agents are frequently misaligned with the humans' true, unobservable objectives, and thus act only as proxies. Optimizing for a misspecified proxy reward function often induces reward hacking, resulting in a policy misaligned with the human's true objectives. An alternative is to perform RL from human feedback, which involves learning a reward function from scratch by collecting human preferences over pairs of trajectories. However, building such datasets is costly. To address the limitations of both approaches,  we propose Preference-Based Reward Repair (PBRR): an automated iterative framework that repairs a human-specified proxy reward function by learning an additive, transition-dependent correction term from preferences. A manually specified reward function can yield policies that are highly suboptimal under the ground-truth objective, yet corrections on only a few transitions may suffice to recover optimal performance. To identify and correct for those transitions, PBRR uses a targeted exploration strategy and a new preference-learning objective. We prove that in tabular domains, when the ground-truth return is linear in known trajectory features, a variant of PBRR with explicit optimistic exploration has a cumulative regret that matches, up to constants, that of prior preference-based RL methods. In addition, on a suite of reward-hacking benchmarks, PBRR consistently outperforms baselines that learn a reward function from scratch from preferences or modify the proxy reward function using other approaches, requiring substantially fewer preferences to learn high performing policies. 

\end{abstract}

\section{Introduction}
\label{sec:intro}

\begin{wrapfigure}{R}{0.61\textwidth} 
\vspace{-6mm}
  \centering
  \includegraphics[width=0.6\textwidth]{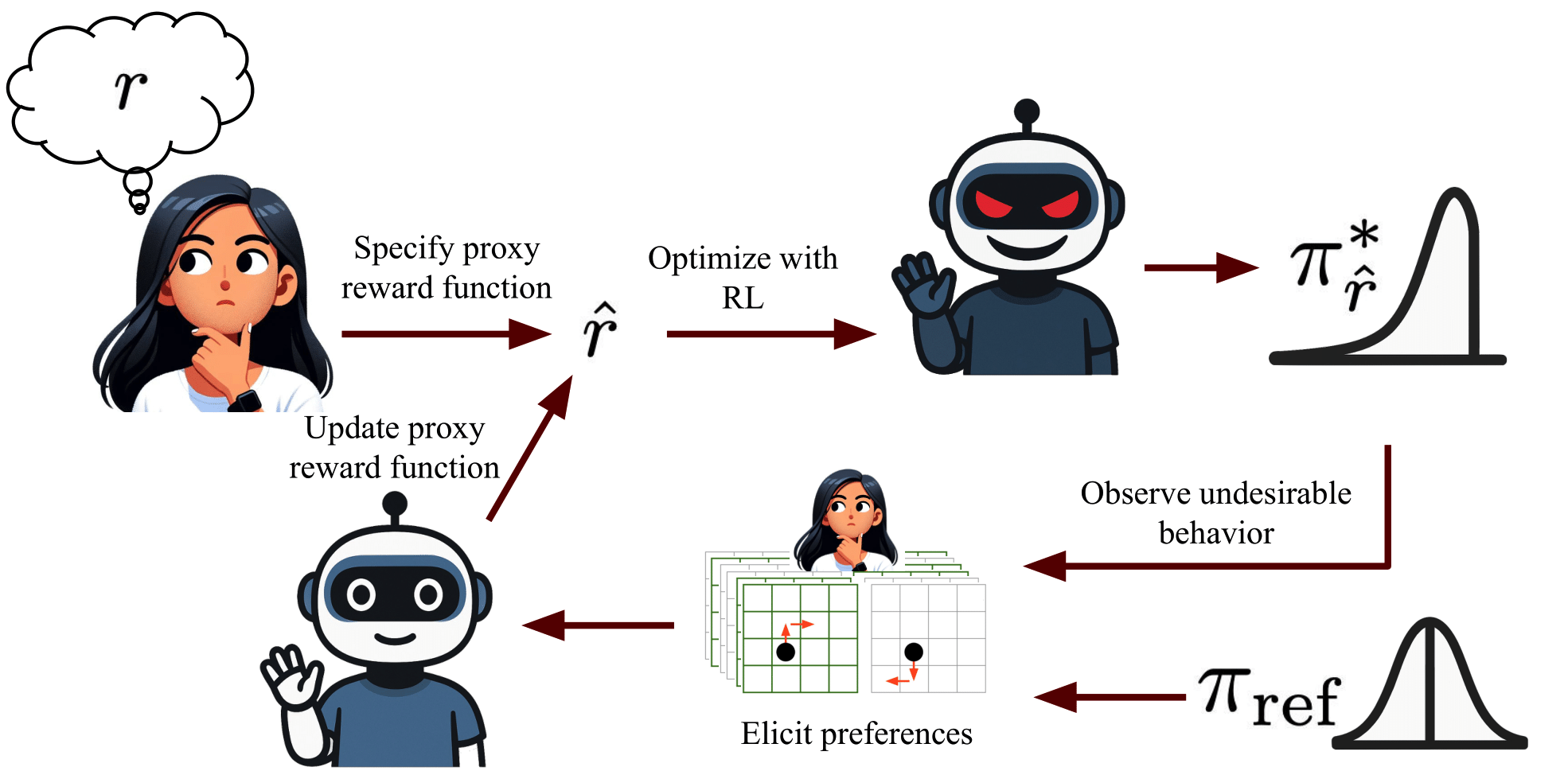} 
  \caption{Illustration of Preference-Based Reward Repair (PBRR). A human specifies a proxy reward function $\hat{r}$, which is optimized for with reinforcement learning (RL) to produce a policy $\pi^*_{\hat{r}}$. Preferences between trajectories from $\pi^*_{\hat{r}}$ and a safe reference policy $\pi_{\text{ref}}$ are elicited to identify instances of unaligned behavior. The preferences are used to update $\hat{r}$ and the process repeats, iteratively aligning the proxy reward function with the human’s unobservable ground-truth reward function $r$.}
  \label{fig:pbrr_overview}
\vspace{-5mm}
\end{wrapfigure}


The reward hypothesis states that ``all of what we mean by goals and purposes can be well thought of as maximization of the expected value of the cumulative sum of reward'' \citep{sutton2018reinforcement}. This idea underpins much of reinforcement learning (RL): if we can specify the right reward function, then optimizing for it should yield the desired behavior. However, manually designing a reward function that fully captures a human designer’s true objectives is rarely possible \citep{amodei2016concrete}.

One approach is to instead rely on \textit{proxy} reward functions---simpler specifications that reflect the intended but unobservable ground-truth objective. Unfortunately  even well-considered proxies, when optimized for by RL, often fail to produce policies that achieve the desired behavior; a failure mode informally known as \emph{reward hacking} \citep{krakovna2020specification, pan2022effects}. For example, in an autonomous driving task, maximizing mean velocity---a proxy for traffic flow---could lead vehicles to block highway on-ramps. 
The common recourse is an iterative, trial-and-error design process. A designer specifies a proxy reward function, trains an agent, observes the resulting behavior, and then manually edits the reward function to remove unwanted incentives \citep{booth2023perils, knox2023reward}. We conjecture that while this process can eventually produce usable reward functions and aligned policies, it is slow, ad hoc, and depends on RL expertise that many domain experts do not have. Automating this process would make RL more practical, for example in domains such as pandemic lockdown policy design \citep{kompella2020reinforcement},  autonomous driving \citep{wu2021flow, dosovitskiy2017carla}, clinical decision making \citep{dallaman2014uvapadova,petersen2019deep, eastman2021reinforcement}, energy management \citep{henry2021gym, orfanoudakis2024ev2gym}, or tax policy optimization \citep{mi2023taxai}.



Another path to alignment is to remove the need for any explicit  human reward design through  learning a reward function from human preferences over trajectories, i.e., reinforcement learning from human feedback (RLHF). However, standard RLHF approaches typically require large datasets of human preferences, which are often prohibitively costly to collect \citep{casper2023open}. Some work~\citep{novoseller2020dueling,pacchiano_dueling} has proposed methods for RLHF with cumulative regret guarantees, using strategic exploration under uncertainty to require fewer preferences, but rely on restrictive assumptions such as discrete state-action spaces and specific human preference generation processes. Scaling up uncertainty-based approaches for RLHF to more complex domains is non-trivial and empirical success has been mixed~\citep{ji2024reinforcement,das2024provably,dwaracherla2024efficient,mehta2023sample}.



\eb{The next part of this paragraph is a key point and it's getting buried. I try making a separate paragraph where you clearly summarize the challenge, building on your sentences at the end of this paragraph, and then a paragraph with your solution. E.g. human specified rewards are often brittle and manual "observation" of resulting policies followed by updating the reward is a common but informal approach, and RLHF approaches are automatable but can be very data intensive (and typically focus on a two step process -- gather data then optimize -- rather than iterative. and then I would propose the solution as a new paragraph} \commentsh{Added a draft sentence above to alleviate this.}

Human specified reward functions are often misaligned, requiring an informal and manual reward correction process, while RLHF approaches are potentially data intensive. To address these limitations,
we introduce \textbf{Preference-Based Reward Repair (PBRR)}, an iterative framework for efficiently and automatically repairing a human-specified proxy reward function using preferences. For many tasks, a human can readily specify a proxy reward function that reflects the unobservable ground-truth objective they have in mind, but lacks robustness to the ways in which an RL agent might exploit it. However, a limited number of targeted adjustments to the proxy reward function may be enough to restore near-optimal behavior. PBRR makes these adjustments as an automated process, in contrast to the  manual iterative reward correction process that humans often perform. 
To automate reward function repair, PBRR leverages two core components: (i) a targeted exploration strategy, which elicits preferences between trajectories generated by the policy trained with the proxy reward function and those from a supplied reference policy, and (ii) a new preference-learning objective to update the proxy reward function only over transitions incorrectly assigned high reward. Figure \ref{fig:pbrr_overview} provides an overview.

On sequential decision process benchmark environments that highlight the challenges of reward hacking~\citep{pan2022effects}, 
PBRR substantially outperforms approaches that learn a reward function from scratch from preferences--i.e., standard RLHF---or attempt to repair the proxy reward function using alternative strategies.  
While our contributions are primarily empirical, we also provide supporting theory: we prove that in the tabular, linear dueling bandit setting of past theoretical work, where the ground-truth return is linear in known trajectory features, a variant of PBRR matches the cumulative regret bounds of a prior strategic RLHF method~\citep{pacchiano_dueling} up to constant terms.
Our contributions are three-fold:
\begin{itemize}[leftmargin=*]
    \item We introduce Preference-Based Reward Repair (PBRR), a method for efficiently repairing a human-specified proxy reward function using a new exploration method and learning objective. 
    \item We prove a variant of PBRR matches, up to constants, the sublinear cumulative regret bounds of ~\cite{pacchiano_dueling} in the same linear dueling bandit regime.
    \item We show that PBRR effectively repairs a proxy reward function even when that initial proxy reward induces a substantially suboptimal policy, and consistently matches or outperforms all baselines on a suite of reward-hacking benchmarks. 
\end{itemize}


\section{Background and Setting}
\label{sec:background}
\eb{The background/setting section is long. I'd try to get it to 1/2 a page. There doesn't need to be a subsection, we don't need to write out the D/mu in that format, etc. I would check anything important and non-standard is not buried. For example, you could put at the very end that we elicit preferences over full traj.}

Consider an MDP $\mathcal{M} \triangleq (S, A, \Omega, \gamma, p_0, r)$ with state space $S$, action space $A$, transition dynamics $\Omega: S \times A \rightarrow \Delta(S)$, discount factor $\gamma \in [0,1]$, and initial state distribution $p_0$. The horizon is $H$ and the ground-truth reward function $r: S \times A \times S \rightarrow \mathbb{R}$. $\mathcal{M} \setminus r \triangleq (S, A, \Omega, \gamma, p_0, \textunderscore)$ is an environment without a specified reward function, where the placeholder $\textunderscore$ denotes the unspecified reward function. Let $r$ be the (unobservable) ground-truth reward function, $\hat r$ a candidate approximation (e.g., learned or specified proxy), and $\tilde r$ an arbitrary reward function.

A policy $\pi: S \times A \rightarrow [0,1]$ maps states to action distributions. Its expected discounted return under $\tilde r$ from start distribution $p_0$ is $J_{\tilde r}(\pi)$. An \emph{optimal policy} for $\tilde r$ is any $\pi^*_{\tilde r} \in \arg\max_\pi J_{\tilde r}(\pi)$. We say $\hat r$ is misspecified in environment $\mathcal{M}$ with ground-truth  reward $r$ if $J_r(\pi^*_{\hat r}) < J_r(\pi^*_r)$. Unless noted, policy performance refers to expected discounted return under $r$. Let $\tau$ denote a trajectory starting at state $s^{\tau}_{ 0} \sim p_0$: 
$\tau = (s^{\tau}_{0},a^{\tau}_{0}, s^{\tau}_{1},a^{\tau}_{1}, ..., s^{\tau}_{{H}})$. Let   a trajectory's return be $\tilde{r}(\tau) = \sum_{t=0}^H \gamma^t \tilde{r}(s_t^\tau, a_t^\tau, s_{t+1}^\tau)$ and  $\mathcal{T}_{\pi}$ denote the support of the trajectory distribution induced by $\pi$ from $p_0$. 

We learn a reward $\hat r$ from trajectory pair preferences. Let \[
\D=\{(\tau_1,\tau_2,\mu)\}_{k=1}^N, \quad
\mu \in \{0,1,\tfrac12\} \text{ with }
0:\tau_1\succ\tau_2,\;
1:\tau_2\succ\tau_1,\;
\tfrac12:\tau_1\sim\tau_2.
\]
As is standard in RLHF \citep{christiano2017deep}, 
we assume a Bradley–Terry preference model:
\[
P(\tau_1\succ \tau_2 \mid \tilde r)=\sigma\!\big(\tilde r(\tau_1)-\tilde r(\tau_2)\big),\quad
\sigma(x)=1/(1+e^{-x}) 
\]
and, unless otherwise stated, fit $\hat r$ by minimizing the cross-entropy loss:

\begin{equation}
\label{eq:loss}
\begin{gathered}
\mathcal{L}_{\text{pref}}(\hat{r};\D_t) =
- \hspace{-7mm}\sum_{(\tau_1, \tau_2, \mu) \in \D} \hspace{-5mm} (1-\mu) \log {P}(\tau_1 \succ \tau_2 | \hat{r}) + \mu \log {P}(\tau_1 \prec \tau_2 | \hat{r}).
\end{gathered}
\end{equation}
We assume a preference $\mu$ is elicited over a pair of trajectories, rather than shorter trajectory segments, to mitigate issues relating to misspecified human preference models (see Appendix.~\ref{app:traj_pref_justification}). 
\section{Related Work}
Alignment has received extensive attention, particularly in the context of large language models. Here we focus instead on sequential decision processes.

Under more restrictive assumptions-- knowledge of the human’s MDP \citep{mechergui2024expectation}, availability of corrective actions \citep{jiang2024reinforcement} or feature attribution-based explanations \citep{mahmud2023explanation}, or that the specified reward function induces near-optimal performance \citep{hadfield2017inverse} subject to only additional calibration \citep{fu2025reward}-- prior work has explored how to align agent behavior in MDPs despite a human's misspecified reward function. Our approach doesn't require these assumptions, nor are they satisfied in the evaluation environments we consider. Other work infers a posterior over plausible reward functions from human data to mitigate errors in the reward functions \citep{eisenstein2023helping,mahmud2023explanation, coste2023reward}, and disjoint work shows how such a prior (either learned or provided directly by a stakeholder) can be leveraged for efficient exploration~\citep{novoseller2020dueling}. However it can be challenging for stakeholders to provide Bayesian priors, and learning them may be brittle. We instead only require a stakeholder to provide a single proxy reward function. \eb{I think this is a natural comparison and makes sense to include and argue why it's not likely to be realistic.}




In RLHF for large language models, given access to a sufficiently performant reference policy and an estimated but flawed reward function, penalizing KL-divergence between action distributions during training can induce a high-performing policy (see, e.g., \citet{ziegler2019fine, ouyang2022training, bai2022training, glaese2022improving, openai2022chatgpt, touvron2023llama}). In MDPs, \citet{laidlaw2024correlated} highlight that using different divergence measures can improve policy performance.
In contrast, our work focuses on settings where no such high-performing reference policy is available.

\commentsh{There is a lot of work on data-efficient RLHF, especially outside LLM settings. IMO we shouldn't dedicate so much space to this line of related work unless we move this section to the appendix.}
Another line of work focuses on efficiently learning a reward function from preferences. Theoretical results for RL in discrete state and action spaces are promising~\citep{novoseller2020dueling,pacchiano_dueling} but rely on quantifying priors or precise measures of uncertainty over the reward function, which is unclear how to replicate in more complex settings. In large language model settings, preference-based algorithms leveraging coarse approximations of uncertainty or optimism have yielded modest empirical benefit~\citep{mehta2023sample, xie2024exploratory, das2024provably}.

Concurrent to our work, \citet{cao2025residual} also assume access to a proxy reward function and learn an additive correction term from preferences, demonstrating benefits on robotic manipulation tasks. Our work differs in several important ways. We study settings where a misspecified proxy reward function induces highly suboptimal behavior. We then propose an exploration strategy that effectively corrects the proxy reward function by leveraging a suboptimal reference policy, as in other RLHF methods. Further, we introduce a new learning objective for repairing a proxy reward function. Together, these components substantially improve performance compared to applying the standard RLHF procedure to update an inputted proxy reward function, which corresponds directly to the baseline of \citet{cao2025residual}, in the reward-hacking benchmark introduced by \cite{pan2022effects}. We also provide a theoretical analysis of our approach, whereas \citet{cao2025residual} focus solely on empirical results.

\commentl{@Stephane, the related lit talks a lot about reward hacking solutions, but there's also a literature that has adapted CE loss functions to be good at learning rankings. I'm not extremely familiar w/ the lit, but I think papers like RankNet or surveys on cross-entropy regularization for ranking might be worth mentioning. The major difference is that the application domain is entirely different.}
\commentsh{@Logan those papers definitely sound interesting I will check them out! IMO our related works section should only focus on methods to fix reward hacking---and I haven't seen the works you mention discussed by other RLHF papers where they would also be relevant. Feel free to push back on this though.}

\section{Methodology}
\label{sec:method}
We now present our Preference-Based Reward Repair (PBRR) algorithm. We assume a human stakeholder initially provides a reward function $\hat{r}_{\text{proxy}}(s,a,s')$. PBRR then iteratively  aligns this human-specified proxy reward function to their ground-truth objective by eliciting preferences. 


Without loss of generality, the ground-truth reward function $r(s,a,s')$ can be written as the proxy reward function $\hat r_{\text{proxy}}(s,a,s')$ plus a correction $g(s,a,s')$. Thus, repairing $\hat r_{\text{proxy}}$ amounts to learning a transition-dependent correction $g$. At iteration $t$, we elicit a preference batch, add it to the dataset $\mathcal{D}_t$ of all preferences collected so far (i.e., $\mathcal{D}_t \supseteq \mathcal{D}_{t-1}$, with $\mathcal{D}_0 = \emptyset$), and fit $g_t$ on $\mathcal{D}_t$, yielding the modified proxy reward function used at the next iteration:
\begin{equation}
\label{eq:r_decompisition}
   \hat{r}_{t+1}(s,a,s') \triangleq \hat{r}_{\text{proxy}}(s,a,s') + g_t(s,a,s')
\end{equation}
$\hat{r}_{t}$ always denotes a modified proxy reward function, while $\hat{r}_\text{proxy}$ denotes the original proxy reward function; with $g_0 \equiv 0$, $\hat{r}_1 = \hat{r}_\text{proxy}$. $g_t(s,a,s')$ is parameterized as a neural network. As shorthand, we write $g(\tau) =\sum_{t=0}^{H} \gamma^t g(s_t, a_t, s_{t+1})$

This specification offers three benefits. First, the stakeholder need only provide a point-estimate reward function---which will then be corrected---rather than a full Bayesian prior over reward functions and uncertainties for Bayesian methods. Second, data-efficiency may increase when the complexity of the additive correction term lies in a lower dimensional space than the full reward function. Third, as we will now show, it enables the design of a loss function that explicitly leverages expected properties of the proxy reward function. 
\eb{Stephane, please take a pass on this paragraph-- the high level is it would be helpful for us to lay out explicitly the nice benefits of this decomp.}\commentsh{I made some small changes but really like what you wrote!}


 Many cases of reward hacking arise because humans misestimate the cumulative effect of small or multi-objective rewards which can dominate long-term outcomes (e.g., an agent in a racing task learns to loop endlessly through checkpoints to accumulate points, not to finish the race), or because humans mispredict which actions will maximize expected return (e.g., an RL agent exploits a bug in its environment). These types of over-optimistic reward functions cover most examples of reward-hacking from \citet{krakovna2020specification}, and exactly constitute the harmful errors in proxy reward functions identified by \citet{shang2026errors}. 
 \commentsh{@Emma I removed your reference to classic hard exploration problems (commented below) to save on word count but lmk if you disagree.}A standard way to learn $g_t(s,a,s')$ would be to minimize cross-entropy (Eq.~\ref{eq:loss}). Instead, we introduce an alternate loss function which can perform particularly strongly when the proxy reward function is overly optimistic\footnote{$\hat{r}$ is overly optimistic if, $\forall (s,a,s') \text{, } \hat{r}(s,a,s') \geq r(s,a,s')$}, without requiring this optimism to hold.

\textbf{Repairing a proxy reward function~~~}Based on this intuition, we design a loss function for learning the correction term $g$ that regularizes towards only correcting transitions that are incorrectly assigned high reward, conflicting with observed preferences. We first partition the preference dataset into $\D^+_t$, the set that contains all samples 
where the \emph{original} proxy reward function's induced ranking 
matches the elicited preference $\mu$, and $\D^-_t$, the set that contains all other samples $(\tau_1, \tau_2, \mu)$:

$\D^+_t = \left\{ (\tau_1, \tau_2, \mu) \,\middle|\, \operatorname{sign}\left(\hat{r}_\text{proxy}(\tau_2) - \hat{r}_\text{proxy}(\tau_1)\right) = \operatorname{sign}(\mu - 0.5) \right\}$ and $\D^-_t = \D_t \setminus \D^+_t$. The partition is defined by the fixed proxy $\hat{r}_\text{proxy}$, not by the modified reward $\hat{r}_\text{proxy} + g$, so a pair's membership does not change as $g$ is learned. A tied pair ($\mu = \frac{1}{2}$) lies in $\D^+_t$ only if $\hat{r}_\text{proxy}$ also assigns both trajectories equal return, and otherwise lies in $\D^-_t$, where both indicators in $\mathcal{L}^-$ below are zero, so it contributes only through $\mathcal{L}_\text{pref}$. In our experiments, preference labels are sampled from a Bradley--Terry model and are therefore binary, so ties do not arise.

We then learn the corrective term $g$ by minimizing the following three-term loss:
\begingroup
\setlength{\abovedisplayskip}{4pt}   
\setlength{\belowdisplayskip}{4pt}   
\begin{align}
\label{eq:reg-pref-loss}
\mathcal{L}(g; \hat{r}_\text{proxy}, \D_t) \triangleq \;
& \mathcal{L}_\text{pref}(\hat{r}_\text{proxy} + g; \D_t) + \lambda_1 \underbrace{\tfrac{1}{|\D_t^+|} \sum_{(\tau_1,\tau_2)\in \D_t^+}\!\big[g(\tau_1)^2 + g(\tau_2)^2\big]}_{\mathcal{L}^+}  \nonumber \\
& + \lambda_2 \underbrace{\tfrac{1}{|\D_t^-|} \sum_{(\tau_1,\tau_2)\in \D_t^-}\!\big[\mathbbm{1}\{\tau_1 \succ \tau_2\}g(\tau_1)^2 + \mathbbm{1}\{\tau_2 \succ \tau_1\}g(\tau_2)^2\big]}_{\mathcal{L}^-}
\end{align}
\endgroup
\e{Can we instead put all the full terms in this equation and then use underbraces to define them as L+, L- etc? I think that'd be clearer.}\commentsh{done.}
The first term, 
\( \mathcal{L}_{\text{pref}} \), is a standard preference loss that encourages the modified reward function \( \hat{r}_\text{proxy} + g\) to satisfy the preferences in \( \D_t \). The second term $\mathcal{L}^+$
regularizes the correction term \( g \) towards zero on trajectory pairs where the proxy reward function agrees with the human preference. This assumes that when the proxy reward function correctly ranks a pair of trajectories, its assigned reward for the transitions in those trajectories is consistent with the ground-truth reward function; $\mathcal{L}^+$ prevents unnecessary adjustments that could degrade an otherwise correct reward signal. Finally, the third term $\mathcal{L}^-$
focuses on trajectory pairs that were misclassified by the original proxy reward function $\hat{r}_\text{proxy}$. Adjusting the correction term \( g\) to correctly classify such trajectories is generally underspecified-- one could (a) increase the reward for the preferred trajectory, or (b) decrease the reward for the not preferred trajectory. $\mathcal{L}^-$ prioritizes option (b), regularizing the correction term \( g \) to zero on transitions in the preferred trajectories, which will prioritize a negative correction for undesirable behaviors.  

To ensure our approach still learns the ground-truth reward function when our assumptions about the specified proxy reward function fail to hold, we decay $\lambda_1$ and $\lambda_2$ over iterations (see Appendix~\ref{app:lambdas_decay_desc}). Although Eq. \ref{eq:reg-pref-loss} leverages the assumption that the proxy reward function is optimistic, our algorithm does not require this assumption, nor does our theoretical analysis in Section \ref{sec:regret_analysis} or Section \ref{sec:reward_hacking_analysis}.

\label{sec:expl-method}

\textbf{Constructing a preference dataset~~~}We next describe how data are gathered to repair the proxy reward function. We assume access to a reference policy,  e.g., constructed from heuristics or a previously used policy. Our hypothesis is that such a policy could provide a valuable contrast to when the proxy reward function's induced policy still leads to behavior considered undesirable by the stakeholder.

\begin{algorithm}[H]
\caption{Preference-Based Reward Repair (PBRR)}
\label{alg:exploration}
\begin{algorithmic}[1]
\STATE {\bf Input:} Proxy reward function $\hat{r}_\text{proxy}$, reference policy $\pi_{\text{ref}}$, number of iterations $T$
\STATE Initialize $g_0(s, a, s') \gets 0$ for all $(s, a, s')$; $\D_0 \gets \emptyset$
\FOR{$t = 1$ to $T$}
    \STATE Compute $\pi^*_{\hat{r}_t}$ given the proxy reward function $\hat{r}_{t} = \hat{r}_\text{proxy} + g_{t-1}$ \footnotemark
   \STATE $\pi_1 = \pi^*_{\hat{r}_t}$,  $\pi_2 = \pi_{\mathrm{ref}}$
    \IF{$C_1 > 0$}
    \STATE Compute $\Pi_t$, non-dominated policy set
    \IF{$\pi^*_{\hat{r}_t} \notin \Pi_t$ or $\pi_{\mathrm{ref}} \notin \Pi_t$ or  
    $C_1 f(\pi^*_{\hat{r}_t},\pi_{\mathrm{ref}}) \leq  \max_{\pi_1,\pi_2 \in \Pi_t} f(\pi_1,\pi_2)$}
    \STATE $\pi_1,\pi_2 =  \arg\max_{\pi_1,\pi_2 \in \Pi_t} f(\pi_1,\pi_2)$
    \ENDIF 
    \ENDIF
    \STATE Collect trajectories $\T_{\pi_1}$ and $\T_{\pi_2}$ and sample trajectory pairs $(\tau_1, \tau_2)$ with $\tau_1 \in \T_{\pi_1}, \tau _2\in \T_{\pi_2}$
    \STATE Elicit preferences $\mu$ over each pair $(\tau_1, \tau_2)$ and set $\D_t \gets \D_{t-1} \cup \{(\tau_1, \tau_2, \mu)\}$
    \STATE Learn additive correction $g_{t}$ using Equation~\ref{eq:reg-pref-loss} with $\D_t$, yielding $\hat{r}_{t+1} = \hat{r}_\text{proxy} + g_t$
\ENDFOR
\STATE {\bf Output:} Final modified reward function $\hat{r}_{T+1}$
\end{algorithmic}
\end{algorithm}

Accordingly, PBRR prioritizes eliciting preferences between trajectories sampled from the policy that optimizes for the corrected proxy reward function and the reference policy, matching the exploration strategy of \citet{xie2024exploratory}. However, in some settings, this exploration strategy may be insufficient to correct the proxy reward function to induce an optimal policy. Therefore, we follow ~\cite{pacchiano_dueling} and define an undominated policy set: the set of policies that remain potentially optimal given the observed data. From this set, we identify the pair of policies with the largest divergence in expected feature values under a weighted covariance norm. If the reference policy and policy optimizing for the corrected proxy reward function have a divergence within a constant of this maximal value, we use the reference policy and corrected proxy reward function's induced policy for exploration. If not, the algorithm can use the maximum divergence policy pair. This strategy enables a principled fallback for additional optimistic exploration when needed, and is analogous to prior work in contextual bandits that defaults to explicit optimistic exploration only when necessary \citep{bastani2021mostly}. Our PBRR algorithm is detailed in Algorithm \ref{alg:exploration}, where we define $C_1 \in \{0\} \cup [1, \infty)$ as the tolerance for using $(\pi^*_{\hat{r}_t}, \pi_{\text{ref}})$ instead of the maximum divergence policy pair, and $f$ and $\Pi_t$ are defined in Section \ref{sec:regret_analysis} for when $C_1>0$.

\footnotetext{In practice, we train a policy using $\hat{r}_t$ but are not guaranteed to find an optimal policy $\pi^*_{\hat{r}_t}$.}

\commentsh{@Emma I edited this section for conciseness but your original version is commented right beneath. Let me know if we should revert back.}
\subsection{Regret Analysis}
\label{sec:regret_analysis}
We now show that if the ground-truth return for a trajectory can be expressed as a linear function of the trajectory embedding, $r(\tau) = \langle \phi(\tau), w^* \rangle$, then PBRR achieves $\sqrt{T}$ cumulative regret. We draw upon recent theoretical results from \citep{pacchiano_dueling}. 
Our key observation is that if the reference policy $\pi_{\text{ref}}$ and optimal policy for the repaired reward function $\pi_{\hat{r}_t}$ lie in the set of possibly optimal policies under the ground-truth reward function given the observed data, and the features induced by their policies maximize the uncertainty with respect to paired feature covariance matrix up to a constant of the maximizing policy pair, then sampling trajectories from the support of $\pi_{\text{ref}}$ and $\pi_{\hat{r}_t}$ will yield bounded cumulative regret within a constant factor of selecting the maximizing uncertainty pair. If these conditions do not hold, our algorithm will instead reduce\footnote{It is possible to define a variant of our algorithm to handle the various subcases of if $\pi_{\text{ref}}$ is not in the non-dominated policy class $\Pi_t$, if $\pi_{\hat{r}_t}$ is not in the non-dominated policy class $\Pi_t$, or to consider if either either $\pi^*_{\hat{r}_t}$ or  $\pi_{\text{ref}}$ can be used as one of the uncertainty-maximizing policy pair; these cases do not impact our theoretical results and so for simplicity we keep the algorithm as is.} to selecting the maximizing uncertainty pair. Proofs and additional details are provided in Appendix~\ref{sec:regret_bound_proofs}.
\commentsh{@Emma do we not want to explicitly state the optimistic reward assumption?}
\begin{assumption}
\label{asm:linear}
The trajectory return is linear in a feature trajectory embedding, $r(\tau) = \langle \phi(\tau), w \rangle.$
\end{assumption}
\begin{assumption}
\label{asm:bt_prefs}
Preferences over trajectories are sampled from the Bradley-Terry preference model defined over the difference in return between trajectories.
\end{assumption}
\begin{assumption} We assume that 
$\|\mathbf{w}^*\| \leq W$ for some known $W > 0$.
\label{asm:bounded}
\end{assumption}
\begin{assumption} For all trajectories $\tau$ we assume that $\|\phi(\tau)\| \leq B$ for some known $B > 0$.
\label{asm:bounded_features}
\end{assumption}

In Theorem \ref{thm:known_trans_model_regret_bound} we show that when the dynamics model is known, PBRR inherits the same cumulative regret bounds as ~\cite{pacchiano_dueling} up to constants. We provide full details in Appendix \ref{sec:regret_bound_proofs}. 

We first define the undominated set of policies as the set of policies that remain potentially optimal under the current uncertainty in the linear reward model parameters: \\$\Pi_t \triangleq \left\{ \pi_i | (\phi(\pi_i) - \phi(\pi))^T w_t + \gamma_t(\delta) || \phi(\pi^i) - \phi(\pi) ||_{V^{-1}_t} \geq 0 \; \forall \; \pi \right\}$\\
We also define the regularized feature covariance matrix with respect to the difference in trajectory feature values: $V_t \triangleq  \sum_{l=1}^{t-1} (\phi(\tau_l^1) - \phi(\tau_l^2))(\phi(\tau_l^1) - \phi(\tau_l^2))^T + \kappa \lambda I_d$. Next, we consider the difference in expected feature embeddings of two policies under the known dynamics model, measured in the inverse covariance norm: $f_{mk}(\pi_1,\pi_2) = || \phi(\pi_1) - \phi(\pi_2) ||_{V^{-1}_t}$.
And finally we define a measure of the non-linearity of the sigmoid function over the parameter
space: $\kappa \triangleq \sup_{x \in B_B(d),\, w \in B_W(d)} \frac{1}{\sigma'(w^\top x)}$, where
$\sigma'$ denotes the derivative and $B_B(d)$, $B_W(d)$ denote the $\ell_2$-norm balls of radius
$B$ and $W$ in dimension $d$. We then have:

\begin{theorem}
\label{thm:known_trans_model_regret_bound}
Let $\delta \leq \frac{1}{e}$ and $\lambda \geq B/\kappa$. Then under Assumptions \ref{asm:linear}, \ref{asm:bt_prefs}, \ref{asm:bounded}, and \ref{asm:bounded_features}, and that the dynamics model is known, for $f=f_{mk}$ and $\Pi_t = \Pi_{t,mk}$, with probability at least $1-\delta$, the expected regret of Algorithm \ref{alg:exploration} is bounded by 
\begin{equation}
\mathrm{Regret}_T \le \tilde O\!\left( C_1 \big(\kappa\sqrt{\lambda}\,W + \kappa\sqrt{d} + BW\sqrt{d}\big)\sqrt{Td} \right)
\end{equation}
where 
$\tilde{\mathcal{O}}$ hides logarithmic factors in $T,\frac{1}{\delta},B,\frac{1}{\kappa},\frac{1}{\lambda},\frac{1}{d}$.
\end{theorem} 

We explicitly retain the $C_1$ factor to show that our regret guarantees are slightly looser than the results of \cite{pacchiano_dueling}, owing to our alternative trajectory selection strategy.

In Theorem \ref{thm:unknown_trans_model_regret_bound} we show that when the dynamics model is unknown, PBRR also inherits the same cumulative regret bounds as ~\cite{pacchiano_dueling} up to constants. We defer further details to Appendix \ref{sec:regret_bound_proofs}, which redefines the undominated policy set to account for uncertainty in the unknown dynamics model $\hat{\mathbb{P}}_t$, and defines $f_u$ to compute the expected trajectory feature difference using $\hat{\mathbb{P}}_t$.

\begin{theorem}
\label{thm:unknown_trans_model_regret_bound}
 Under Assumptions \ref{asm:linear}, \ref{asm:bt_prefs}, \ref{asm:bounded}, and \ref{asm:bounded_features}, for $f=f_{u}$ and $\Pi_t = \Pi_{t,u}$, the regret of Algorithm \ref{alg:exploration} is bounded by 
\begin{equation}
R_T \leq \tilde{\mathcal{O}}\!\left( C_1 \left( 
\kappa d \sqrt{T} + H^{3/2} \sqrt{|\mathcal{S}||\mathcal{A}| d T H} 
+ H |\mathcal{S}| \sqrt{|\mathcal{A}| d T H} 
\right) \right),
\end{equation}
for all $T$ simultaneously with probability at least $1-15\delta$, where 
$\tilde{\mathcal{O}}$ hides logarithmic factors in $\delta,|S|$ and $|A|$. 
\end{theorem}
\vspace{-2mm}

Theorems \ref{thm:known_trans_model_regret_bound} and \ref{thm:unknown_trans_model_regret_bound} suggest that it may be possible to select the reference policy and policy that optimizes for the current proxy reward function while matching---up to constants---the regret bounds of prior work.

\subsection{Reward Hacking Analysis}
\label{sec:reward_hacking_analysis}
Explicit exploration ($C_1 \neq 0$) sometimes is empirically dominated by setting $C_1=0$. 
We now additionally show that when $C_1=0$ and preferences are labeled by regret, PBRR induces a policy that performs no worse than the reference policy once the induced policy stops changing (Thm.~\ref{thm:pbrr-performance}) and resolves two specific instantiations of reward hacking (Cors.~\ref{corr:no-skalse-hacking}, \ref{corr:no-laidlaw-hacking}).

We begin with the following assumptions, which differ from those in Section \ref{sec:regret_analysis}:

\begin{assumption}
\label{asm1:vanish}
The regularization weights vanish $\lambda_1, \lambda_2 \to 0$ as $t \to \infty$ in Algorithm~\ref{alg:exploration}.
\end{assumption}

\begin{assumption}
\label{asm1:regret_prefs}
    Preferences over trajectories are noiseless and determined by the difference in regret between trajectories, measured against an optimal policy, i.e., the discounted sum of negative optimal advantages (Eq. \ref{eq:det-regret-pref}; see Appendix~\ref{app:regret_pref_theory_just} for how this differs from regret measured against the behavior policy) and $\hat r$ is fit with the corresponding regret preference model.\footnote{This assumption is strong in our experiments, where preferences are elicited over full trajectories. It is unclear in this setting if humans would use regret to label preferences.  However, prior work suggests that when preferences are elicited over partial trajectory segments, regret aligns more closely with human judgments \citep{knox2022models}. Therefore whether this assumption is plausible depends on the specific setting in which preferences are compared. See Appendix~\ref{app:regret_pref_theory_just} for further discussion.  }
\end{assumption}

\begin{assumption}
\label{asm1:c_is_0}
    $C_1=0$ in Algorithm \ref{alg:exploration}.
\end{assumption}

\begin{assumption}
\label{asm1:infinite_data}
 Each trajectory set is finite and contains the support of the corresponding policy, i.e., $\mathrm{support}(\pi) \subseteq \T_\pi$ and Line~13 adds every pair
$(\tau_1,\tau_2) \in \T_{\pi_1} \times \T_{\pi_2}$ to $\D_t$.
\end{assumption}

\begin{assumption}
\label{asm1:same_start}
All trajectories begin in the same start state $s_0$. This assumption is without loss of generality, as outlined in Appendix \ref{app:regret_pref_theory_just}.
\end{assumption}

\begin{assumption}
\label{asm1:consistent}
The repaired reward function $\hat r$ is consistent with the preference dataset: for every
$(\tau_1,\tau_2,\mu)\in\D$,
$\mathrm{sign}\left[P(\tau_1 \succ \tau_2 \mid \hat r) - \tfrac12\right]
= \mathrm{sign}\left(\tfrac12 - \mu\right)$
\end{assumption}


Assumptions~\ref{asm1:vanish} and ~\ref{asm1:c_is_0} match the version of PBRR we implement for our empirical results. Under these assumptions, at any iteration at which the policy induced by the repaired reward
function does not change, PBRR is guaranteed to have repaired the proxy reward function such
that it induces a policy at least as performant as the reference policy:

\begin{theorem}
Suppose Assumptions~\ref{asm1:vanish} through \ref{asm1:consistent} hold. Then at any iteration
$t$ of Algorithm~\ref{alg:exploration} at which $\pi^*_{\hat r_t} = \pi^*_{\hat r_{t-1}}$, the optimal policy
for the repaired reward function matches or outperforms the reference policy:
\[
J_r(\pi^*_{\hat r_t}) \ge J_r(\pi_{\mathrm{ref}}).
\]
In particular, this holds for all $t > t_0$ if $\pi^*_{\hat r_t}$ is constant for $t \ge t_0$.
\label{thm:pbrr-performance}
\end{theorem}

Theorem~\ref{thm:pbrr-performance} implies that reward hacking is resolved once the induced policy stops changing, following specific instantiations of the two different definitions of reward hacking from \citet{Skalse} and \citet{laidlaw2024correlated}:

\begin{corollary}[Unhackability in the sense of \citet{Skalse}]
At any iteration $t$ as in Theorem~\ref{thm:pbrr-performance}, the reward functions $\hat r_t$ and $r$
are unhackable relative to the policy set $\{\pi^*_{\hat r_t}, \pi_{\mathrm{ref}}\}$.
\label{corr:no-skalse-hacking}
\end{corollary}


\begin{corollary}[Unhackability in the sense of \citet{laidlaw2024correlated}]
At any iteration $t$ as in Theorem~\ref{thm:pbrr-performance}, $\hat r_t$ is unhackable with respect
to $r$ and $\pi_{\mathrm{ref}}$.
\label{corr:no-laidlaw-hacking}
\end{corollary}

Proofs and additional details are in Appendix \ref{app:theory}.

\section{Experiments}
\label{sec:results}
We now empirically evaluate PBRR. Our settings largely involve high-dimensional state spaces where the ground-truth reward function is not linear.\footnote{At least, not in a feature space that is likely to be known to the decision maker in advance.} Defining undominated policy sets for complex, non-linear reward functions learned from preferences is intractable without further restrictions on the policy and reward class, as is finding the best uncertainty-maximizing policy pairs.  Therefore, in our empirical results we set $C_1=0$, which implies, from Line 6 of Algorithm \ref{alg:exploration}, that for exploration PBRR always uses the reference policy and the policy that optimizes for the corrected proxy reward function. Consequently, PBRR as evaluated empirically is not covered by the regret bounds in Section \ref{sec:regret_analysis}, but is covered by the theoretical analysis in Section \ref{sec:reward_hacking_analysis}. We find that repairing a proxy reward function with PBRR and $C_1=0$ induces substantially better performance with fewer preferences than either learning a reward function from scratch via RLHF or repairing a proxy reward function using alternative methods. 
\eb{I don't think we need these refs -- only for pointing to the appendices .}
\subsection{Environments}
\label{sec:env_descriptions}
We evaluate PBRR across the reward hacking benchmark environments used in \citet{pan2022effects}, summarized in Table \ref{tab:envs}. More details are in Appendix \ref{app:environments}. All preference labels are sampled from the Boltzmann distribution under the environment's ground-truth reward function over full trajectories. 
Simulating labels in this way is standard in RLHF evaluations (e.g., \citep{christiano2017deep, lee2021b, liang2022reward, metcalf2024sample}) and is what makes our particular domains tractable to study, since eliciting the significant number of preferences from expert human labelers would be prohibitively
costly. We address the fidelity of this simulation in Appendix~\ref{app:traj_pref_justification}, which explains why we elicit preferences over full trajectories rather than segments, and in Appendix~\ref{app:pref_noise} where we evaluate PBRR under increased preference noise.

\begin{table}[tb]
\centering
\setlength{\tabcolsep}{2.5pt}
\renewcommand{\arraystretch}{1.06}

\begin{adjustbox}{max width=\linewidth}
\begin{tabular}{L{2cm} L{2.9cm} L{1.3cm} L{1.3cm} c L{2.6cm} L{3.4cm} L{2.4cm}}
\toprule
\textbf{Name} & \textbf{Objective} & \textbf{State} & \textbf{Action} & \textbf{H} &
\textbf{Ref. policy} & \textbf{ $\hat{r}_{\text{proxy}}$ \& Summary of $\pi^*_{\hat{r}_{\text{proxy}}}$} & \textbf{Simulator} \\
\midrule

\textbf{Pandemic Mitigation} \citep{kompella2020reinforcement} & Design COVID-19 pandemic lockdown regulations to balance economic and health outcomes &
Cont., 312-dim &
Disc. $\{-1,0,1\}$ &
192 &
BC on real-world mitigation strategies from \citet{kompella2020reinforcement} &
 Omits political cost of lockdowns $\Rightarrow$ policy keeps high lockdown restrictions even at low infection rates &
Modified SEIR simulator \citep{kompella2020reinforcement} \\

\addlinespace[2pt]
\textbf{Glucose Monitoring}\newline \citep{dallaman2014uvapadova,fox2020deep} & Administer insulin to patient with Type I diabetes to prioritize patient health outcomes &
Cont., 96-dim &
1-d Cont. in $[0,1]$ &
5760 &
BC on \emph{few} expert demos \citep{willis1999proportional} &
Prioritizes reducing financial cost of treatment $\Rightarrow$ policy sacrifices patient health for low costs &
FDA-approved simulator \citep{dallaman2014uvapadova,fox2020deep} \\

\addlinespace[2pt]
\textbf{Traffic Control}\newline \citep{wu2021flow} & Control autonomous vehicle (AV) fleet on highway to maximize traffic flow &
Cont., 50-dim &
10-d Cont. in $[0,1]$ &
300 &
BC on \emph{few} human driving demos \citep{treiber2000congested} &
Maximizes mean velocity $\Rightarrow$ AVs block on-ramp so other vehicles can't merge onto highway&
FLOW highway simulator \citep{wu2021flow} \\

\addlinespace[2pt]
\textbf{AI Safety Gridworld} \citep{leike2017ai} & Water all tomatoes in a grid-world & 
Disc., \newline 36-dim, &
Disc. $\{0,1,2,3\}$ &
100 &
Policy trained to only water a subset of tomatoes &
Tomatoes look watered when they are not if the agent visits the sprinkler state $\Rightarrow$ policy never waters all tomatoes &
Toy environment from \citet{leike2017ai}\\
\bottomrule
\end{tabular}
\end{adjustbox}
\caption{Reward-hacking environments, used by \cite{pan2022effects}. H = horizon. BC = behavior cloning. Cont.~= Continuous. Disc.~= Discrete.}
\label{tab:envs}
\end{table}

\commentsh{Done!}

\subsection{Baselines}
\label{sec:baselines}
We compare PBRR against state-of-the-art and natural ablation baselines. More details are in Appendix \ref{app:baselines}.
\emma{Might be worth calling this proxy-reward rather than just state constrainedPPO}\commentsh{I only disagree because such a long name---State-Constrained-Proxy-Reward-PPO---makes the plot legend clunkier/take up more space; let me know if you feel strongly about this though}\eb{That's reasonable. It' not worth changing now, might be worth brainstorming something shorter but in this direction before we put it up on arxiv}
\begin{itemize}[leftmargin=*,nosep]
\item \textbf{State-Constrained-PPO~~~} A natural baseline is to optimize for the proxy reward function using the reference policy as a constraint in PPO, with no additional data collection. ~\cite{laidlaw2024correlated} showed that using a state-based divergence measure often yields better performance, so we optimize over the set of best-performing divergences amongst their proposals. 
\item 
\textbf{Online-State-Constraint~~~} This method also uses the best divergence-regularized objectives of \cite{laidlaw2024correlated}, but now the reward function is learned from scratch. 
Preferences are elicited between trajectories sampled from the learned policy and the reference policy. 
\item \textbf{Online-RLHF~~~} \emma{Maybe add exploratory? Does RRM also try to maximize the uncertainty?} \commentsh{Yeah RRM does; also I disagree with adding ``exploratory" in the baseline name, as \cite{christiano2017deep} never emphasize exploration and I imagine if that was their goal they might have developed another approach----this exploration strategy is clearly flawed}(\citet{christiano2017deep}) Their method learns a reward model from scratch and maintains a reward ensemble.  Trajectory pairs with the highest predictive uncertainty are selected for preference elicitation. To construct a stronger baseline, the reference policy is used for initial exploration data. See Appendix \ref{app:online_rlhf_abalations} for further ablations of this baseline drawing on work in data-efficient RLHF for robotics (e.g., \cite{lee2021pebble, liang2022reward, metcalf2024sample}.
\item \textbf{Residual Reward Modeling (RRM)~~~} (\cite{cao2025residual}) 
This method learns the correction term $g$ for Eq.~\ref{eq:r_decompisition} using the standard cross-entropy loss.  It elicits preferences over trajectory pairs with the highest predictive uncertainty sampled from the policy optimized for the current proxy reward function. 
\item \textbf{RRM + State-Constraint~~~} Here we adapt the Online-State-Constraint baseline to learn the correction term $g$ for Eq. \ref{eq:r_decompisition} to repair a proxy reward function.
\item \textbf{$r$-PPO}: PPO using the ground truth reward model. Serves as an oracle upper bound.
\end{itemize}

\vspace{-2mm}
\begin{figure}[tb]
    \centering    \includegraphics[width=0.48\textwidth]{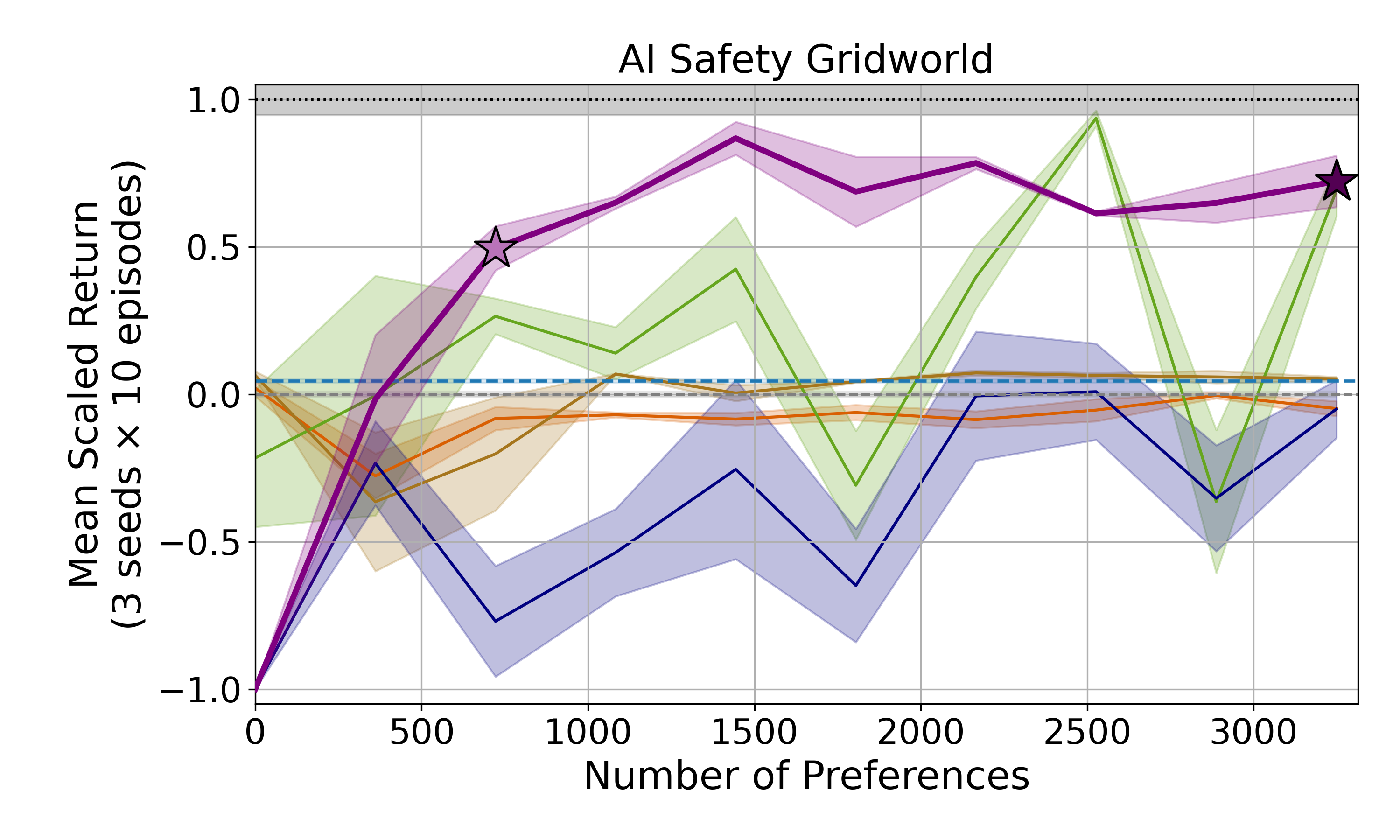}
    \includegraphics[width=0.48\textwidth]{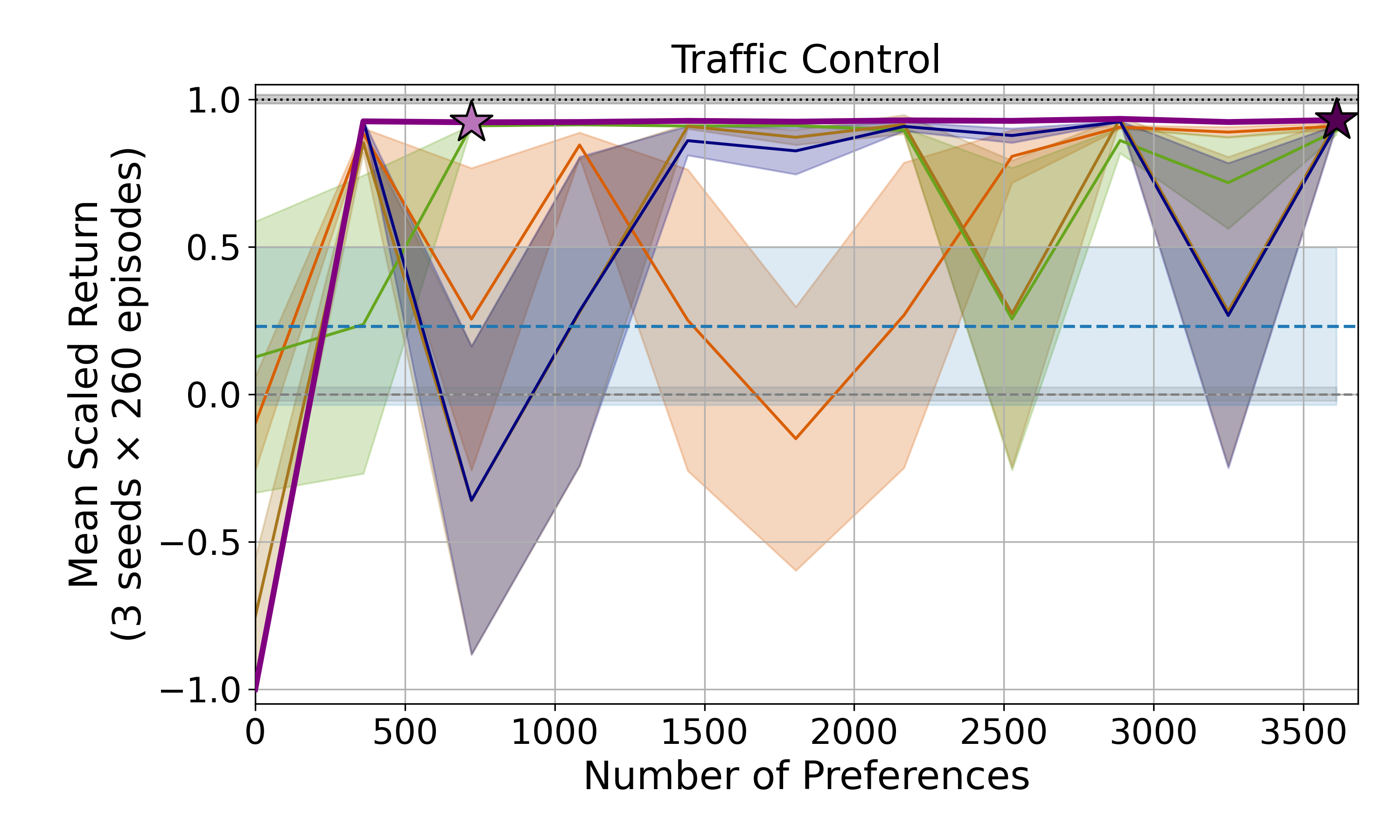}\\[0.5em]
    \includegraphics[width=0.48\textwidth]{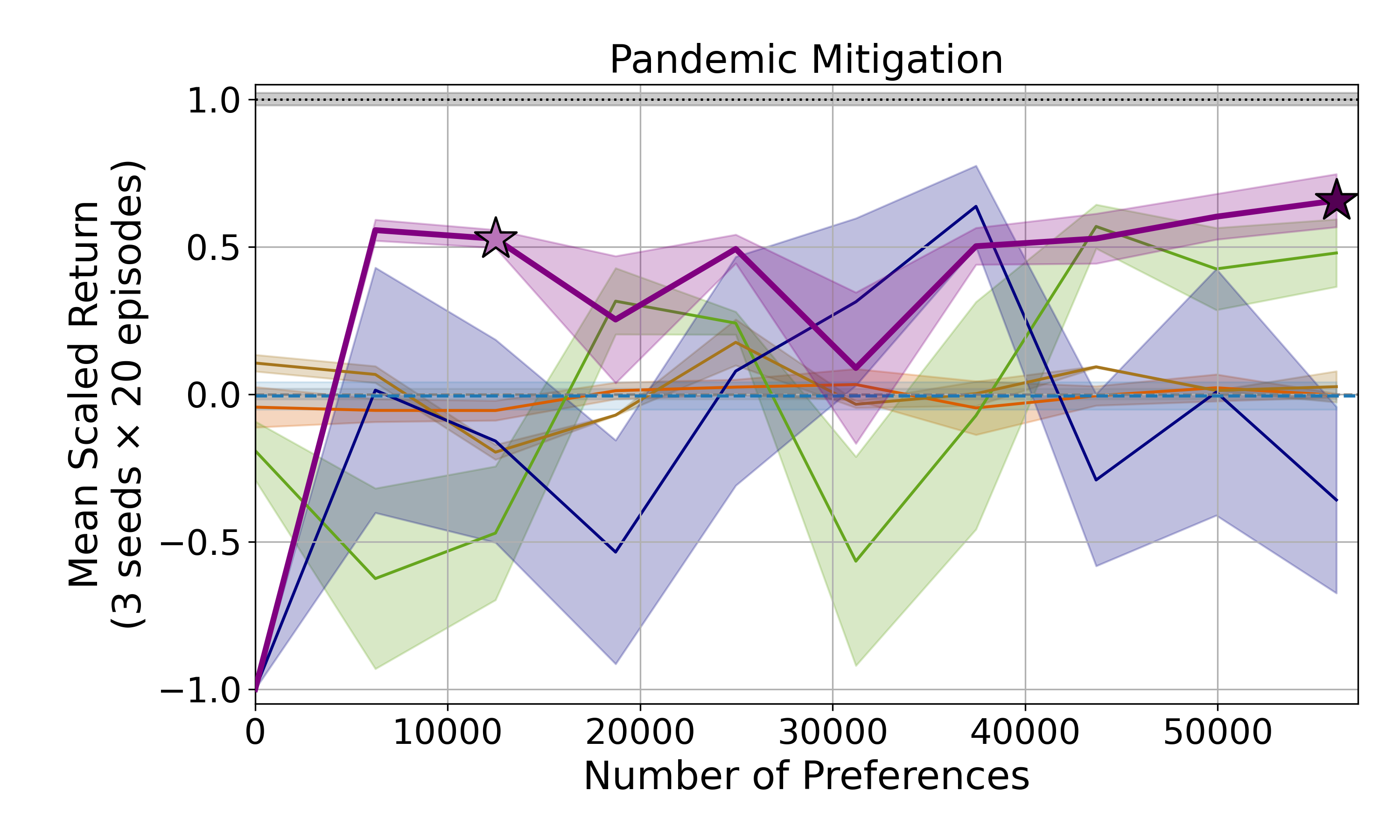}
    \includegraphics[width=0.48\textwidth]{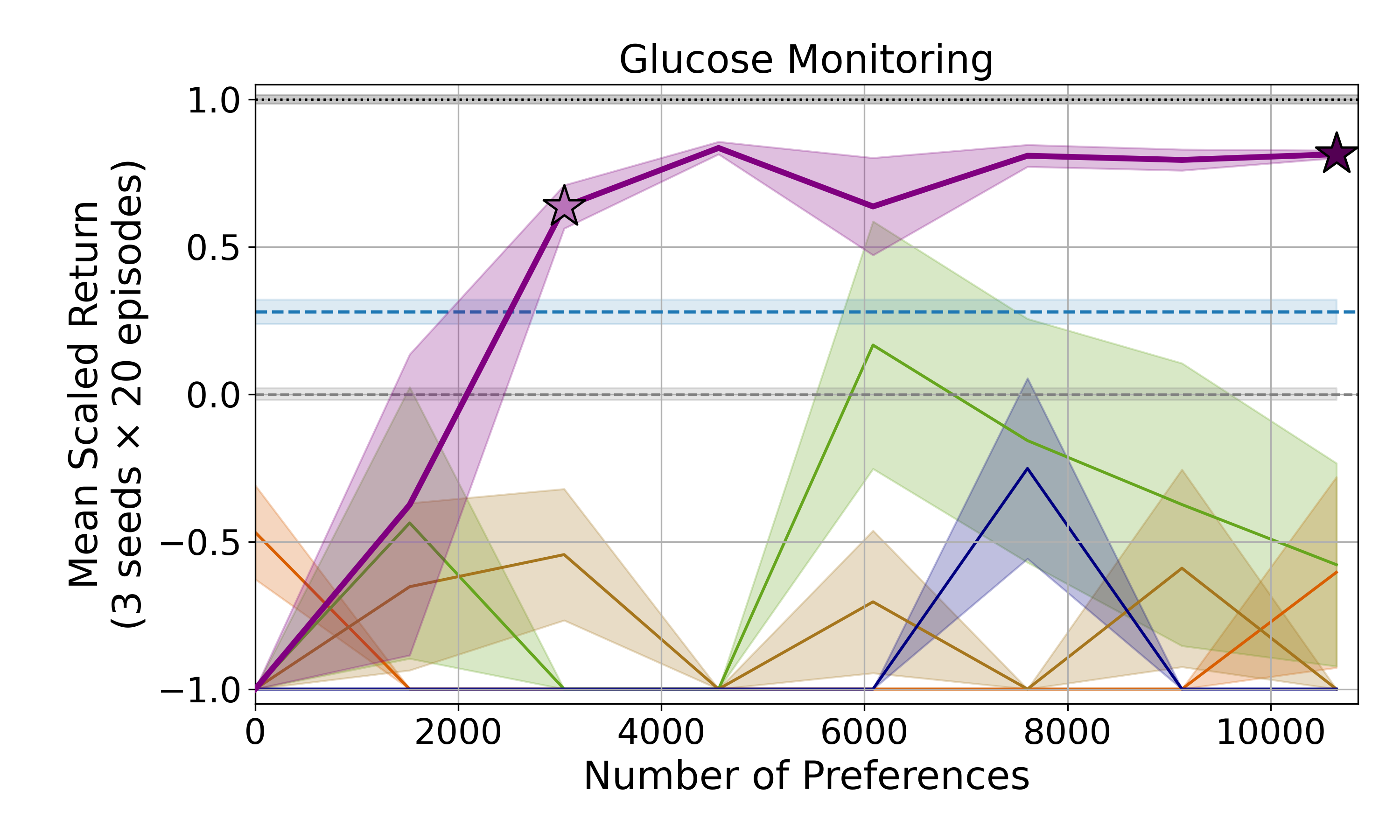}

    \includegraphics[width=0.65\textwidth]{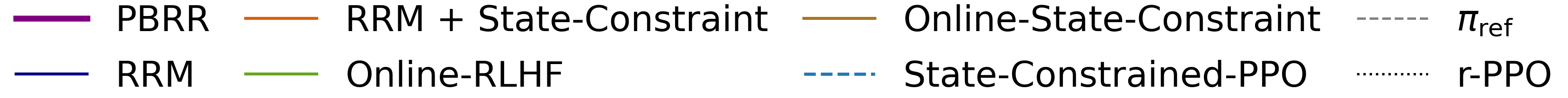}
    \caption{Mean return under the ground-truth reward function achieved by PBRR compared to the baselines from Section \ref{sec:baselines}, averaged over 3 random seeds. The mean return for each seed is scaled so that, after scaling, the reference policy has a mean return of $0$ and $r$-PPO has a mean return of $1$. Values are clipped to $[-1, 1]$, with values below $-1$ indicating very poor  performance. Shaded regions indicate the standard error. Unscaled, unclipped returns and further plotting details are provided in Appendix \ref{app:unscaled_res} and Appendix \ref{app:plot_details}, and statistical significance is reported in Appendix \ref{app:significance}.\textcolor{lightpurple}{\ding{72}} and \textcolor{darkpurple}{\ding{72}} mark PBRR's performance after two proxy reward function updates and the final update respectively.}
    \label{fig:main_results}
\vspace{-6mm}
\end{figure} 
\subsection{Main results: repairing $\hat{r}_\text{proxy}$ with preferences}
\label{sec:main_results}
\eb{I know I suggested splitting results and discussion, but reading it now, I think it'll be better to stronger to integrate the two. It also would be useful to reorganize this into key takeaways, with a bit of support for each. For example:
(1) jump start performance, (2) asymptotic performance, (3) stability (4) ablations on our method}
Results are shown in Figure \ref{fig:main_results}, with additional experiments in Appendix \ref{app:additional_results}. Using PBRR to repair the proxy reward function with preferences is more data efficient than learning a reward function \textit{ab initio} without the human-specified prior, and alternative approaches for repairing the proxy reward function. Moreover, PBRR outperforms State-Constrained-PPO, indicating that comparisons to the reference policy enable efficient learning of $g$, even when the reference policy itself is not performant enough to successfully employ any divergence-based method we consider. In Appendix \ref{app:significance} we report statistical significance: in all environments except Traffic Control, PBRR outperforms every baseline with statistical significance in at least one iteration; in all environments, after the first round of preference elicitation, PBRR is never outperformed by any baseline. 


\textbf{\textcolor{lightpurple}{\ding{72}}
Better Jump Start Performance ~~~} After the first two updates of the proxy reward function, PBRR attains substantially higher performance than all baselines in all environments except for Traffic Control, where PBRR matches the performance of Online-RLHF.  \emma{It would be nice to have the bold to indicate the take home message. Could be better initial performance, but it's more about the quickly learning. Feel free to tweak the tagline}

\textbf{\textcolor{darkpurple}{\ding{72}} Strong Final Performance~~~} Within the preference budgets considered, PBRR matches or outperforms all baselines in every environment. The performance of RRM indicates that the proxy reward function alone does not provide enough exploration guidance to learn a correction term even after eliciting a large dataset of preferences; see Appendix \ref{app:safety_gw_cao_analysis} for a qualitative analysis. PBRR overcomes this limitation by leveraging the reference policy to direct exploration.

\textbf{Stability~~~} PBRR is substantially more stable in the AI safety gridworld, Traffic, and Pandemic environments. While other methods can eventually match its performance with enough preferences, their instability makes them impractical without ground-truth evaluation. Unlike PBRR, they show oscillatory or degrading performance as more preferences are collected, since small changes in the reward function can cause large changes in policy performance. Appendix~\ref{app:safety_gw_rlhf_analysis} analyzes this effect, explaining why Online-RLHF is unstable in the AI Safety Gridworld.

\textbf{Outperforming the reference policy $\pi_{\text{ref}}$~~~} PBRR induced policies that outperform the reference policy in all environments. The reference policy used by PBRR need not be performant; it must only provide a useful comparison to the proxy reward function's induced policy. In Appendix \ref{app:pbrr_random_ref_pol}, we empirically show that PBRR continues to match or outperform all baselines when using a randomly initialized reference policy.

\textbf{Optimism Assumption~~~} PBRR does not require the input reward function to be optimistic---note the decay parameters in Algorithm \ref{alg:exploration}---though it may do particularly well in such a setting. We explore the performance of PBRR when the input reward function is not optimistic in all domains, and find that PBRR continues to be the most performant method. In the Glucose Monitoring environment, the optimism assumption does not hold; policies that maximize patient health outcomes, i.e., are optimal under the ground-truth reward function, are penalized under the proxy reward function due to their financial cost. PBRR still outperforms all baselines. Appendix~\ref{app:pessimistic_proxy} reports similar findings in the other
environments when repairing proxy reward functions that are not optimistic, where
PBRR continues to outperform the baselines and thus remains a strong choice when the optimism assumption does not hold.



\begin{figure}[h]
\vspace{-2mm}
    \centering
    \includegraphics[width=0.48\textwidth]{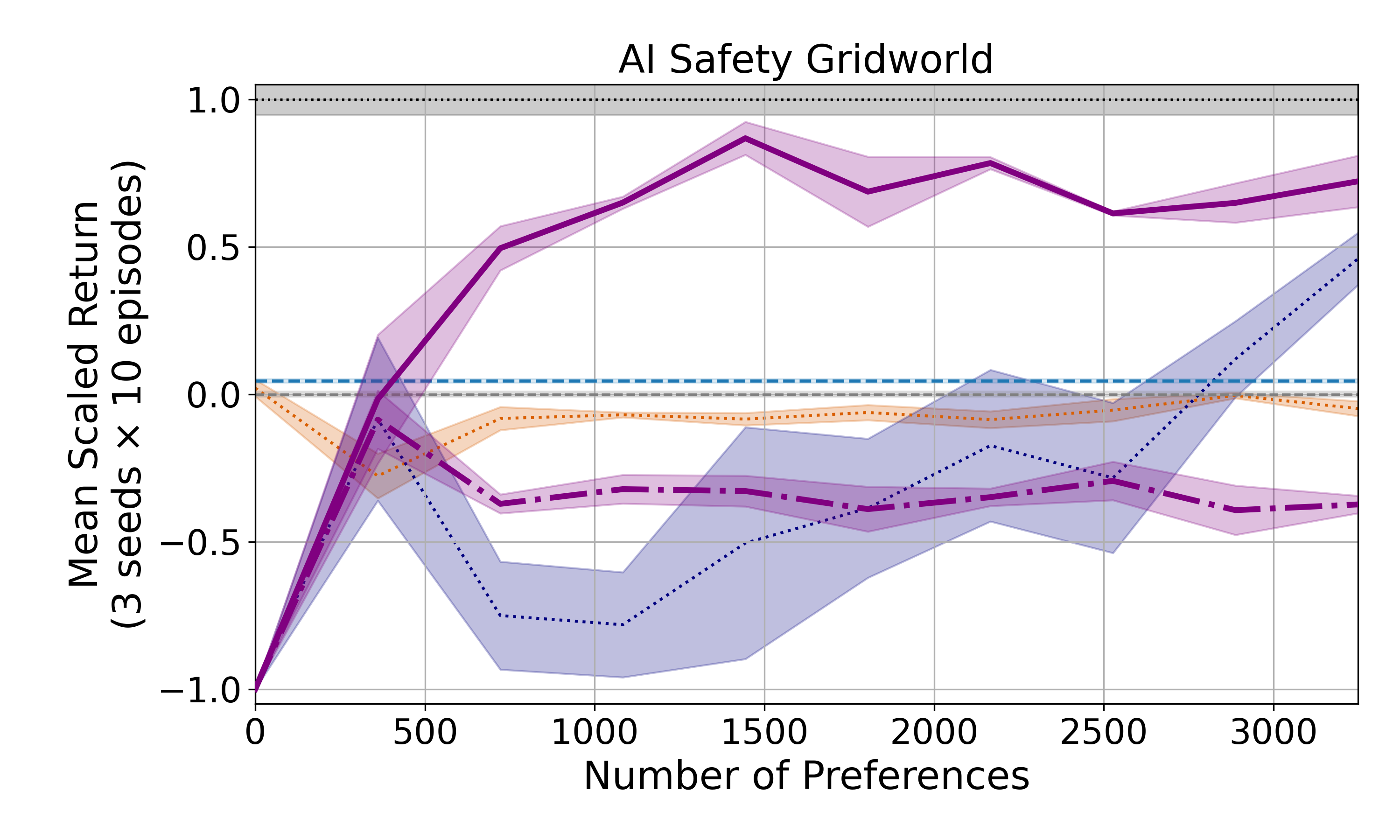}
    \includegraphics[width=0.48\textwidth]{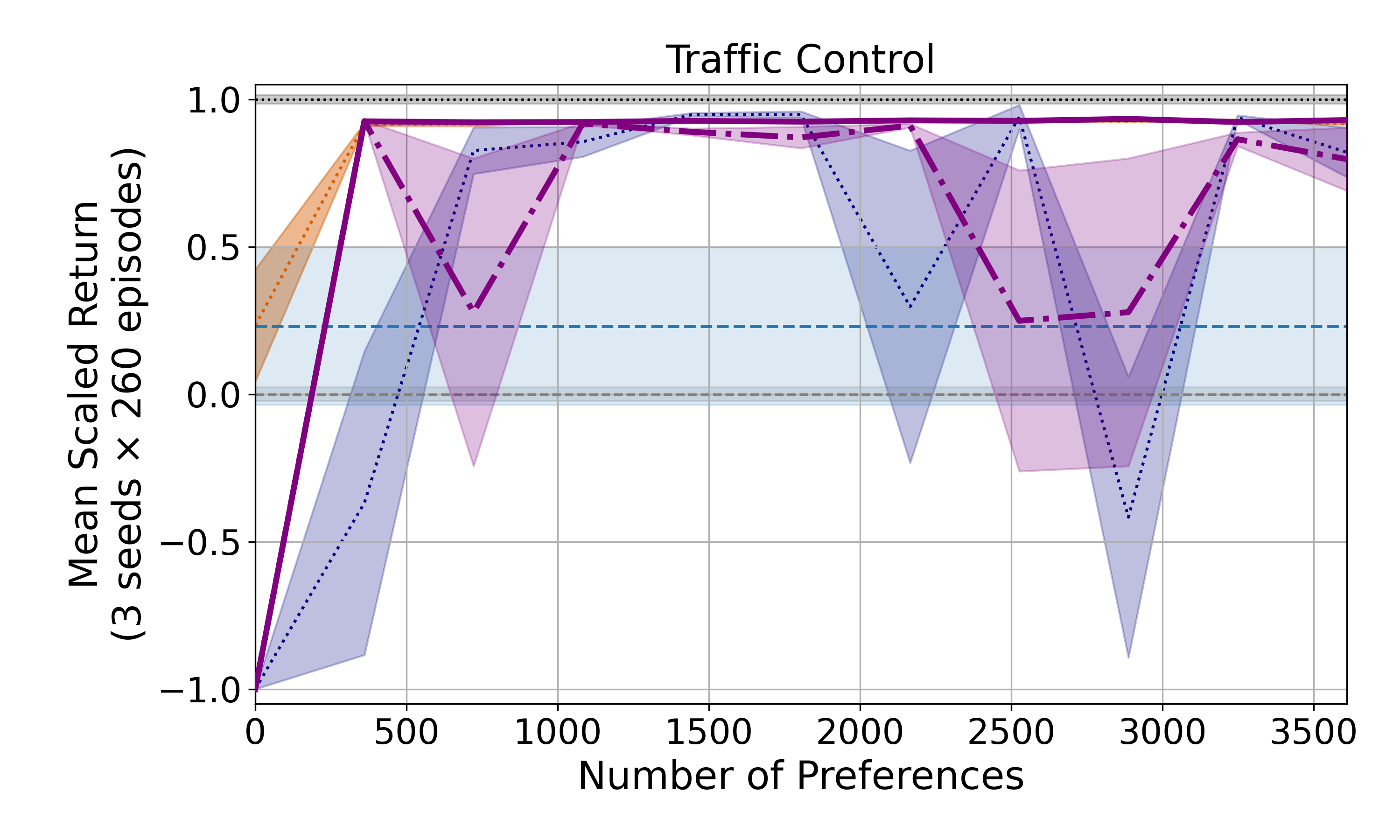}\\[0.5em]
    \includegraphics[width=0.48\textwidth]{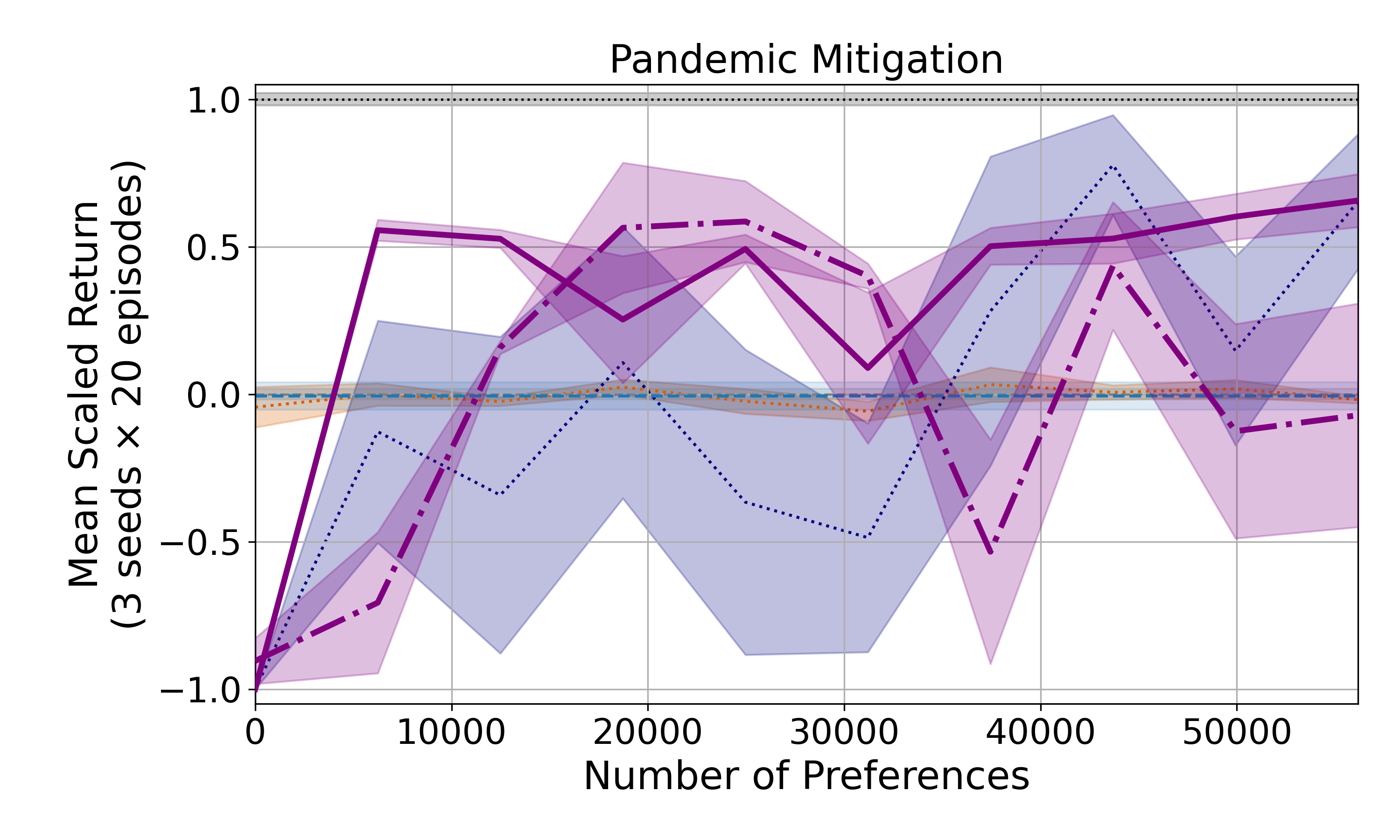}
    \includegraphics[width=0.48\textwidth]{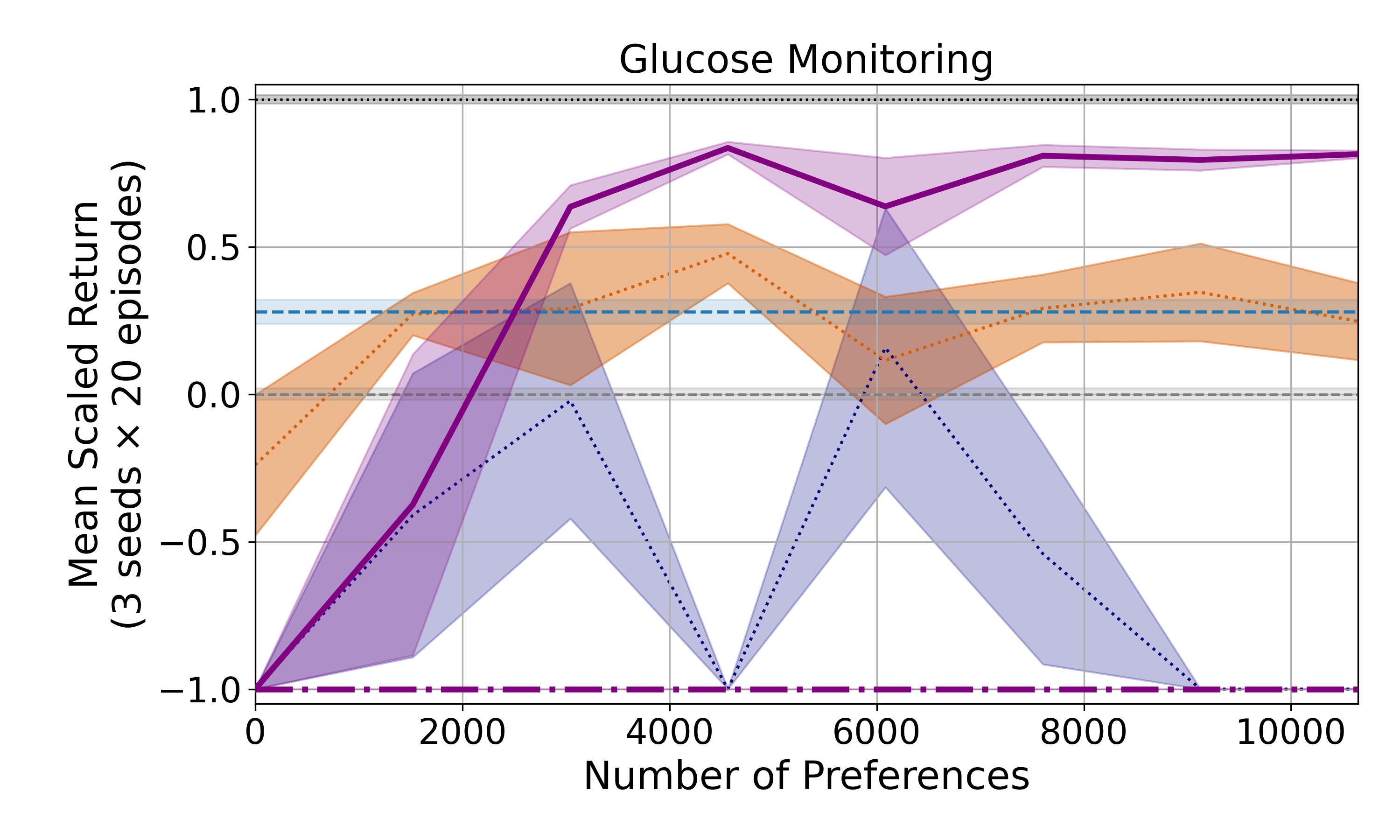}
    \includegraphics[width=0.99\textwidth]{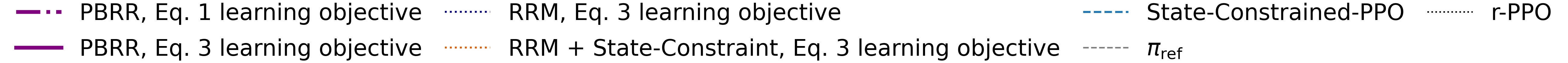}
    \caption{Mean return under the ground-truth reward function achieved by PBRR, compared to (i) PBRR using the standard preference-learning objective (Eq.\ref{eq:loss}) instead of our proposed objective (Eq.\ref{eq:reg-pref-loss}), and (ii) other methods that repair the proxy reward function equipped with our proposed objective. See Figure \ref{fig:main_results} for plotting details.}
    \label{fig:abalation}
\vspace{-3mm}
\end{figure}

\label{ref:main_results}
\subsection{Ablation Study: Preference Learning Objective vs. Exploration Strategy }
\label{sec:abalations}

We also sought to isolate the contributions of our preference-learning objective (Eq.~\ref{eq:reg-pref-loss}) and exploration strategy (outlined in Algorithm \ref{alg:exploration}). Figure \ref{fig:abalation} shows that repairing the proxy reward function with PBRR but using the standard loss in Eq. \ref{eq:loss} instead of Eq. ~\ref{eq:reg-pref-loss} yields substantially less stable performance and a lower mean-return under the ground-truth reward function across all environments. Without $\mathcal{L}^+$ and $\mathcal{L}^-$ from Eq. \ref{eq:reg-pref-loss}, the updated proxy reward function incorrectly assigns a higher reward to the suboptimal actions taken by the reference policy. Appendix~\ref{app:safety_gw_pbrr_no_reg_analysis} provides a qualitative analysis in the AI safety gridworld, and Appendix~\ref{app:pbrr_reg_ablation} reports further ablations of each regularization term to empirically validate their effectiveness, as they are not covered by our theoretical analysis. Conversely, using PBRR’s objective (Eq.~\ref{eq:reg-pref-loss}) within the RRM or RRM+State-Constraint baselines fails to match PBRR’s data efficiency in all environments except Traffic Control. This shows that both 
PBRR’s preference-learning objective and  exploration strategy are important to efficiently repair the proxy reward function with preferences.

Appendix~\ref{app:retraining_ppo} empirically shows that the proxy reward function repaired by PBRR can reliably be used to retrain new policies under different random seeds. Moreover, optimizing the repaired reward function with PPO for longer yields the same or better performance for the Traffic and Pandemic environments, highlighting that the repaired reward function is robust to extended RL optimization---although in the Glucose Monitoring environment, extended RL optimization beyond that considered when updating the proxy reward function eventually leads to performance deterioration.

\section{Conclusion}
We introduce Preference-Based Reward Repair (PBRR), a framework to repair a human-specified proxy reward function by learning a transition-level correction from preferences. Across a diverse set of  benchmarks, spanning autonomous vehicle traffic control, pandemic lockdown regulation design, and insulin regulation for diabetes patients, PBRR achieves higher performance with greater stability and fewer preferences than methods that either learn a reward function from scratch or modify the proxy reward function using alternative strategies. Our ablations show that both components of PBRR are necessary: the exploration strategy that leverages a supplied reference policy to identify transitions over which the proxy reward function is incorrect, and the preference-learning objective that encourages only correcting the proxy reward function on transitions where it incorrectly assigns high reward. Our theoretical analysis further shows that a variant of PBRR attains a sub-linear regret bound comparable with prior work.
Overall, our results suggest that repairing a human-specified proxy reward function with PBRR, rather than learning a reward function from scratch, is a data-efficient path to alignment in complex, sequential decision making tasks. 
\newpage



\bibliography{main}
\bibliographystyle{tmlr}

\appendix

\textbf{Appendix}

\section{Why elicit preferences over trajectories instead of trajectory segments?}
\label{app:traj_pref_justification}

We focus on eliciting preferences over pairs of trajectories, rather than over pairs of trajectory segments. In practice, eliciting preferences over full trajectories may introduce credit assignment issues that are resolved when eliciting preferences over shorter trajectory segments. The latter is more common than the former (e.g., \cite{christiano2017deep}), and we hypothesize that this credit assignment issue results in the noise we observe in our results in Section \ref{sec:results}.

Despite the limitation of eliciting preferences over full trajectories rather than shorter segments, this remains a more principled approach for simulating human feedback. In particular, \cite{knox2022models} studied different models of human preferences, finding that the difference in regret between trajectory segments is more predictive of the human preference label than the difference in the sum of rewards. To simulate human preference labels---as we do in this work to avoid collecting preferences from real humans---we therefore should label preferences in accordance with regret under the ground-truth reward function. Unfortunately, for the real-world tasks we consider, computing the regret with respect to the ground-truth reward function is exceedingly difficult and computationally expensive due to the continuous state and action spaces. Consequently, we label preferences over trajectory pairs by the difference in the sum of rewards. These trajectories begin and end in the same state. Therefore, preferences follow the change-in-expected-return model, also proposed by \cite{knox2022models}. 

Labeling preferences over shorter trajectory segments determined by the difference in sum of rewards---which \cite{knox2022models} call the partial return preference model---- results in preference labels that empirically do not match human judgments. We follow the recommendation of \cite{knox2022models} and avoid using this partial return preference model to simulate preference labels. 

The tradeoff however is that credit assignment for reward learning becomes more challenging due to our reliance on eliciting preferences over full trajectories. We argue that simulating higher-fidelity preference labels is more principled, as it better reflects how our methods would perform when no ground-truth reward function is available and real human preferences must be elicited. In other words, trajectory-level preferences allow us to focus on evaluating whether a method can repair a misspecified proxy reward function, rather than conflating this with its capacity to learn a reward function from a misspecified model of human preferences. See \cite{knox2022models} for further discussion on different models of human preferences and their pitfalls, noting that the preference models they consider are equivalent when eliciting preferences over full trajectories that begin and end in the same state in MDPs with deterministic transition dynamics.

\section{Environment Details}
\label{app:environments}
We use the same environments and  configurations as \cite{laidlaw2024correlated} and \cite{pan2022effects}, except for the AI safety gridworld. To make the task from the AI safety gridworld domain harder, we reconfigure the placement of the tomatoes in the grid-world. Full implementation details, including this grid-world configuration, are available in our codebase. We provide a brief description of each environment, as well as its accompanying proxy reward function and reference policy below.

\textbf{Pandemic Mitigation ~~~} The agent controls the level of lockdown restrictions imposed on a population by observing COVID-19 test outcomes, as simulated by a modified SEIR model \citep{kompella2020reinforcement}. The proxy reward function captures epidemiological and economic outcomes but omits the political cost associated with aggressive lockdown regulations. Optimizing for the proxy reward function induces a policy that maintains a high lockdown level even when infection rates are low. The supplied reference policy is trained via behavioral cloning on a combination of government strategies used during the pandemic.

Observations are vectors of 312 continuous values; an action is a single discrete value in $\{-1, 0,1\}$; the horizon is 192 time steps. 

\textbf{Glucose Monitoring ~~~} The agent controls the insulin administered to a simulated patient with Type I diabetes to maintain healthy glucose levels in an FDA approved simulator \citep{dallaman2014uvapadova, fox2020deep}. The proxy reward function prioritizes reducing the financial cost of insulin and hospital visits, while the ground-truth reward function is a standard measure of health risk. Optimizing for the proxy reward function induces a policy that minimizes financial costs but not the patient's overall health risk. The supplied reference policy is trained via behavioral cloning on \textit{only a handful} of demonstrations executed by a PID controller tuned by \cite{willis1999proportional}, illustrating a case where only a handful of expert demonstrations are available. 

Observations are vectors of 96 continuous values; an action is a single continuous value in $[0,1]$; the horizon is 5760 time steps. 

\textbf{Traffic Control ~~~} The agent controls a fleet of autonomous vehicles on an on-ramp attempting to merge into traffic on a highway with simulated human drivers \citep{wu2021flow}. The proxy reward function prioritizes maximizing the mean velocity of all vehicles. When optimized for, it induces a policy where the autonomous vehicles block the on-ramp, allowing highway traffic to maintain maximum speed while preventing other vehicles from entering. The ground-truth reward function instead prioritizes the mean commute time of all vehicles. The supplied reference policy is trained via behavioral cloning on \textit{only a handful} of demonstrations executed by an Intelligent Driver Model \citep{treiber2000congested} aimed to mimic human driving. 

Observations are vectors of 50 continuous values; actions are vectors of 10 continuous values in $[0,1]$; the horizon is 300 time steps. 

\textbf{AI Safety Gridworld ~~~} In the only toy-task we consider, the agent moves around a grid-world with the objective of watering all tomatoes by visiting tomato-containing states. There exists a sprinkler state that makes the tomatoes look watered, i.e., the agent attains positive reward under the proxy reward function, when the tomatoes are not actually watered, i.e., the agent attains no reward under the ground-truth reward function. This environment was introduced by \cite{leike2017ai}. The supplied reference policy is trained using the ground-truth reward function in a grid-world layout that contains only a small subset of the tomatoes present in the task we use for evaluation.

Observations are vectors of 36 discrete values; an action is a single discrete value in $\{0,1,2,3\}$; the horizon is 100 time steps. 

\section{Optimizing for $\hat{r}$ while avoiding reward hacking}
\label{app:constrained-ppo-outline}
Some prior work (e.g., \citep{ziegler2019fine, ouyang2022training, bai2022training, glaese2022improving, openai2022chatgpt, touvron2023llama}) assumes access to a reference policy $\pi_{\mathrm{ref}}$, and then optimizes for the following objective:
\begin{equation}
\label{eq:constrained-RL}
    \text{maximize} \quad
J_{\hat{r}}(\pi)
\;-\;
\beta F(\pi, \pi_{\mathrm{ref}})
\end{equation}

where $F$ is some measure of divergence between $\pi$ and $\pi_{\mathrm{ref}}$. For example, $F$ can be defined as the expected KL divergence between the action distributions of $\pi$ and $\pi_{\mathrm{ref}}$: 
\[
F(\pi, \pi_{\mathrm{ref}}) = (1-\gamma) \mathbb{E}_{\pi}\!\Biggl[
    \sum_{t=0}^{\infty}
        \gamma^{t}\,
        D_{\mathrm{KL}}\!\bigl(
            \pi(\,\cdot\mid s_t)
            \,\big\|\pi_{\mathrm{ref}}(\,\cdot\mid s_t)
        \bigr)
\Biggr]
\]

\cite{laidlaw2024correlated} proposes other divergence measures that dictate how $\pi$ can deviate from $\pi_{\mathrm{ref}}$, which we consider in our empirical results. Eq. \ref{eq:constrained-RL} applies a divergence penalty between the learned policy $\pi$ and $\pi_{\mathrm{ref}}$ to balance between optimizing for $\hat{r}$ without straying too far from a known behavior distribution, i.e., $\pi_{\mathrm{ref}}$. While this approach can attain better performance under the ground-truth reward function than only optimizing for the misspecified proxy reward function, it requires a sufficiently performant reference policy or proxy reward function $\hat{r}$ to attain near-optimal performance under the ground-truth reward function. Otherwise, constraining the learned policy to be close to the reference policy as defined by $F$ will fundamentally limit the performance of the resulting policy. On the other hand, reducing the divergence penalty by lowering $\beta$ can weaken the constraint on $\pi$, allowing it to exploit misspecifications in $\hat{r}$ and potentially learn a policy that is substantially sub-optimal with respect to the ground-truth reward function. For this reason, \citet{laidlaw2024correlated} use a reasonably performant reference policy in their experiments. In contrast, we operate in a setting where no such policy exists; under these conditions, optimizing Eq.~\ref{eq:constrained-RL} with any of the divergence measures considered by \citet{laidlaw2024correlated} and any choice of $\beta$ is unlikely to yield near-optimal performance with respect to $r$. Therefore, in our setting, $\hat{r}$ itself must be updated.

\section{Baselines for learning and repairing reward functions}
\label{app:baselines}
Each baseline described below either learns a reward function \textit{ab initio} or assumes access to a manually specified proxy reward function and learns an additive correction term \( g(s, a, s') \) as used in Equation~\ref{eq:r_decompisition}. In both cases, the additive correction term and the reward function are parametrized as a neural network or an ensemble of neural networks, depending on the baseline. Here we describe how each baseline constructs a batch of trajectory pairs to elicit preferences over. Upon eliciting preferences, the resulting $(\tau_1, \tau_2, \mu)$ samples are added to dataset $\D_t$ and---unless otherwise stated---the reward function or correction term parameters are updated using the standard preference loss in Eq. \ref{eq:loss} given $\D_t$. 

Our approach and all baselines use Proximal Policy Optimization (PPO)~\citep{schulman2017proximal} to train a policy with the current estimate of the reward function $\hat{r}$. Like prior RLHF approaches (e.g., \cite{christiano2017deep, lee2021pebble, lee2021b}), we decouple reward learning and policy learning: PPO samples environment rollouts independently of those used to update the reward function. As a result, differences between methods lie primarily in their reward modeling and data collection strategies, rather than in their policy optimization. 

\subsection{Online-RLHF Baseline} 
This baseline adapts the methodology of \cite{christiano2017deep} and \cite{lee2021pebble} to our problem setting. At each iteration, an ensemble of randomly initialized reward models is updated using all collected preferences over trajectories. A new batch of preferences for elicitation is constructed as follows: 

At each iteration we sample up to 200 trajectories from the current policy, which are then added to a candidate batch. For a more fair comparison with our approach that assumes access to a safe reference policy, the candidate batch also contains trajectories sampled from the supplied reference policy. We note that the candidate batch initially contains trajectories sampled from the reference policy----in addition to a policy trained with a randomly initialized reward model---rather than trajectories sampled from a policy pre-trained with the state-entropy objective proposed by \cite{lee2021pebble}, which is a follow-up work to \cite{christiano2017deep}; this baseline harnesses the safe reference policy for exploration. 

All possible pairs of trajectories are constructed from the candidate batch, and the top $k$ pairs with the highest variance over the predicted preference probabilities---computed via preference model $P( .|\tilde{r})$ with respect to the reward model ensemble---are labeled with preferences. Note that this baseline may elicit preferences between trajectories sampled from the reference policy and the current policy if those pairs have the highest variance in predicted preference probability.

\subsection{Residual Reward Model (RRM) Baseline} 
This baselines mirrors the method of \citet{cao2025residual}, and matches the Online-RLHF baseline except a correction term \( g \) for Eq. \ref{eq:r_decompisition} is learned instead of learning a reward function \textit{ab initio}. A \texttt{tanh} function is also applied to the output of the learned correction term \( g \) to match the implementation of \citet{cao2025residual}. The primary difference between this baseline implementation and \citet{cao2025residual} is that they adopt Soft Actor-Critic (SAC) while we use Proximal Policy Optimization (PPO) for a stringent comparison with our other baselines.

\subsection{State-Constraint Reward Learning Baselines} Here we describe both the Online-State-Constraint and RRM + State-Constraint baselines. These baselines adapt recent insights from alignment research in large language model (LLM) settings, which suggest that constraining the learned policy to remain close to a reference policy can improve performance when learning from preferences. We build on the methodology proposed by \citet{dong2024rlhf}, modifying it to suit our setting. In particular, unlike \citet{dong2024rlhf}, our policies are not parameterized as LLMs and therefore we cannot exploit stochastic decoding to generate diverse outputs for preference elicitation. To get around this, we sample trajectories from two different policies instead; we sample one trajectory from the learned policy and one from the reference policy. 

At each iteration, after learning a reward function from the constructed dataset of preferences, we learn a policy from that reward function using the divergence-regularized objective in Eq. \ref{eq:constrained-RL}. We follow the insights of \cite{laidlaw2024correlated} and select the divergence measure to be the one that induces the best performance as outlined in Appendix \ref{app:contrained-ppo}. 

We consider two baselines that follow this approach; one that learns a reward function \textit{ab initio} and one that repairs an existing reward function. For each baseline we additionally implement two versions: one that keeps $\pi_\text{ref}$ fixed---as presented in Section \ref{sec:results}, and one that updates $\pi_\text{ref}$ to be the policy learned in the previous iteration---as presented in Appendix 
\ref{app:pi_ref_updating_exp_res}.

\textbf{Online-State-Constraint ~~~} At each iteration $t$, we roll-out the reference policy $\pi_\text{ref}^t$ and the current learned policy $\pi^t_{const}$, where $\pi^t_{const}$ is found by optimizing for the objective in Eq. \ref{eq:constrained-RL} with $\hat{r}_t$ and $\pi_\text{ref}^t$. The specific divergence measure used for this objective is chosen as the measure that induces the best performance as described in Appendix \ref{app:contrained-ppo}. $\sqrt{k}$ trajectories are then sampled from $\pi_\text{ref}^t$ and $\pi^t_{const}$ respectively, and $k$ exhaustive $(\tau_1, \tau_2)$ pairs are constructed where $\tau_1 \sim \pi_\text{ref}^t$ and $\tau_2 \sim \pi_{const}^t$. Preferences are elicited over all $k$ pairs and added to the dataset $\D$. Depending on the baseline variant, the reference policy at the next iteration is moved such that $\pi_\text{ref}^{t+1} = \pi_\text{ref}^{t}$ or $\pi_\text{ref}^{t+1} = \pi_{const}^{t}$.

\textbf{RRM + State-Constraint ~~~}  
This baseline follows the same procedure as described above, with the exception that a correction term \( g \) is learned and applied to the initial proxy reward function, rather than learning a reward function \textit{ab initio}.

\section{Implementation Details}
\label{app:imp_details}
All experiments were implemented with Python 3.9 and PyTorch 2.7 \citep{paszke2019pytorch}, largely building off of the codebase provided by \cite{laidlaw2024correlated}. 

\subsection{Policy and Reward Model Architectures}

We use the same policy network architectures as \cite{laidlaw2024correlated}. For the pandemic, traffic, and AI safety gridworld environments, we use feedforward policy networks with the following architectures: two hidden layers of 128 units for the pandemic environment, four hidden layers of 512 units for the traffic environment, and four hidden layers of 512 units for the AI safety gridworld environment. In the glucose environment, the policy is implemented as a three-layer LSTM, each layer containing 64 units.

When learning a reward function \textit{ab initio}, we parameterize the reward function as a feedforward neural network with 5 hidden layers of 256 units for the pandemic environment and 5 hidden layers of 512 units for the glucose, traffic, and AI safety gridworld environments. For the Online-RLHF baseline, we maintain an ensemble of 5 reward functions. When learning a corrective term $g$ for a specified proxy reward function, we parameterize $g$ as a feedforward neural network with 5 hidden layers of 512 units for the pandemic, glucose, and AI safety gridworld environments, and 3 hidden layers of 256 units for the traffic environment. For the RRM baseline, we maintain an ensemble of 3 reward functions. Table \ref{tab:reward-training-hyperparams} shows the additional hyperparameters used for training the reward function and correction term. The Adam optimizer was used to learn the reward function or correction term parameters with default Pytorch hyperparameter values. 

\begin{table}[h]
\centering
\caption{Hyperparameters used for learning a reward function \textit{ab initio} and learning a correction term to repair a proxy reward function across all environments.}
\begin{tabular}{llccc}
\toprule
\textbf{Environment} & \textbf{Repair Proxy Reward Function?} & \textbf{Learning Rate} & \textbf{Weight Decay} & \textbf{Epochs} \\
\midrule
Pandemic & True & 0.001   & False & 200 \\
Pandemic & False     & 0.0001  & False & 200 \\
Glucose  & True & 0.0001  & False & 200 \\
Glucose  & False    & 0.0001  & True  & 200 \\
Traffic  & True & 0.0001  & True  & 50  \\
Traffic  &False     & 0.001   & False & 50  \\
AI safety gridworld    & True & 0.0001  & True  & 200 \\
AI safety gridworld    & False     & 0.0001  & True  & 200 \\
\bottomrule
\end{tabular}
\label{tab:reward-training-hyperparams}
\end{table}
\commentl{@Stephane, using scientific notation for the learning rate column might make it a bit easier to read.}

All hyperparameters listed above were found via the following process: for each environment, we constructed a dataset of exhaustive preferences between trajectories sampled from a policy trained with the ground-truth reward function and the proxy reward function respectively, and trajectories sampled from the reference policy. Separately for learning a reward function from scratch and a corrective term for Eq. \ref{eq:r_decompisition}, we performed a grid-search over the following hyperparameter candidates when learning from the constructed dataset of preferences:
\begin{itemize}
    \item learning rate: [0.001, 0.0001]
    \item weight decay: [True, False]
    \item number of epochs: [50, 100, 200]
    \item number of hidden layers: [3, 5]
    \item number of units per hidden layer: [256, 512]
\end{itemize}

We chose the hyperparameter values that invoked the lowest loss on a held out test-set. Our objective was to identify the hyperparameters that most effectively enabled the learned reward function to distinguish between trajectories of varying optimality.

\subsection{Policy and Reward Model Initialization}

For the glucose, traffic, and pandemic environments, at each iteration, i.e., before any policy is learned, we always initialize the policy weights to be the reference policy weights. Following the methodology of \cite{laidlaw2024correlated}, we randomly initialize the policy weights for the AI safety gridworld  environment because otherwise we do not observe reward hacking when optimizing for the initial proxy reward function. 

For all environments and experiments, at each iteration we re-initialize the reward function or correction term's weights. We then re-train the reward function or correction term on all collected samples 
$(\tau_1, \tau_2, \mu) \in \mathcal{D} = \bigcup_{t} \mathcal{D}_t$, not just the most recently collected batch of preferences $\mathcal{D}_t$. 

\subsection{Controlling divergence from $\pi_\text{ref}$}
\label{app:contrained-ppo}
For the State-Constrained-PPO, Online-State-Constraint, and RRM + State-Constraint baselines, we optimize for the objective in Eq. \ref{eq:constrained-RL} when constructing a policy. To attain the best performance possible given only the proxy reward function and a reference policy, we pick the divergence measure that induces the highest performance under the ground-truth reward function. This methodology, via privileged access to the ground-truth reward function, ensures that the State-Constrained-PPO baseline induces the best performance achievable without updating the proxy reward function. Note that only privileged access to the ground-truth reward function is used to select the divergence measure parameters for the State-Constrained-PPO, Online-State-Constraint, and RRM + State-Constraint baselines; our approach never uses privileged information. 

For the Pandemic environment, we use the same reference policy as \cite{laidlaw2024correlated}, and therefore choose the divergence measure $F$ and constant $\beta$ from Eq. \ref{eq:constrained-RL} that induces the best performance in their paper: $F$ is the KL-divergence between state occupancy measures and $\beta = 0.06$. For all other environments, we perform a joint search over the choice of occupancy measure---either state or state-action occupancy---and the divergence measure and $\beta$ coefficient specified in Table~\ref{tab:env-divergence-coeffs}. For each combination, we evaluate performance using the ground-truth reward function and select the setting that yields the highest mean return across 3 random seeds. For the Traffic environment, $F$ is the KL-divergence between state-action occupancy measures and $\beta = 0.0005$; for the Glucose environment, $F$ is the KL-divergence between state-action occupancy measures and $\beta = 0.06$; for the AI safety gridworld environment, $F$ is the KL-divergence between state-action occupancy measures and $\beta = 0.08$. Note that our search did not include using action-occupancy measures due to their consistently inferior performance in \cite{laidlaw2024correlated}.

\begin{table}[h]
\centering
\caption{Coefficient values we searched over for each divergence measure, environment, and occupancy measure.}
\begin{tabular}{llccccccc}
\toprule
\textbf{Environment} & \textbf{Divergence} & \multicolumn{7}{c}{\textbf{Coefficient Values}} \\
\cmidrule(lr){3-9}
 & & 1 & 2 & 3 & 4 & 5 & 6 & 7 \\
\midrule
AI safety gridworld  & $\sqrt{\chi^2}$ & 0.0005 & 0.001  & 0.0025 & 0.005  & 0.01   & 0.025  & 0.05  \\
AI safety gridworld   & KL              & 0.8    & 0.4    & 0.16   & 0.08   & 0.04   & 0.016  & 0.008 \\
Traffic & $\sqrt{\chi^2}$ & 2e-6   & 4e-6   & 1e-5   & 2e-5   & 4e-5   & 1e-4   & 2e-4  \\
Traffic & KL              & 0.005  & 0.0025 & 0.001  & 0.0005 & 0.00025 & 0.0001 & 0.00005 \\
Glucose & $\sqrt{\chi^2}$ & 0.0005 & 0.001  & 0.0025 & 0.005  & 0.01   & 0.025  & 0.05  \\
Glucose & KL              & 0.0015 & 0.003  & 0.006  & 0.015  & 0.03   & 0.06   & 0.15  \\
\bottomrule
\end{tabular}
\label{tab:env-divergence-coeffs}
\end{table}

\subsection{Hyperparameters for PPO}
\label{sec:hyperparam-tuning}

We used the same hyperparameters for PPO as \cite{laidlaw2024correlated}, with two exceptions. To reduce the running time of our approach and all baselines that learn a reward function, which each require training at least one new policy per reward-learning iteration, we reduce the number of PPO training iterations and the PPO batch size. We ensure that with the reduced training batch size and PPO training iterations, optimizing for the proxy reward function still consistently induces reward hacking behavior, and optimizing for the ground-truth reward function still consistently induces the highest achievable expected return or is less than a standard deviation away. The reduced number of iterations and training batch sizes for each environment are shown in Table \ref{tab:ppo-training-hyperparams}.

\begin{table}[h]
\centering
\caption{PPO training hyperparameters for each environment that differ from the ones originally used by \cite{laidlaw2024correlated}.}
\begin{tabular}{lcc}
\toprule
\textbf{Environment} & \textbf{PPO Batch Size} & \textbf{Training Iterations} \\
\midrule
Pandemic & 3,840   & 100 \\
Glucose  & 10,000  & 150 \\
Traffic  & 80,000  & 100 \\
AI safety gridworld   & 1,000   & 100 \\
\bottomrule
\end{tabular}
\label{tab:ppo-training-hyperparams}
\end{table}

\subsection{Learning from preferences experimental details}

\textbf{How many preferences are elicited per iteration?~~~ }For each environment, we elicit $k^2$ preferences per iteration. If a method rolls out two policies $\pi_1$ and $\pi_2$ per iteration, e.g., PBRR samples trajectories from $\pi_\text{ref}$ and $\pi^*_{\hat{r}_t}$, then $k$ trajectories are sampled from each policy and preferences are elicited between all $k^2$ pairs $(\tau_1, \tau_2)$ where $\tau_1 \sim \pi_1$ and  $\tau_2 \sim \pi_2$. All baselines that roll out a single policy per iteration select $k^2$ trajectory pairs from a candidate batch via ensemble-based uncertainty estimates (see Appendix \ref{app:baselines} for details on how this candidate batch is constructed for each relevant method). $k = 19$ for the AI safety gridworld and traffic environments, $k=39$ for the glucose environment, and $k=79$ for the pandemic environment. These values were chosen to balance between distinguishing the sample efficiency of different methods and limiting the computational cost of training a new policy after every reward function update. 

\textbf{How are trajectories for preference elicitation constructed?~~~} For all environments except glucose, we elicit preferences over full trajectories of length $H$: $H=100$ for AI safety gridworld, $H=192$ for pandemic, and $H=300$ for traffic. In the glucose environment, episodes span $H=5760$ timesteps. If a simulated patient dies, the episode enters an absorbing state where all observations, actions, and rewards are zero until the horizon is reached. To avoid out-of-memory issues when learning from preferences, we split each glucose trajectory into three equal segments (each of length 1920), discarding any segments that consist entirely of absorbing-state transitions. Preferences are then elicited over the remaining segments.

\textbf{How are preferences labeled?~~~} We label all trajectory pairs with preference labels sampled from the Boltzman distribution given the ground-truth reward function. We use $\gamma=0.99$ when labeling preferences using $r$, and when learning $\hat{r}$ from preferences. This is the discount factor used by \cite{laidlaw2024correlated} when learning from $r$ for the same benchmark environments.

\subsection{PBRR $\lambda_1$ and $\lambda_2$}
\label{app:lambdas_decay_desc}
In Section \ref{sec:method} we outline how the proxy reward function is updated with PBRR via the preference-learning objective in Eq. \ref{eq:reg-pref-loss}. In practice, we set $\lambda_1 = \lambda_2 = 10$, and then divide both terms by the number of trajectory pairs collected where the proxy reward function's induced ranking agrees with the human preference label, $|\mathcal{D}^+|$. In effect, this decays $\lambda_1$ and $\lambda_2$ as the proxy reward function is repaired. Specifically, at iteration $t$, $\lambda_1^t = \lambda_2^t = \frac{10}{|\mathcal{D}^+|}$. 

\section{Online-RLHF Ablations}
\label{app:online_rlhf_abalations}
A line of prior work \citep{lee2021pebble, liang2022reward, metcalf2024sample, park2022surf} explored how to improve the data efficiency of RLHF methods across a suite of robotics control tasks. The techniques introduced in these works are complementary to our proposed approach, PBRR, and could in principle be combined with it. We leave such an exploration for future work. In this section, we examine whether the data-efficiency strategies proposed by \citet{lee2021pebble, liang2022reward, metcalf2024sample} substantially enhance the initial performance of the Online-RLHF baseline. We find that PBRR continues to outperform the initial performance achieved by Online-RLHF, even when these strategies are applied. We compare to the following methods:

\textbf{Online-RLHF~~~} The baseline is used for the main results in Section \ref{sec:results}.

\textbf{Online-RLHF + \citet{lee2021pebble}~~~}
Identical to the Online-RLHF baseline, except that, following the unsupervised pre-training procedure introduced by \citet{lee2021pebble}, we first train an exploration policy using their entropy-maximization objective. Trajectories collected from this exploration policy are then added to the candidate batch used to construct trajectory pairs for preference elicitation. As in Online-RLHF, the candidate batch also includes trajectories sampled from the reference policy.

\textbf{Online-RLHF + \citet{liang2022reward}~~~}
Identical to the Online-RLHF + \citet{lee2021pebble} baseline, except that, following the approach of \citet{liang2022reward}, the learned reward function is augmented with an exploration bonus term that encourages the policy to visit states where the reward estimate is uncertain. This exploration bonus is gradually decayed to $0$ once half the total number of preferences used in Figure~\ref{fig:main_results} have been collected.

\begin{figure}[!htbp]
    \centering
    \includegraphics[width=0.48\textwidth]{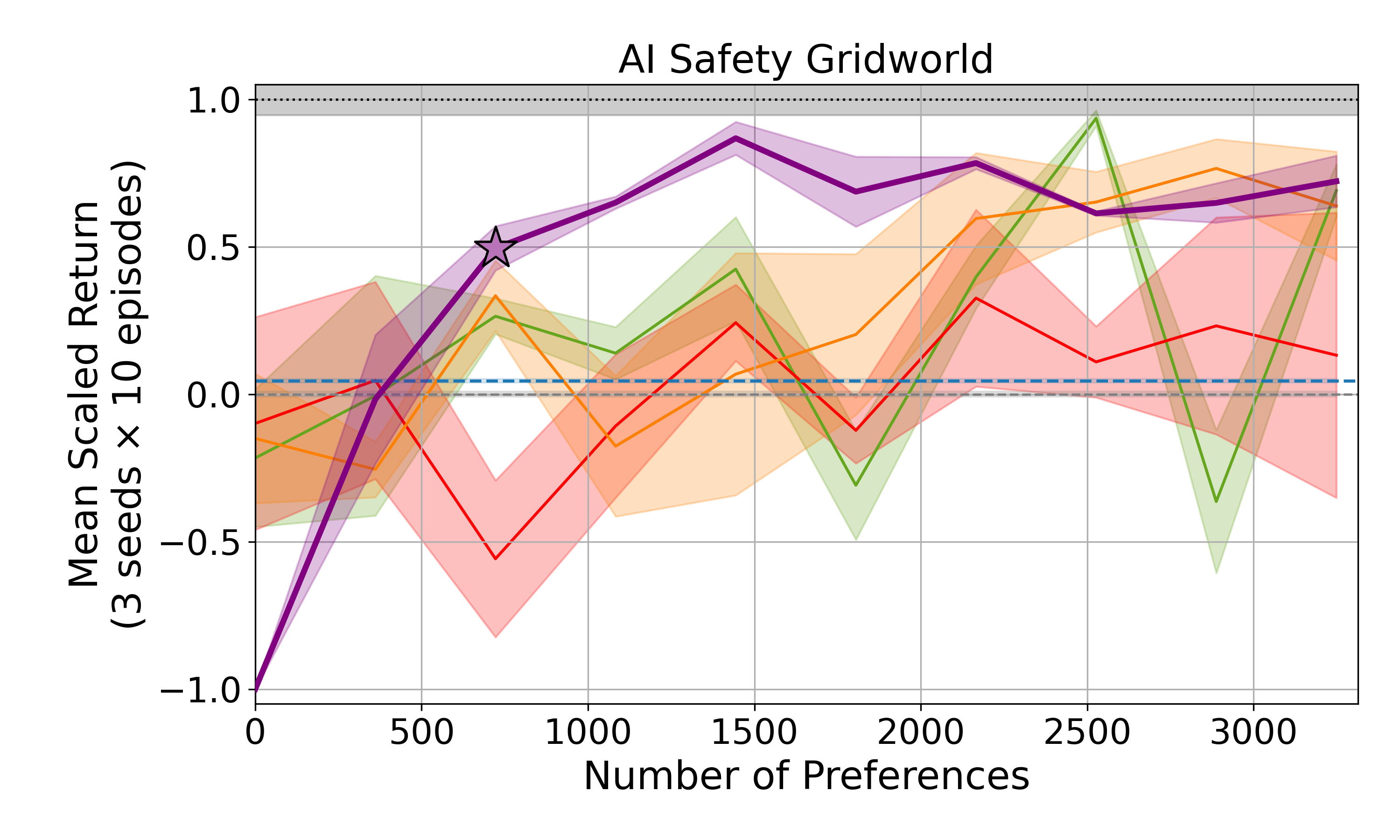}
    \includegraphics[width=0.48\textwidth]{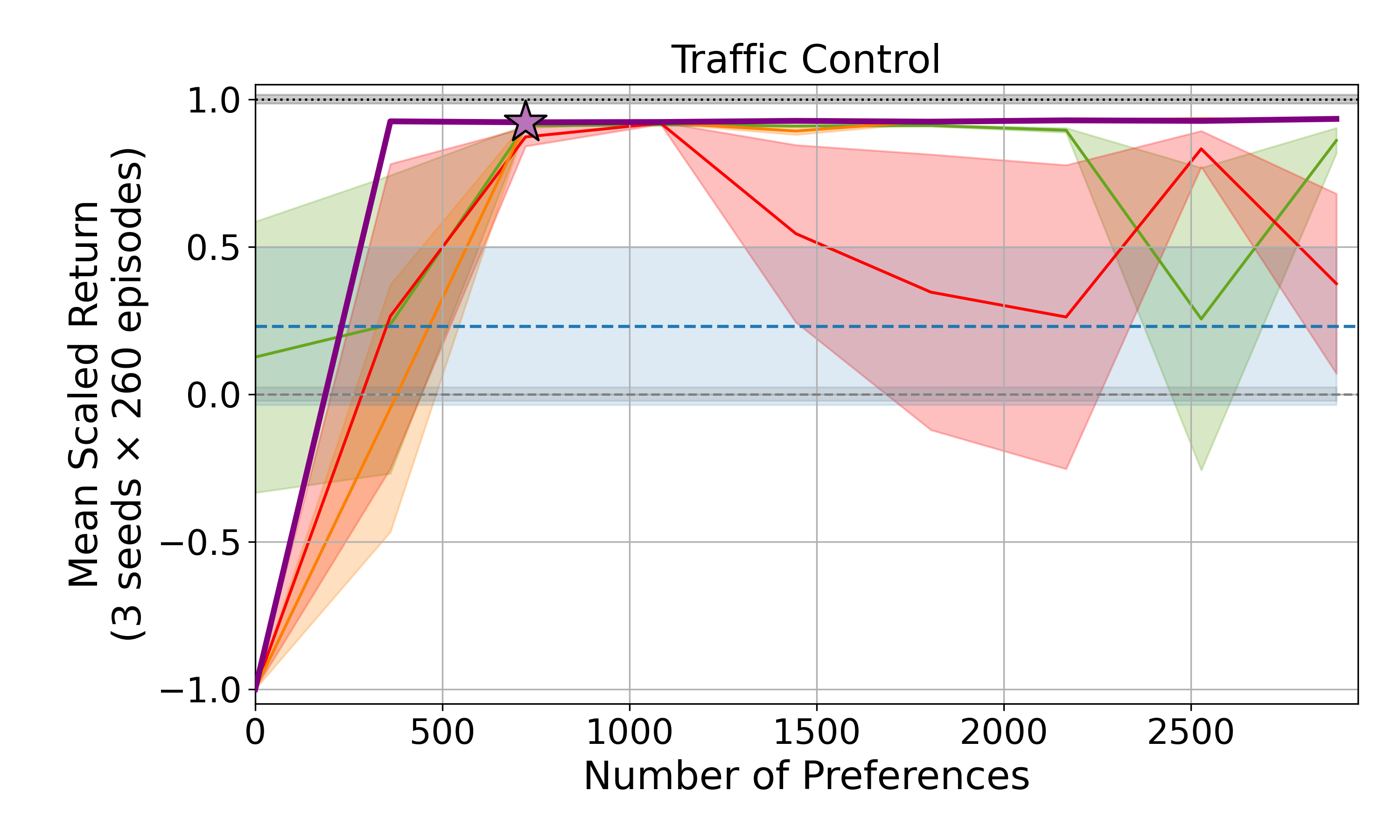}\\[0.5em]
    \includegraphics[width=0.48\textwidth]{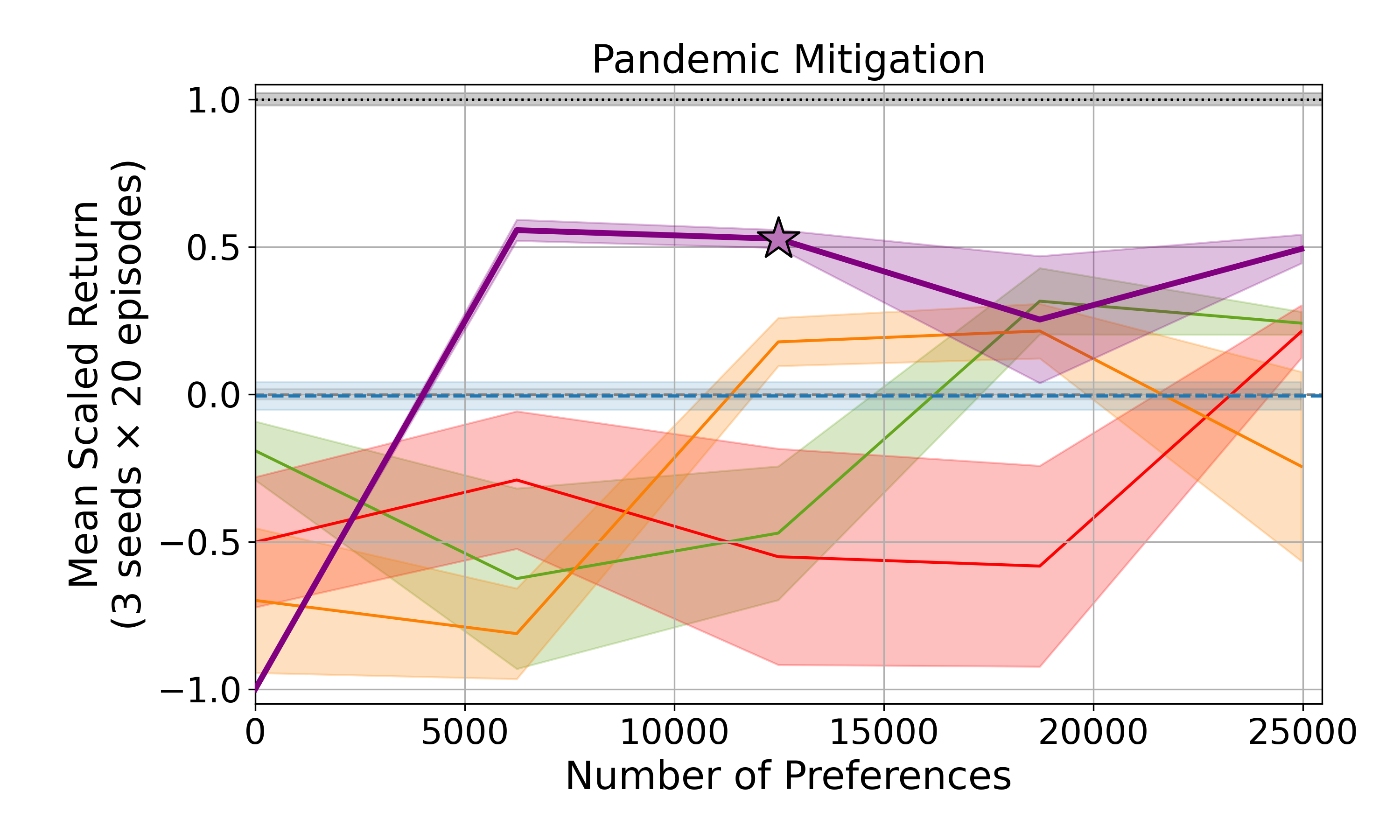}
    \includegraphics[width=0.48\textwidth]{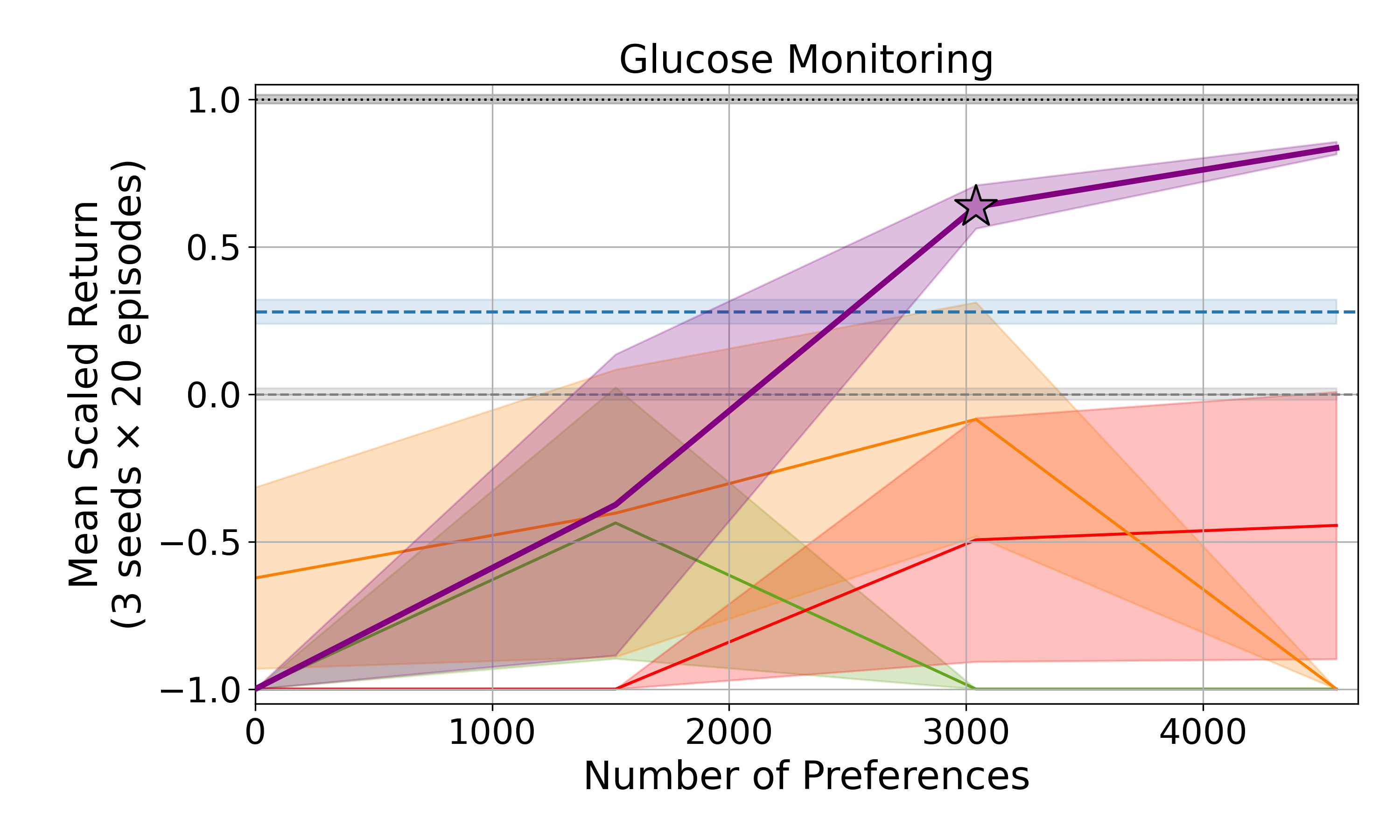}
    \includegraphics[width=0.85\textwidth]{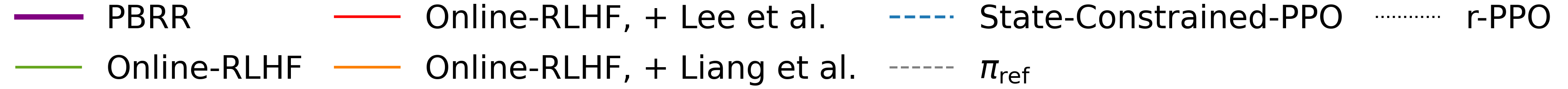}
    \caption{Mean return under the ground-truth reward function achieved by the Online-RLHF baseline and its variants that utilize proposed data-efficiency techniques from \citet{lee2021pebble} and \cite{liang2022reward}. See Figure \ref{fig:main_results} for plotting details.}
    \label{fig:online_rlhf_abaltion}
\end{figure}

As shown in Figure \ref{fig:online_rlhf_abaltion}, PBRR consistently outperforms the initial performance achieved by the Online-RLHF baselines, even when augmented with these data-efficiency techniques, for all environments except Traffic Control. In Traffic Control, PBRR matches the performance of Online-RLHF + \cite{liang2022reward} after eliciting a single batch of preferences. We hypothesize that exploration procedures designed to maximize state coverage (e.g., \cite{lee2021pebble}) or reward uncertainty (e.g., \citet{liang2022reward}) may be less effective in non-ergodic MDPs where most states correspond to undesirable outcomes. For instance, in the Glucose Monitoring environment, many treatment policies yield poor health outcomes for the patient, and exploring such regions of the state space---even if those regions induce the most uncertainty in predicted reward---may provide little information about what constitutes a good treatment policy.

We additionally compare PBRR against the following baseline in the AI safety gridworld:

\textbf{Online-RLHF + \citet{metcalf2024sample}~~~} Identical to the Online-RLHF + \citet{lee2021pebble} baseline, except that, following the approach of \citet{metcalf2024sample}, a self-supervised dynamics representation is jointly learned with the reward model parameters using their proposed training procedure.

\begin{figure}[!htbp]
    \centering
    \includegraphics[width=0.48\textwidth]{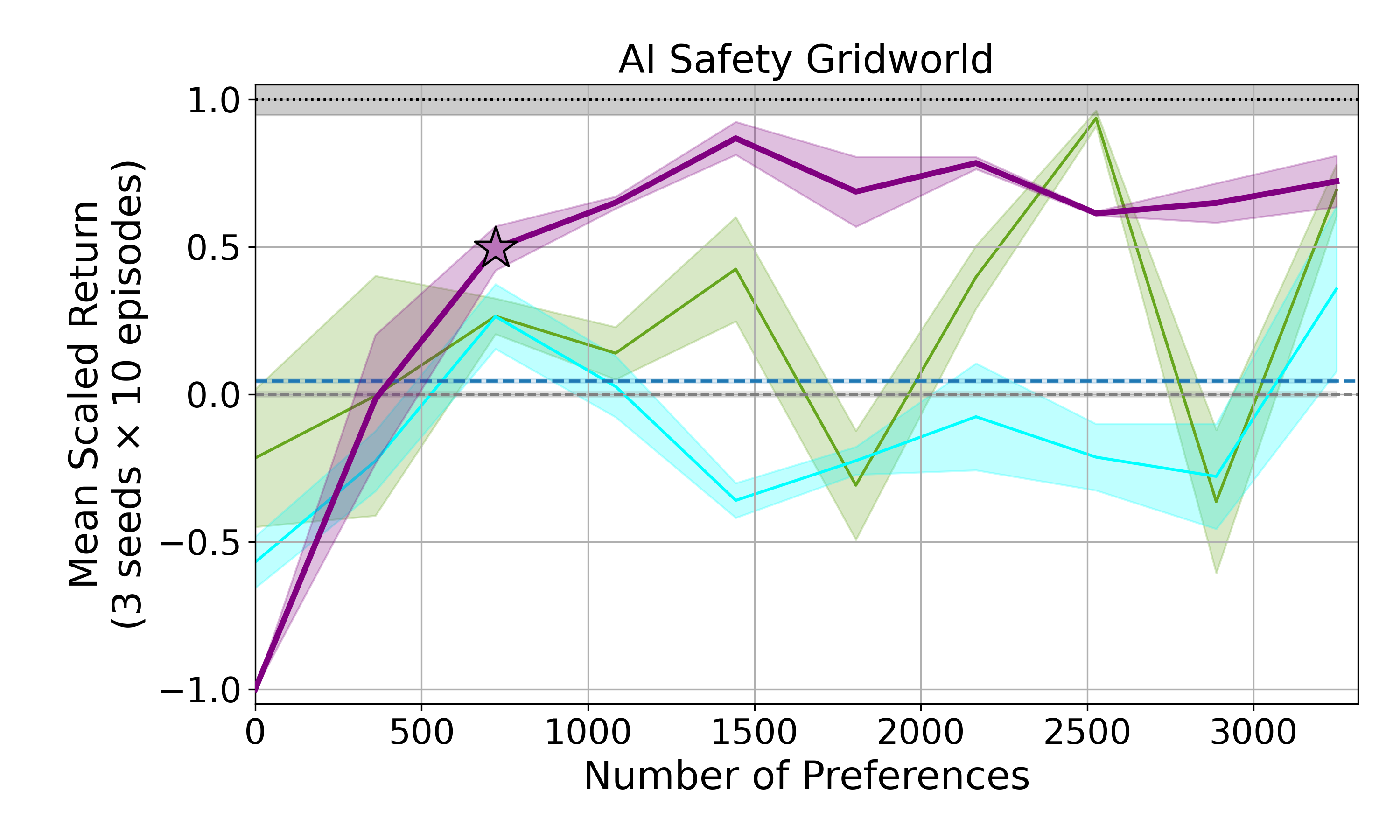}
    \includegraphics[width=0.85\textwidth]{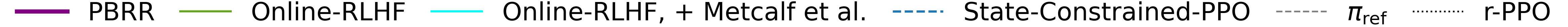}
    \caption{Mean return under the ground-truth reward function achieved by the Online-RLHF baseline and the Online-RLHF + \citet{metcalf2024sample}. See Figure \ref{fig:main_results} for plotting details.}
    \label{fig:online_rlhf_abaltion_2}
\end{figure}

The results in Figure \ref{fig:online_rlhf_abaltion_2} show that Online-RLHF + \citet{metcalf2024sample} underperforms PBRR and Online-RLHF; we suspect that the learned dynamics aware representation is not sufficient to learn a reward function over. 

\newpage
\section{AI Safety Gridworld Qualitative Analysis}
\label{app:safety_gw_qual_analysis}

\begin{wrapfigure}{R}{0.33\linewidth} 
    \centering
    \includegraphics[width=\linewidth]{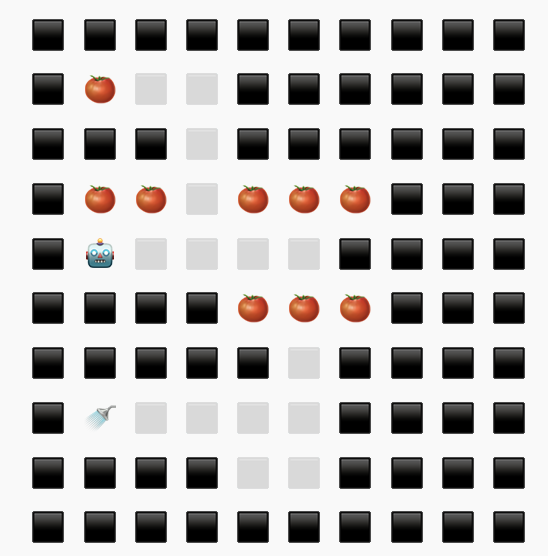}
    \caption{AI Safety Gridworld. The robot icon indicates the agent's initial start state, gray squares are passable states and black states are impassable. The tomatoes and sprinkler state are also illustrated. The robot only accrues reward under the ground-truth reward function upon its first visit to a tomato-containing state; an optimal policy visits all tomato-containing states. The robot additionally accrues reward under the proxy reward function when visiting the sprinkler state; an optimal policy for the proxy reward function moves to and stays in the sprinkler state.}
    \label{fig:aisafetygw}
    \vspace{-10mm}
\end{wrapfigure}

Figure \ref{fig:aisafetygw} shows the board layout we use for the AI safety gridworld environment. Here we qualitatively analyze why some approaches fail to learn a performant policy in this simple environment. While the other environments are less interpretable than AI safety gridworld, we suspect the methods we consider in this section may share poor performance in those environment for similar reasons.

\subsection{Why does Online-RLHF perform poorly?}
\label{app:safety_gw_rlhf_analysis}

The results in Figure \ref{fig:main_results} show that the Online-RLHF baseline exhibits oscillatory performance; its performance with respect to the ground-truth reward function decreases upon acquiring new preferences, only to increase again in subsequent iterations. This instability reflects the difficulty of reward learning in these reward-hacking benchmark environments. For instance, after acquiring 1,800 preferences in AI Safety Gridworld, the Online-RLHF baseline observes many trajectory pairs in which visiting more tomato-containing states is always preferred to visiting fewer. As a result, it learns a reward function that assigns positive reward whenever the agent enters a tomato-containing state. In contrast, the ground-truth reward function assigns positive reward only when a tomato is watered—i.e., upon its first visit. Consequently, the learned reward function induces a looping policy that remains in a single tomato-containing state, whereas the ground-truth reward function induces a policy that visits all tomato-containing states. This failure mode illustrates how RLHF methods can conflate instrumental goals (e.g., visiting a tomato-containing state) with terminal goals (e.g., visiting all tomato-containing states) even after learning from a large dataset of preferences; see \citet{marklund2025misalignment} for further characterization of this misalignment type. We suspect that the Online-RLHF baseline learns similarly misaligned reward functions in the other environments, while empirically PBRR does not exhibit this type of misalignment.

\subsection{Why does RRM perform poorly?}
\label{app:safety_gw_cao_analysis}

The results in Figure \ref{fig:main_results} show that the RRM performs poorly in this environment after collecting a substantial number of preferences. This result appears particularly surprising given that \cite{cao2025residual} follow the same methodology and achieve strong results on a suite of robotics tasks. Upon observing the policy that RRM learns after the first iteration, $\pi^*_{\hat{r}_0}$ , we note that it only aims to visit the sprinkler state to attain a high reward under the proxy reward function. Trajectories are then sampled from $\pi^*_{\hat{r}_0}$ and used to update the proxy reward function, inducing policy $\pi^*_{\hat{r}_1}$ at the next iteration. Trajectories sampled from $\pi^*_{\hat{r}_0}$ that visit more tomatoes---specifically the ones in coordinates $(5,4), (5,5), (5,6)$ that are on the way to the sprinkler---are always preferred to trajectories that visit less tomatoes. Therefore, the proxy reward function is updated to assign a higher reward for visiting those states. Consequently, the policy derived from the updated proxy reward function does not explore the other states---such as the tomato-containing states in the top half of the grid-world----because the states including and around $(5,4), (5,5), (5,6)$ are predicted to have higher reward. This process repeats, where the proxy reward function is only ever updated for a particular region of the grid-world---incorrectly assigned high reward---and therefore performance does not improve. More broadly, exploiting the proxy reward function does not necessarily induce an effective exploration policy---an observation also noted by \cite{xie2024exploratory}. We suspect \cite{cao2025residual} achieve strong performance in their robotics tasks as a result of learning from a proxy reward function that induces a policy that makes meaningful progress toward the true objective. This is notably not the case in the settings we consider, although the proxy reward function can still be updated with relatively few preferences to induce near-optimal performance via PBRR. 

\subsection{Why does PBRR with the objective in Eq. \ref{eq:loss} perform poorly?}
\label{app:safety_gw_pbrr_no_reg_analysis}

The results in Figure \ref{fig:abalation} show that, in this environment, PBRR with the preference-learning objective in Eq. \ref{eq:loss} substantially under-performs PBRR's default implementation, which uses the preference-learning objective in Eq. \ref{eq:reg-pref-loss}. To understand why, we note that $\pi_{\text{ref}}$ by construction only visits the tomato-containing states at coordinates $(1,1), (3,1), (3,2)$. Throughout training, PBRR observes preferences over trajectories sampled from $\pi_{\text{ref}}$ and $\pi^*_{\hat{r}_t}$. It is usually the case---inevitably during the early iterations---that trajectories from $\pi_{\text{ref}}$ are preferred over trajectories from $\pi^*_{\hat{r}_t}$. Without the regularization terms of Eq. \ref{eq:reg-pref-loss}, minimizing the cross-entropy loss from Eq. \ref{eq:loss} over the collected dataset of preferences then results in high predicted reward being assigned to the transitions encountered by $\pi_{\text{ref}}$. As a result, $\pi^*_{\hat{r}_t}$ begins to visit those transitions at subsequent iterations, but does not explore other parts of the state space such as the tomato-containing states on the right hand side of the grid-world. In effect, minimizing the standard preference loss induces unbounded updates to the proxy reward function, resulting in a policy that over-values the actions taken by $\pi_{\text{ref}}$ and does not explore other, potentially optimal actions. The regularization terms we add for Eq. \ref{eq:reg-pref-loss} encourage the updated proxy reward function to remain sufficiently optimistic by encouraging only decrements to the predicted reward when the proxy reward function induces a ranking that doesn't match the elicited preference. 

\section{Illustrative scenarios: minimal MDPs}
\label{app:example_mdps}
\begin{figure}[h]
    \centering
    \begin{subfigure}[t]{0.48\textwidth}
        \centering
        \includegraphics[width=\textwidth]{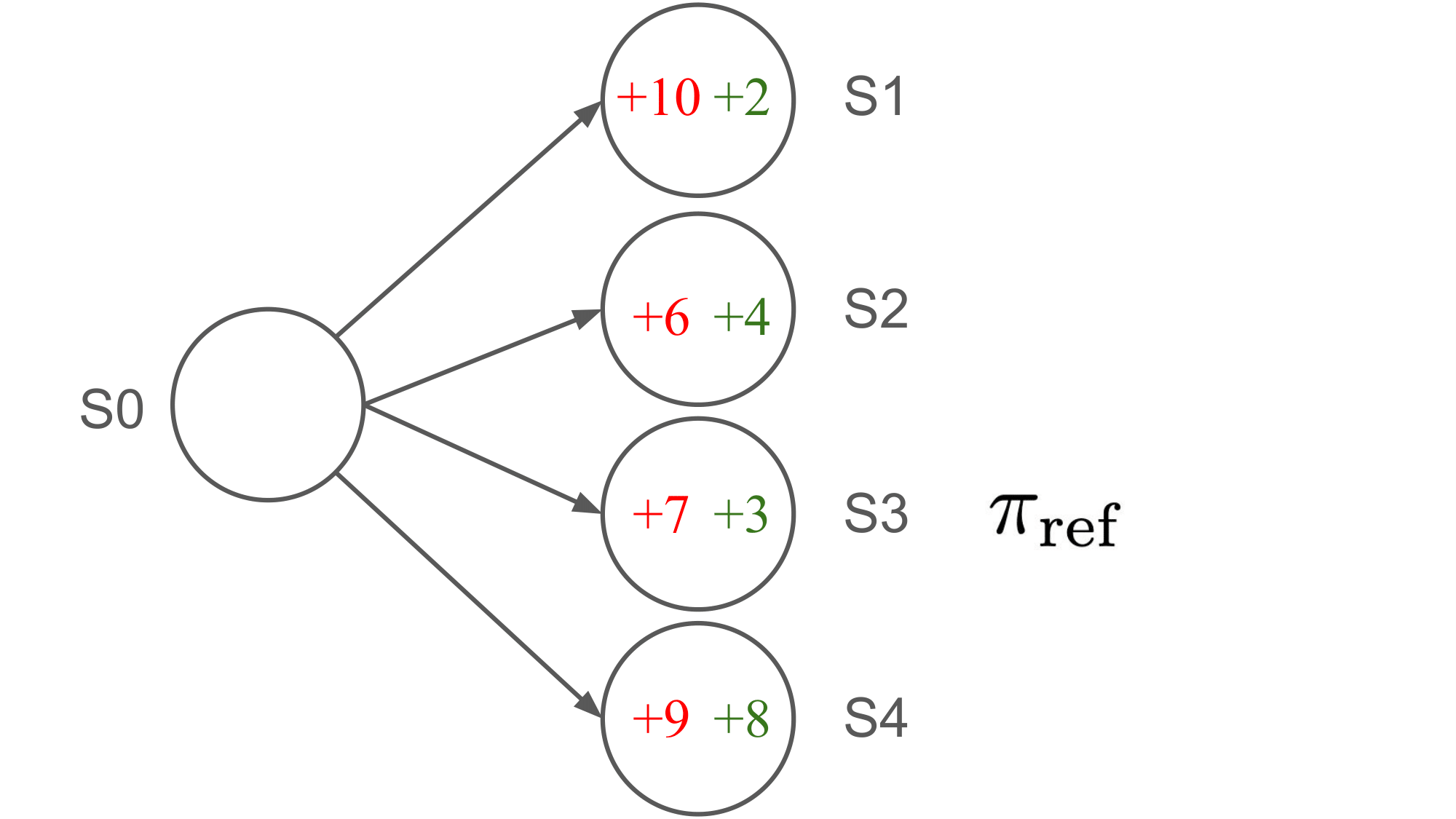}
        \subcaption{MDP 1}
    \end{subfigure}
    \hfill
    \begin{subfigure}[t]{0.48\textwidth}
        \centering
        \includegraphics[width=\textwidth]{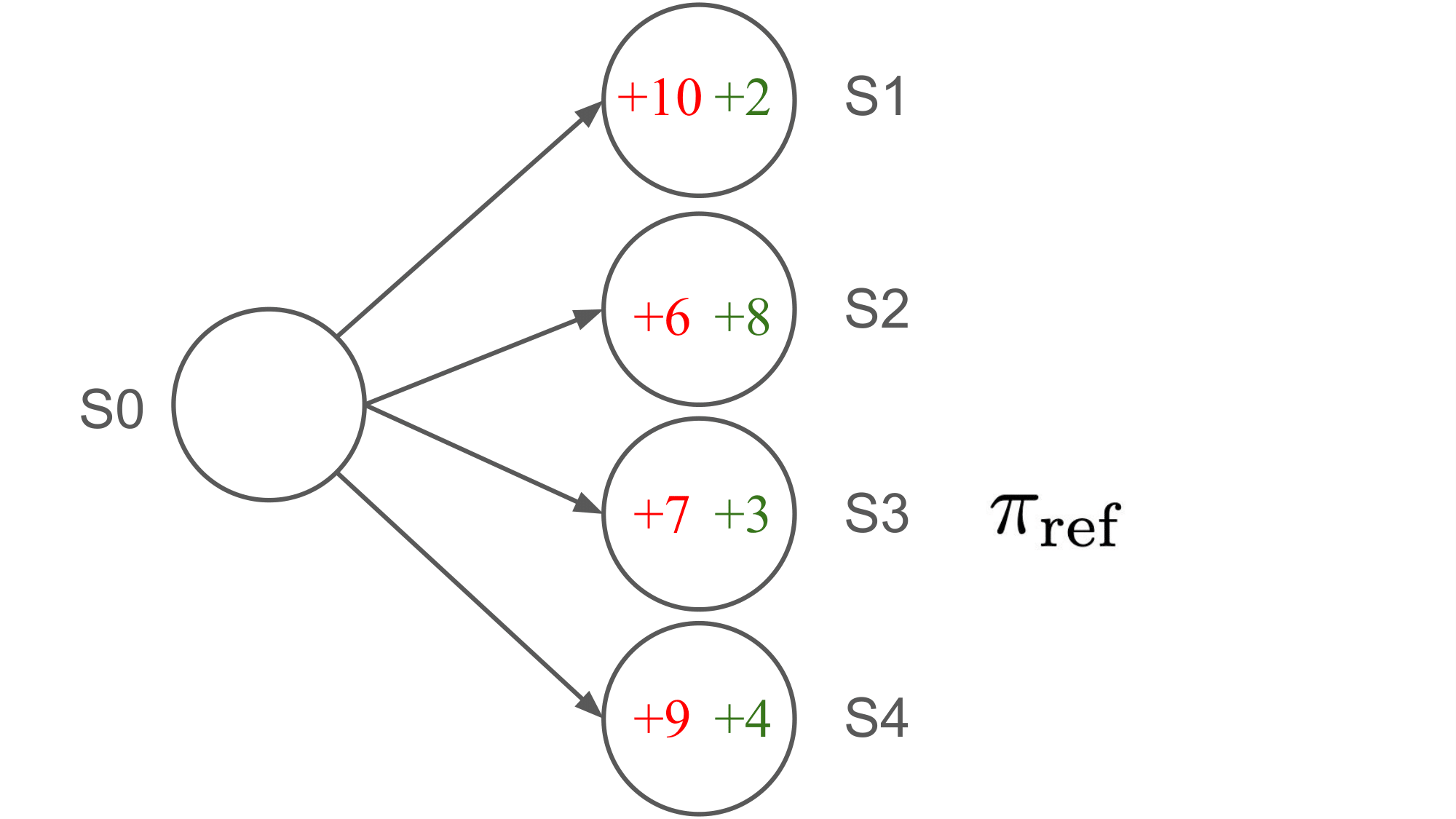}
        \subcaption{MDP 2}
    \end{subfigure}
    
    \caption{Example MDPs highlighting the strengths and weaknesses of PBRR, as outlined in Appendix~\ref{app:example_mdps}. For both MDPs, assume deterministic transition dynamics and $4$ available actions from state $s_0$. $\pi_{\text{ref}}$ marks the state that is visited by the supplied reference policy. The proxy reward function's outputted rewards are in red, and the ground-truth reward function's outputted rewards are in green. Rewards are only defined over states and the time horizon $H=1$.}
    \label{fig:example_mdps}
\end{figure}

In this section we present simple MDPs that highlight PBRR's strengths and weakness respectively. 

We first show that there exists MDPs where PBRR can learn an optimal policy much faster than random sampling: 

\begin{theorem}\label{thm:instance_better_than_unif}
There exists an MDP in which PBRR recovers an optimal policy after \(O(1)\) preference query, whereas a uniform-exploration baseline requires \(O(|S|)\) preferences in expectation.
\end{theorem}

\begin{proof}
\textbf{MDP and rewards.} Consider a horizon-\(H=1\) MDP with start state \(s_0\). From \(s_0\) there are \(n\) actions \(a_1,\dots,a_n\) leading deterministically to terminal states \(s_1,\dots,s_n\), respectively. Rewards are tabular over next-states: taking \(a_i\) yields immediate reward \(r(s_i)\).

Let the initial \emph{proxy} reward function satisfy
\[
\hat r_0(s_1)\;>\;\hat r_0(s_n)\;>\;\hat r_0(s_i)\quad\text{for all }i\notin\{1,n\}.
\]
Let the \emph{ground-truth} reward function satisfy
\[
r(s_n)\;>\;r(s_2)\;>\;r(s_i)\quad\text{for all }i\notin\{2,n\},
\]
so \(a_n\) is uniquely optimal. Assume a reference policy $\pi_{\text{ref}}$ is supplied that always chooses action $a_2$. Note that the reference action \(a_2\) is strictly better than \(a_1\) under \(r\). Assume noiseless preferences labeled by \(r\) over single-step transitions: \((s_0,a_i,s_i)\succ(s_0,a_k,s_k)\) iff \(r(s_i)>r(s_k)\). 


\textbf{PBRR needs one preference.} First, \(\pi^*_{\hat r_0}\) is constructed from the initial proxy reward function $\hat r_0$. \(\pi^*_{\hat r_0}\) chooses \(a_1\) (since \(\hat r_0(s_1)\) is maximal), while the supplied reference \(\pi_{\text{ref}}\) deterministically chooses \(a_2\). PBRR constructs the pair
\[
(s_0,a_1,s_1)\ \text{vs.}\ (s_0,a_2,s_2),
\]
which is labeled \((s_0,a_2,s_2)\succ(s_0,a_1,s_1)\) because \(r(s_2)>r(s_1)\) by construction. In accordance with Eq. \ref{eq:reg-pref-loss}, the proxy reward function is updated by decrementing \(\hat r(s_1)\) below \(\hat r(s_2)\) while leaving all uncompared transitions unchanged. This update produces \(\hat r_1\) with
\[
\hat r_1(s_2)>\hat r_1(s_1)\quad\text{and}\quad \hat r_1(s_i)=\hat r_0(s_i)\ \text{for all }i\notin\{1,2\}.
\]
Since initially \(\hat r_0(s_n)>\hat r_0(s_i)\) for all \(i\notin\{1,n\}\), and the update leaves \(\hat r(s_n)\) and \(\hat r(s_2)\) unchanged, we still have
\[
\hat r_1(s_n)>\hat r_1(s_i)\quad\text{for all }i\neq n.
\]
Hence \(\pi^*_{\hat r_1}\) selects \(a_n\), which is optimal under \(r\). Therefore PBRR reaches an \(r\)-optimal policy after a single preference, i.e., \(O(1)\).

\textbf{Uniform exploration needs \(O(|S|)\) preferences.} Consider a baseline that, at each iteration, selects two actions \(a_i\) and  \(a_k\) uniformly from \(\{a_1,\dots,a_n\}\) and elicits a preference comparing \((s_0,a_i,s_i)\) against \((s_0,a_k,s_k)\). Assuming the same update rule as PBRR above, the only comparison that can demote the action chosen by the policy induced by the proxy reward function \(a_1\) below \(a_i\) for $i \not = 1$ (and thereby expose \(a_n\) as the new argmax, given the update) is the pair involving \(i=1\). Each iteration hits \(i=1\) with probability \(2/n\), so the waiting time to the first such informative comparison is geometric with mean \(n/2\). Thus the baseline requires \(\mathbb{E}[\#\text{preferences}]=n/2=O(n)=O(|S|)\) (since \(|S|=n+1\)) before it can recover the optimal action \(a_n\).
\end{proof}

MDP 1 shows a specific illustration of one such MDP that satisfied Theorem \ref{thm:instance_better_than_unif}. We now step through the procedure executed by PBRR following Algorithm \ref{alg:exploration} in MDP 1. First the proxy reward function initially induces a policy that visits $s_1$. Upon observing a preference between this trajectory and the trajectory sampled from $\pi_{\text{ref}}$--- $s_3$ $\succ$ $s_1$--- the proxy reward function is updated so that the proxy reward for $s_1$ is lower than the proxy reward for $s_3$. Therefore, the updated proxy reward function at the next iteration induces a policy that visits $s_4$, which is optimal with respect to the ground-truth reward function. MDP 1 illustrates a scenario where the proxy reward function induces a substantially sub-optimal policy, but updating the proxy reward function at only a single state can repair it so as to induce an optimal policy under the ground-truth reward function. PBRR succeeds in such scenarios. 

We next present a scenario, shown in MDP 2 (Figure~\ref{fig:example_mdps}, right) where PBRR with  $C_1=0$ (no explicit additional exploration) can fail to learn the optimal policy. Here after the same initial step as in MDP, the proxy reward function is updated so that the induced policy visits $s_4$. The preference collected at this iteration will be $s_4$ $\succ$ $s_3$. Because the objective in Eq.~\ref{eq:reg-pref-loss} discourages updating the proxy reward function when it induces a ranking that matches the elicited preference, the proxy reward function is not updated further. But the proxy reward function does not induce an optimal policy with respect to the ground-truth reward function. MDP 2 highlights an example where PBRR with $C_1=0$ (no explicit additional exploration) will induce a policy that outperforms the reference policy but would not induce optimal performance.

\section{Regret Bound Proofs}
\label{sec:regret_bound_proofs}
\commentsh{@Emma I didn't want to change this below without your approval, but I think that $\pi_i$ should now be $\pi_t$ given that I changed Algorithm \ref{alg:exploration} so that $t$ indicates the iteration of PBRR (to match the regret bound on the order of $T$). Let me know if you agree/disagree.}
We now define some additional notation, following ~\cite{pacchiano_dueling}. Note when the dynamics model is known, one can compute the expected features $\phi$ under a policy $\pi_i$, which we denote as $\phi(\pi_i)$ 
\begin{eqnarray}
V_t &=&  \sum_{l=1}^{t-1} (\phi(\tau_l^1) - \phi(\tau_l^2))(\phi(\tau_l^1) - \phi(\tau_l^2))^T + \kappa \lambda I_d \\
g_t(w) &=& \sum_{l=1}^{t-1} \sigma(<\phi(\tau_l^1) - \phi(\tau_l^2),w>)(\phi(\tau_l^1) - \phi(\tau_l^2))+ \lambda W \\
w^L_t &=& \arg\min_{w s.t. ||w|| \leq W } || g_t(w) - g_t(\hat{w}^{MLE}_t) ||_{V^{-1}_t} \\
f_{mk}(\pi_1,\pi_2) &=& || \phi(\pi_1) - \phi(\pi_2) ||_{V^{-1}_t} \\
\alpha_{d,T}(\delta) &=& 20BW \sqrt{d \log \!\left(\frac{T(1+2T)}{\delta}\right)} \\
\beta_t(\delta) &=& \sqrt{ \lambda }  W + \sqrt{\log(1/\delta) + 2 d \log \left(1 + \frac{tB}{\kappa \lambda d} \right)} \\
C_t(\delta) &=& {w \; s.t.\; ||w - w^L_t||_{V_t} \leq 2 \kappa \beta_t(\delta)}\\
\gamma_t(\delta) & = & 2 \kappa \beta_t(\delta) + \alpha_{d,T}(\delta) \\
\Pi_t & = & \left\{ \pi_i | (\phi(\pi_i) - \phi(\pi))^T w_t + \gamma_t(\delta) || \phi(\pi^i) - \phi(\pi) ||_{V^{-1}_t} \geq 0 \; \forall \; \pi \right\} \\ 
\end{eqnarray}
When cross entropy loss is used to fit preference data and the reward model is linear, the resulting loss can be expressed as 
\begin{equation}
\mathcal{L}^{\lambda}_t(w) = \sum_{l=1}^{t} (o_l \log (\sigma( \langle \theta(\tau^1_l) - \theta(\tau^2_l)),w \rangle )) - \frac{\lambda}{2} ||w||^2_2 +  (1-o_l) \log (\sigma( 1- \langle \theta(\tau^1_l) - \theta(\tau^2_l)),w \rangle )),
\end{equation}
where $o_l=1$ or $0$ depending on which trajectory is preferred.

As the maximum likelihood estimator $w_{MLE}$ may not satisfy the required boundness assumption, prior work~\cite{pacchiano_dueling} defined a projected version $w^L_t$  of the weight vector $w$. 

We also define the event that the true $w^*$ iles in the specified confidence interval $C_t(\delta)$ on all time steps as 
\begin{equation}
\mathcal{E}_\delta = \{ \forall t \geq 1, \mathbf{w}_\star \in \mathcal{C}_t(\delta) \}.
\end{equation}

While in the main text we show the big-O version to avoid the additional notation complexity required to define the terms, we now state a more precise version of Theorem~\ref{thm:known_trans_model_regret_bound}:
\begin{theorem}
Let $\delta \leq 1/\mathrm{e}$ and $\lambda \geq B/\kappa$. Then under Assumptions \ref{asm:linear} through \ref{asm:bounded_features}, for $f=f_{mk}$ and $\Pi_t = \Pi_{t,mk}$, with probability at least $1-\delta$, the expected regret of Algorithm \ref{alg:exploration} is bounded by 
\begin{equation}
Regret_t \leq C_1 (4 \kappa \beta_T(\delta) + 2 \alpha_{d,T}(\delta)) \sqrt{ 2 T d \log \left(1 + \frac{TB}{\kappa d} \right)}
\end{equation}
\end{theorem}
\begin{proof}
Our proof closely follows the proof of Theorem 1 in \citet{pacchiano_dueling}. While their proof was for a different algorithm, we note that their Lemma 7, Corollary 1, Lemma 2 and Lemma 8 all continue to hold in our setting, as they do not depend on the specific policies chosen for exploration. 

The first part of the proof of our theorem exactly follows the proof of Theorem 1 \citep{pacchiano_dueling}, where conditioned on the event $\mathcal{E}_\delta$ holding, they bound the regret due to executing the two exploration policies $\pi_1$ and $\pi_2$:
\begin{eqnarray}
2r_t &=& (\phi(\pi^*) - \phi(\pi_t^1))^\top \mathbf{w}^* + (\phi(\pi^*) - \phi(\pi_t^2))^\top \mathbf{w}^* \nonumber \\
&=& (\phi(\pi^*) - \phi(\pi_t^1))^\top \mathbf{w}_t^L + (\phi(\pi^*) - \phi(\pi_t^1))^\top (\mathbf{w}^* - \mathbf{w}_t^L) 
+ (\phi(\pi^*) - \phi(\pi_t^2))^\top (\mathbf{w}^* - \mathbf{w}_t^L) 
+ (\phi(\pi^*) - \phi(\pi_t^2))^\top \mathbf{w}_t^L  \nonumber \\
&\leq& (\phi(\pi^*) - \phi(\pi_t^1))^\top \mathbf{w}_t^L + (\phi(\pi^*) - \phi(\pi_t^2))^\top \mathbf{w}_t^L \nonumber \\
& &\quad + \|\mathbf{w}^* - \mathbf{w}_t^L\|_{V_t} \cdot \|\phi(\pi^*) - \phi(\pi_t^1)\|_{V_t^{-1}} 
+ \|\mathbf{w}^* - \mathbf{w}_t^L\|_{V_t} \cdot \|\phi(\pi^*) - \phi(\pi_t^2)\|_{V_t^{-1}} \label{eqn:4terms}
\end{eqnarray}
They then note that in this sum, the last two terms can be bounded by their Corollary 1:
\begin{eqnarray}
\|\mathbf{w}^* - \mathbf{w}_t^L\|_{V_t} \cdot \|\phi(\pi^*) - \phi(\pi_t^1)\|_{V_t^{-1}} 
+ \|\mathbf{w}^* - \mathbf{w}_t^L\|_{V_t} \cdot \|\phi(\pi^*) - \phi(\pi_t^2)\|_{V_t^{-1}} \\
\leq (2\kappa \beta_t(\delta) + \alpha_{T,d}(\delta)) \cdot \Big( \|\phi(\pi^*) - \phi(\pi_t^1)\|_{V_t^{-1}} 
+ \|\phi(\pi^*) - \phi(\pi_t^2)\|_{V_t^{-1}} \Big) \label{eqn:last2}
\end{eqnarray}
Up to this point, the proof is identical. We now seek to bound the first two terms in Equation~\ref{eqn:4terms}. First note that  
\begin{equation}
(\phi(\pi^*) - \phi(\pi_t^1))^\top \mathbf{w}_t^L + (\phi(\pi^*) - \phi(\pi_t^2))^\top \mathbf{w}_t^L 
\leq (2 \kappa \beta_t(\delta) + \alpha_{T,d}(\delta)) \left(|| \phi(\pi^*) - \phi(\pi^1_t)||_{\bar{V}_t^{-1}} + || \phi(\pi^*) - \phi(\pi^2_t)||_{\bar{V}_t^{-1}} \right) \label{eqn:wd}
\end{equation} 
since Line 9 in Algorithm~\ref{alg:exploration} ensures that the selected $\pi_1$ and $\pi_2$ always lie in $\Pi_t$, and therefore holds from the definition of $\Pi_t$.  

In addition, Line 9 in Algorithm~\ref{alg:exploration} ensures if $\pi^*_{\hat{r}_t}=\pi_1$ and $\pi_{\mathrm{ref}}=\pi_2$ are used as the exploration policies, then they must satisfy:
\begin{equation}
\max_{\pi_i,\pi_j \in \Pi_t} f(\pi_i,\pi_j) \leq C_1 f(\pi^*_{\hat{r}_t},\pi_{\mathrm{ref}}).
\end{equation}
Therefore     
\begin{equation}
\left\| \phi(\pi^\star) - \phi(\pi_{\text{ref}}) \right\|_{\bar V_t}^{-1} 
+ 
\left\| \phi(\pi^\star) - \phi(\pi_{\text{proxy}}) \right\|_{\bar V_t}^{-1}
\;\;\le\;\;
2 C_1 \cdot 
\left\| \phi(\pi_1) - \phi(\pi_2) \right\|_{\bar V_t}^{-1}. \label{eqn:c1bound}
\end{equation}
The remaining part of the proof follows Theorem 1~\cite{pacchiano_dueling}. Specifically substituting Equations~\ref{eqn:last2}, ~\ref{eqn:wd}  and~\ref{eqn:c1bound} into Equation~\ref{eqn:4terms}, and using that $\pi^* \in \Pi_t$ under the assumed event and Lemma 2, yields
\begin{eqnarray}
2r_t \leq 2 \big( 2\kappa \beta_t(\delta) + \alpha_{T,d}(\delta) \big) \cdot 
\Big( \|\phi(\pi^*) - \phi(\pi_t^1)\|_{V_t^{-1}} + \|\phi(\pi^*) - \phi(\pi_t^2)\|_{V_t^{-1}} \Big) \\
\leq 4 C_1 \big( 2\kappa \beta_t(\delta) + \alpha_{T,d}(\delta) \big) \cdot 
\|\phi(\pi_t^1) - \phi(\pi_t^2)\|_{V_t^{-1}}
\end{eqnarray}
The remaining few steps in the proof of Theorem 1 then bound  

\begin{eqnarray}
Regret_T &=& \sum_{t=1}^T r_t \\
&\leq &
2 C_1 \big( 2\kappa \beta_T(\delta) + \alpha_{T,d}(\delta) \big) \cdot \sum_{t=1}^T
\|\phi(\pi_t^1) - \phi(\pi_t^2)\|_{V_t^{-1}} \\
&\leq &
2 C_1 \big( 2\kappa \beta_T(\delta) + \alpha_{T,d}(\delta) \big) \cdot \sqrt{T \sum_{t=1}^T
\|\phi(\pi_t^1) - \phi(\pi_t^2)\|^2_{V_t^{-1}} } \\
&\leq &
2 C_1 \big( 2\kappa \beta_T(\delta) + \alpha_{T,d}(\delta) \big) \cdot \sqrt{2 T d \log \Big(1 + \frac{TB}{\kappa d} \Big)},
\end{eqnarray}
using Cauchy-Schwarz for the second inequality and Lemma 8~\cite{pacchiano_dueling} for the final inequality. 
\end{proof}


We now provide an analogous proof for the case when the dynamics model is not known. We need some additional notation. Let $N_t(s,a)$ represent the number of times the trajectories have included $(s,a)$ tuples. The proof again follows prior work~\cite{pacchiano_dueling}. They define an alternate covariance matrix that leverages the empirical covariance matrix, and an alternate bonus term and confidence sets needed to account for the uncertainty since the dynamics model is estimated from finite samples.

\begin{equation}
\tilde{\mathbf{V}}_t = \kappa \lambda I_d 
+ \sum_{\ell=1}^{t-1} 
\left( \phi^{\hat{\mathbb{P}}_\ell}(\pi^1_\ell) - \phi^{\hat{\mathbb{P}}_\ell}(\pi^2_\ell) \right)
\left( \phi^{\hat{\mathbb{P}}_\ell}(\pi^1_\ell) - \phi^{\hat{\mathbb{P}}_\ell}(\pi^2_\ell) \right)^\top
\end{equation}
\begin{equation}
\xi^{(t)}_{s,a}(\eta, \delta) 
= \min \left( 2\eta,\; 4\eta \sqrt{\frac{U}{N_t(s,a)}} \right),
\end{equation}
where
\begin{equation}
U = H \log\big(|\mathcal{S}||\mathcal{A}|\big) 
+ \log\!\left(\frac{6 \log(N_t(s,a))}{\delta}\right).
\end{equation} The bonus function is:
\begin{eqnarray}
\hat{B}_t(\pi, \eta, \delta) 
&=& \mathbb{E}_{s_1 \sim \rho, \, \tau \sim \hat{\mathbb{P}}_t^{\pi}(\cdot \mid s_1)} 
\left[ \sum_{h=1}^{H-1} \xi_{s_h, a_h}^{(t)}(\eta, \delta) \right] 
\end{eqnarray}
and the undominated policy set is:
\begin{eqnarray}
\Pi_{t,\mu} =  
& \Bigg\{\pi^i \;\Bigg|\;
\big( \phi^{\hat{\mathbb{P}}_t}(\pi^i) - \phi^{\hat{\mathbb{P}}_t}(\pi) \big)^\top 
\mathbf{w}^L_t 
+ \gamma_t \big\| \phi^{\hat{\mathbb{P}}_t}(\pi^i) - \phi^{\hat{\mathbb{P}}_t}(\pi) \big\|_{\tilde{\mathbf{V}}_{t-1}} \\
&\quad + \widehat{B}_t\!\left(\pi^i, 2SB, \tfrac{\delta}{2|\mathcal{A}||\mathcal{S}|}\right) 
+ \widehat{B}_t\!\left(\pi, 2SB, \tfrac{\delta}{2|\mathcal{A}||\mathcal{S}|}\right) 
\geq 0, \; \forall \pi 
\Bigg\}.
\end{eqnarray}
and we define $f$ as:
\begin{eqnarray}
f_{u}(\pi_t^1, \pi_t^2) 
=  
\gamma_t \big\| \phi_t^{\hat{\mathbb{P}}_t}(\pi^1) - \phi_t^{\hat{\mathbb{P}}_t}(\pi^2) \big\| 
\|_{\tilde{\mathbf{V}}_{t-1}}
+ 2\widehat{B}_t(\pi^1, 2WB, \delta) 
+ 2\widehat{B}_t(\pi^2, 2WB, \delta).
\end{eqnarray}

We now restate our Theorem~\ref{thm:unknown_trans_model_regret_bound}:
\begin{theorem} (Theorem~\ref{thm:unknown_trans_model_regret_bound})
 Under assumptions \ref{asm:linear} through \ref{asm:bounded_features},  for $f=f_{u}$ and $\Pi_t = \Pi_{t,u}$, the regret of Algorithm \ref{alg:exploration} is bounded by 
\begin{equation}
R_T \leq \tilde{\mathcal{O}}\!\left( C_1 \left( 
\kappa d \sqrt{T} + H^{3/2} \sqrt{|\mathcal{S}||\mathcal{A}| d T H} 
+ H |\mathcal{S}| \sqrt{|\mathcal{A}| d T H} 
\right) \right),
\end{equation}
\end{theorem}

\begin{proof} (sketch)
The proof follows the proof of Theorem~2 of \cite{pacchiano_dueling} with the analogous modification as the one we made in our proof of Theorem 1. Specifically, since $\pi^*, \pi_t^1, \pi_t^2 \in \Pi_{t,u}$, note that 
\begin{equation}
f_u(\pi^*, \pi_t^1) + f_u(\pi^*, \pi_t^2) \leq 2 \max_{\pi_i, \pi_j \in \Pi_{t,u}} f_u(\pi_i, \pi_j)
\end{equation}
and from Lines 8--9 (with $C_1 \geq 1$), the definition of $f_{u}$ ensures:
\begin{equation}
\max_{\pi_i, \pi_j \in \Pi_{t,u}} f_u(\pi_i, \pi_j) \leq C_1 f_u(\pi_t^1, \pi_t^2).
\end{equation}
If the condition in Line~8 does not hold, Algorithm~1 keeps $(\pi_t^1, \pi_t^2) = (\pi^*_{\hat{r}_t}, \pi_{\text{ref}})$, and the inequality is the negation of that condition. Otherwise, Line~9 sets $(\pi_t^1, \pi_t^2)$ to the maximizing pair, and the inequality holds since $C_1 \geq 1$. The remainder of the proof follows the rest of the proof of Theorem~2 of \cite{pacchiano_dueling}, which bounds the instantaneous regret by $f_u(\pi^*, \pi_t^1) + f_u(\pi^*, \pi_t^2)$ up to a constant, and then bounds $\sum_t f_u(\pi_t^1, \pi_t^2)$; the latter step holds for any sequence of executed policy pairs.

\end{proof}


\section{Reward Hacking Analysis Proofs}
\label{app:theory}

Here we provide a theoretical analysis motivating PBRR, relying on different assumptions than the regret analysis in Section \ref{sec:regret_analysis}. We show that PBRR's repaired reward function is guaranteed to induce an optimal policy that matches or exceeds the performance of the reference policy (Thm.~\ref{thm:pbrr-performance}) and resolve two specific instantiations of reward hacking (Cors.~\ref{corr:no-skalse-hacking}, \ref{corr:no-laidlaw-hacking}) in the infinite data setting, at any iteration at which the induced policy does not change.

\subsection{Assumptions}
\label{app:regret_pref_theory_just}

For this theoretical analysis we assume all of the assumptions listed in Section \ref{sec:reward_hacking_analysis}. Below we expand on the assumption that preferences are labeled by regret.

We define regret under $\tilde r$ for a trajectory $\tau=(s_0,a_0,\ldots,s_{H})$ as $\mathrm{regret}(\tau\mid \tilde r)\;\triangleq\;\sum_{t=0}^{|\tau|-1}\gamma^t\!\left[V^*_{\tilde r}(s_t)-Q^*_{\tilde r}(s_t,a_t)\right]
\;=\; \sum_{t=0}^{|\tau|-1}\!\left(-\gamma^t A^*_{\tilde r}(s_t,a_t)\right)$ where $A^*_{\tilde r}(s,a)\!\triangleq\!Q^*_{\tilde r}(s,a)-V^*_{\tilde r}(s)$ and $V^*_{\tilde r},Q^*_{\tilde r}$ are the optimal value and action-value functions under $\tilde r$. We then assume a preference between $\tau_1$ and $\tau_2$ is determined by regret: 

\begin{equation}
    \mu \triangleq
\begin{cases}
0 & \text{if } \mathrm{regret}(\tau_1|r) < \mathrm{regret}(\tau_2|r),\\
1 & \text{if } \mathrm{regret}(\tau_1|r) > \mathrm{regret}(\tau_2|r),\\
\frac12 & \text{if } \mathrm{regret}(\tau_1|r) = \mathrm{regret}(\tau_2|r),
\end{cases}
\label{eq:det-regret-pref}
\end{equation}

where regret is the discounted sum of negative optimal advantages:

\begin{align*}
\mathrm{regret}(\tau \mid \tilde{r}) \triangleq 
\sum_{t=0}^{|\tau| - 1} 
-\gamma^t A^*_{\tilde{r}}(s_t^\tau, a_t^\tau).
\end{align*}

This quantity is measured against an \emph{optimal} policy for $\tilde r$, following the terminology of \citet{knox2022models}. It differs from the notion of regret used in much of the recent preference-learning literature, which is measured against the value of the behavior policy; \citet{cho2025policy} refer to the quantity above as the optimal advantage to make this distinction explicit, and \citet{park2026posttraining} and \citet{kim2026regret} argue that regret measured against the behavior policy can better represent human judgment. We use regret in the former sense throughout this analysis.

We also assume $\mathcal{L}_\text{pref}$ uses regret-based preferences. That is, $P(\tau_1 \succ \tau_2 \mid \tilde r) = \sigma\big(\mathrm{regret}(\tau_2 \mid \tilde r) - \mathrm{regret}(\tau_1 \mid \tilde r)\big)$,
which is the preference model referred to in Assumption~\ref{asm1:consistent}.

\citet{knox2022models} shows that human preferences are better described by the difference in regret between trajectory segments, as opposed to the difference in the sum of rewards. In practice, we simulate human preference labels using the difference in sum of rewards between trajectories because regret is intractable to compute in our empirical environments, as discussed in Appendix \ref{app:traj_pref_justification}.

Our assumption about how preferences are labeled (Eq. \ref{eq:det-regret-pref}) for this analysis differs in two key ways from \citet{knox2022models}: (i) we assume preferences are over trajectories, not trajectory segments (ii) we assume preferences are noiselessly determined by regret, not sampled from a Boltzmann distribution. Our analysis holds without assumption (i) as long as preferences are elicited over an exhaustive set of trajectory segments. We argue that (ii) is a reasonable assumption when executing PBRR, where preferences are only collected between trajectories sampled from $\pi_{\hat{r}_t}$---which is initially substantially sub-optimal due to a misspecified proxy reward function $\hat{r}_t$, and $\pi_{\text{ref}}$---which is assumed to be a safe policy. There is clear distinction between the trajectories from these policies---both qualitatively and in terms of regret under $r$, so we assume no preference noise via  Eq. \ref{eq:det-regret-pref}. 

We also assume that all trajectories begin from the same start state $s_0$. This assumption is without loss of generality: for any MDP with initial state distribution $p_0$, we can construct an equivalent MDP by introducing a new start state $s'_0$ that transitions in a single step according to $p_0$. In this construction, all trajectories then originate from $s'_0$, satisfying the assumption of a single start state.

\subsection{Proof of Theorem \ref{thm:pbrr-performance}}
\label{app:thm:pbrr-performance}

\begin{lemma}
\label{lem:ranking-agreement}
Suppose Assumptions~\ref{asm1:regret_prefs} and \ref{asm1:consistent} hold.
Then the repaired reward function $\hat r$ and the ground-truth reward function $r$
induce identical regret-based orderings on every trajectory pair in the preference
dataset: for all $(\tau_1, \tau_2, \mu) \in \D$,
\[
\mathrm{sign}\left[\mathrm{regret}(\tau_1 \mid \hat r) - \mathrm{regret}(\tau_2 \mid \hat r)\right]
=
\mathrm{sign}\left[\mathrm{regret}(\tau_1 \mid r) - \mathrm{regret}(\tau_2 \mid r)\right].
\]
\end{lemma}

\begin{proof}
Fix $(\tau_1,\tau_2,\mu)\in\D$. Under the regret preference model of Appendix~J.1,
$P(\tau_1 \succ \tau_2 \mid \hat r) = \sigma\big(\mathrm{regret}(\tau_2 \mid \hat r) - \mathrm{regret}(\tau_1 \mid \hat r)\big)$,
and $\sigma(x) > \tfrac12 \iff x > 0$, so Assumption~\ref{asm1:consistent} gives
\[
\mathrm{sign}\left[\mathrm{regret}(\tau_2 \mid \hat r) - \mathrm{regret}(\tau_1 \mid \hat r)\right]
= \mathrm{sign}\left(\tfrac12 - \mu\right).
\]
By Assumption~\ref{asm1:regret_prefs}, $\mu$ is determined by Eq.~\ref{eq:det-regret-pref}, so
\[
\mathrm{sign}\left(\tfrac12 - \mu\right)
= \mathrm{sign}\left[\mathrm{regret}(\tau_2 \mid r) - \mathrm{regret}(\tau_1 \mid r)\right].
\]
Equating the two right-hand sides and negating both arguments gives the claim.
\end{proof}


\paragraph{Restatement of Theorem~\ref{thm:pbrr-performance}.}

Suppose Assumptions~\ref{asm1:vanish} through \ref{asm1:consistent} hold. Then at any iteration
$t$ of Algorithm~\ref{alg:exploration} at which $\pi^*_{\hat r_t} = \pi^*_{\hat r_{t-1}}$, the optimal policy
for the repaired reward function matches or outperforms the reference policy:
\[
J_r(\pi^*_{\hat r_t}) \ge J_r(\pi_{\mathrm{ref}}).
\]
In particular, this holds for all $t > t_0$ if $\pi^*_{\hat r_t}$ is constant for $t \ge t_0$.


\begin{proof}

Fix an iteration $t$ with $\pi^*_{\hat r_t} = \pi^*_{\hat r_{t-1}}$ and write $\hat r = \hat r_t$.
At iteration $t-1$, Algorithm~\ref{alg:exploration} compared $\pi^*_{\hat r_{t-1}} = \pi^*_{\hat r_t}$
against $\pi_{\mathrm{ref}}$ (Line~5, Assumption~\ref{asm1:c_is_0}), so by
Assumption~\ref{asm1:infinite_data} every pair in $\T_{\pi^*_{\hat r}} \times \T_{\pi_{\mathrm{ref}}}$ is in
$\D_{t-1}$, the dataset on which $\hat r$ is fit. Hence Lemma~\ref{lem:ranking-agreement}
applies to every such pair.

Pick any $\tau_1 \in \T_{\pi^*_{\hat r}}$ and $\tau_2 \in \T_{\pi_{\mathrm{ref}}}$. By the optimality
of $\pi^*_{\hat r}$ under $\hat r$:

\[
\mathrm{regret}(\tau_1 \mid \hat r) = 0 \;\le\; \mathrm{regret}(\tau_2 \mid \hat r).
\]
By Lemma~\ref{lem:ranking-agreement}, this inequality is preserved under $r$, so
\[
\mathrm{regret}(\tau_1 \mid r) \;\le\; \mathrm{regret}(\tau_2 \mid r).
\]
Note that this implies
\begin{equation*}
\max_{\tau_1} \mathrm{regret}(\tau_1 \mid r)\;\le\; \min_{\tau_2} \mathrm{regret}(\tau_2 \mid r).
\end{equation*}
On the left hand side, the maximum is an upper bound to the expectation over $\tau_1 \in \T_{\pi^*_{\hat r}}$, and the minimum is a lower bound to the expectation over $\tau_2 \in \T_{\pi_{\mathrm{ref}}}$:
\begin{equation*}
\EE_{\tau_1 \in \T_{\pi^*_{\hat r}}}\!\big[\mathrm{regret}(\tau_1 \mid r)\big] \leq \max_{\tau_1} \mathrm{regret}(\tau_1 \mid r)\;\le\; \min_{\tau_2} \mathrm{regret}(\tau_2 \mid r) \leq \EE_{\tau_2 \in \T_{\pi_{\mathrm{ref}}} }\!\big[\mathrm{regret}(\tau_2 \mid r)\big].
\end{equation*}
Therefore, by Assumption~\ref{asm1:infinite_data},
\begin{equation*}
\EE_{\tau_1 \sim {\pi^*_{\hat r}}}\!\big[\mathrm{regret}(\tau_1 \mid r)\big]
\;\le\;
\EE_{\tau_2 \sim {\pi_{\mathrm{ref}}} }\!\big[\mathrm{regret}(\tau_2 \mid r)\big].
\end{equation*}
Under the single-start-state assumption (Assumption~\ref{asm1:same_start}),
\begin{equation*}
\EE_{\tau \sim \pi}\!\big[\mathrm{regret}(\tau \mid r)\big] \;=\; V_r^*(s_0) - V_r^{\pi}(s_0).
\end{equation*}
Substituting this identity and rearranging gives
\begin{equation*}
V_r^*(s_0) - V_r^{\pi^*_{\hat r}}(s_0)
\;\le\;
V_r^*(s_0) - V_r^{\pi_{\mathrm{ref}}}(s_0).
\end{equation*}
Rearranging terms again gives
\begin{equation*}
V_r^{\pi^*_{\hat r}}(s_0) \;\ge\; V_r^{\pi_{\mathrm{ref}}}(s_0)
\end{equation*}
which is equivalent to
\begin{equation*}
J_r(\pi^*_{\hat r}) \;\ge\; J_r(\pi_{\mathrm{ref}}).
\end{equation*}
\end{proof}

\subsection{Theorem~\ref{thm:pbrr-performance}, Corollary \ref{corr:no-skalse-hacking}}
\label{app:no-skalse-hacking}

\citet{Skalse} defines reward hacking as:

\begin{definition} \citep{Skalse}
\label{def:hacking1_}
A pair of reward functions $(\tilde{r}_1, \tilde{r}_2)$ is hackable relative to a policy set $\Pi$ and an environment $(S, A, T, \gamma, p_0, \textunderscore)$ if there exists $\pi, \pi' \in \Pi$ such that
\begin{align*}
    J_{\tilde{r}_1}(\pi) > J_{\tilde{r}_1}(\pi') \text{ and } J_{\tilde{r}_2}(\pi) < J_{\tilde{r}_2}(\pi')
\end{align*}
\end{definition}

This canonical definition of reward hacking intuitively states that, if $(\tilde{r}_1, \tilde{r}_2)$ is not hackable given a set of policies $\Pi$ , then there does not exist any pair of policies in $\Pi$ such that one policy has a strictly higher expected return under $\tilde{r}_1$ while the other has a strictly higher expected return under $\tilde{r}_2$. Framed differently, any increase in expected return under $\tilde{r}_1$ from switching policies in $\Pi$ never decreases the expected return under $\tilde{r}_2$. This definition is particularly strict, as noted by \cite{Skalse}.

\begin{proof}[Proof of Corollary~\ref{corr:no-skalse-hacking}]
Let $\Pi = \{\pi^*_{\hat r_t}, \pi_{\mathrm{ref}}\}$. By Definition~\ref{def:hacking1_}, $(\hat r_t, r)$ is
hackable relative to $\Pi$ only if one policy in $\Pi$ has strictly higher return under $\hat r_t$
and strictly lower return under $r$. Since $\pi^*_{\hat r_t}$ is optimal for $\hat r_t$,
$J_{\hat r_t}(\pi_{\mathrm{ref}}) > J_{\hat r_t}(\pi^*_{\hat r_t})$ is impossible, and by
Theorem~\ref{thm:pbrr-performance}, $J_r(\pi^*_{\hat r_t}) < J_r(\pi_{\mathrm{ref}})$ is impossible.
Hence $(\hat r_t, r)$ is not hackable relative to $\Pi$.
\end{proof}



\subsection{Theorem~\ref{thm:pbrr-performance}, Corollary \ref{corr:no-laidlaw-hacking}}
\label{app:no-laidlaw-hacking}
\cite{laidlaw2024correlated} define reward hacking as:

\begin{definition} \citep{laidlaw2024correlated} 
\label{def:hacking2_}
    Suppose $\tilde{r}$ is a $\rho$-correlated proxy\footnote{For the full definition of $\rho$-correlation, we refer readers to the original paper from \citet{laidlaw2024correlated}.} with respect to $\pi_\text{ref}$. The proxy reward function is hackable with respect to the ground-truth reward function and  $\pi_{\text{ref}}$ if,
    \begin{align*}
        J_r(\pi^*_{\tilde{r}}) < J_r(\pi_\text{ref}).
    \end{align*}
\end{definition}

This definition of reward hacking only considers reward functions that are reasonable optimization targets, i.e., by requiring the inputted reward function to be $\rho$-correlated with the ground-truth reward function, and sets a threshold for reward hacking via the expected return of the reference policy under the ground-truth reward function. This definition is less strict than that of \citet{Skalse}; see \cite{laidlaw2024correlated} for an in-depth comparison, 

\begin{proof}[Proof of Corollary~\ref{corr:no-laidlaw-hacking}]
By Definition~\ref{def:hacking2_}, $\hat r_t$ is hackable with respect to $r$ and $\pi_{\mathrm{ref}}$
only if $J_r(\pi^*_{\hat r_t}) < J_r(\pi_{\mathrm{ref}})$, which Theorem~\ref{thm:pbrr-performance} rules out.
\end{proof}

\section{Plotting Details and Additional Results}
\label{app:additional_results}

\subsection{Plotting details}
\label{app:plot_details}
For Figures \ref{fig:main_results} and \ref{fig:abalation}, for each seed we roll out a policy for the number of episodes specified in Appendix Table \ref{tab:n_episodes_mean_return}. We compute the mean return over all episodes with respect to the ground-truth reward function. We then scale the mean return for policy $\pi$ as follows:
\begin{center}
    $\hat{J}_\text{$r$-scaled}(\pi)  = \frac{\hat{J}(\pi) - \hat{J}_r(\pi_\text{ref})}{ \hat{J}_r(\pi^*_{r}) - \hat{J}_r(\pi_\text{ref})}:$

\end{center}

where $\hat{J}_r$ is the empirical mean return under the ground-truth reward function $r$. We clip $\hat{J}_\text{$r$-scaled}(\pi)$ to be in range $[-1, 1]$. We perform this clipping for visual clarity, noting that any policy with a scaled return less than $0$ does worse than the supplied reference policy, and any policy with a scaled return less than $-1$ is considerably sub-optimal. In Figures \ref{fig:main_results} and \ref{fig:abalation}, we plot the resulting mean scaled return over 3 seeds. 
\begin{table}[!htbp]
\centering
\begin{tabular}{l c}
\hline
\textbf{Environment} & \textbf{Number of Episodes} \\
\hline
Traffic   & 260 \\
Pandemic  & 20  \\
Glucose   & 20  \\
AI safety gridworld    & 10  \\
\hline
\end{tabular}
\caption{Number of episodes used to compute the mean return.}
\label{tab:n_episodes_mean_return}
\end{table}

\subsection{Retraining a policy with an updated proxy reward function}
\label{app:retraining_ppo}

\begin{figure}[tb]
    \centering
    \includegraphics[width=0.32\textwidth]{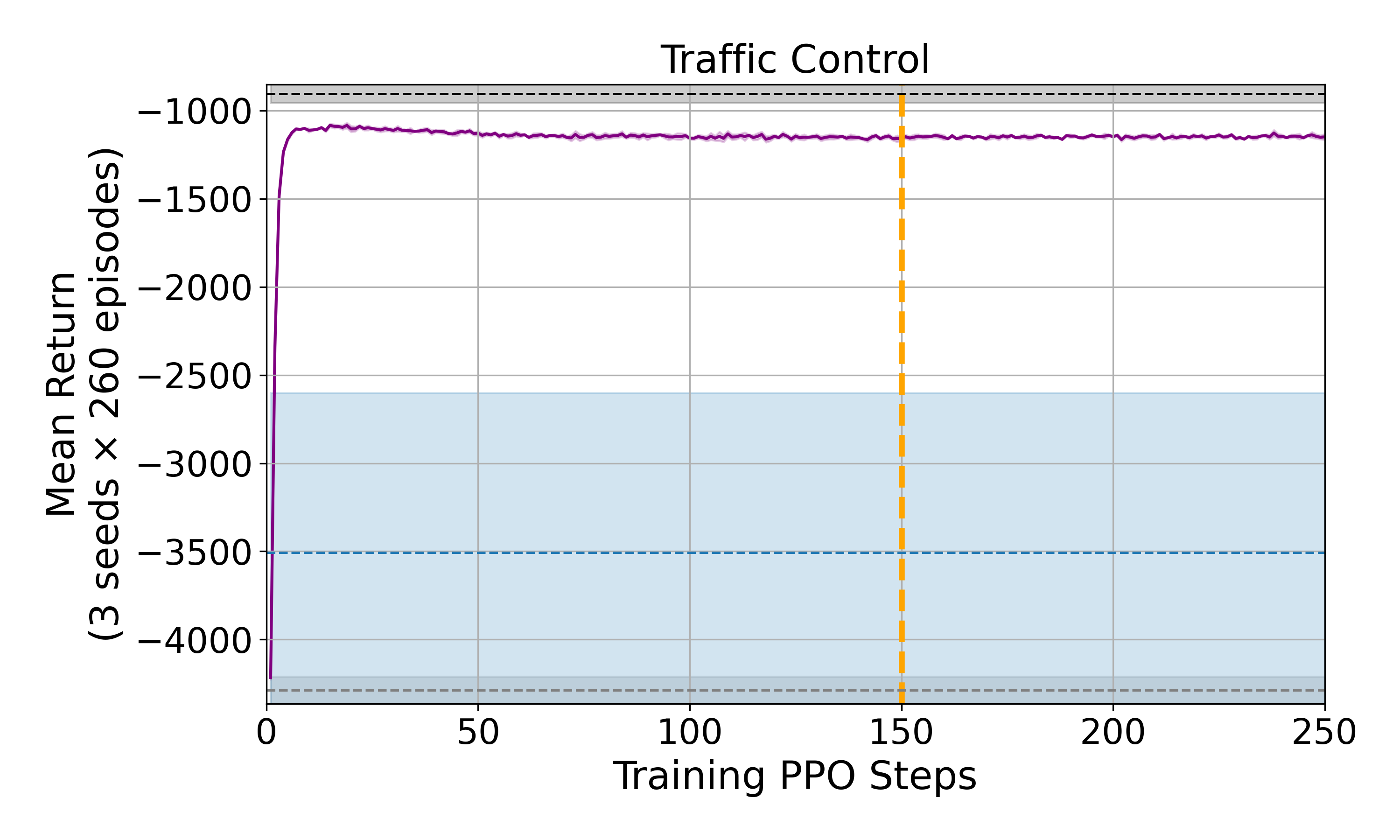}
    \includegraphics[width=0.32\textwidth]{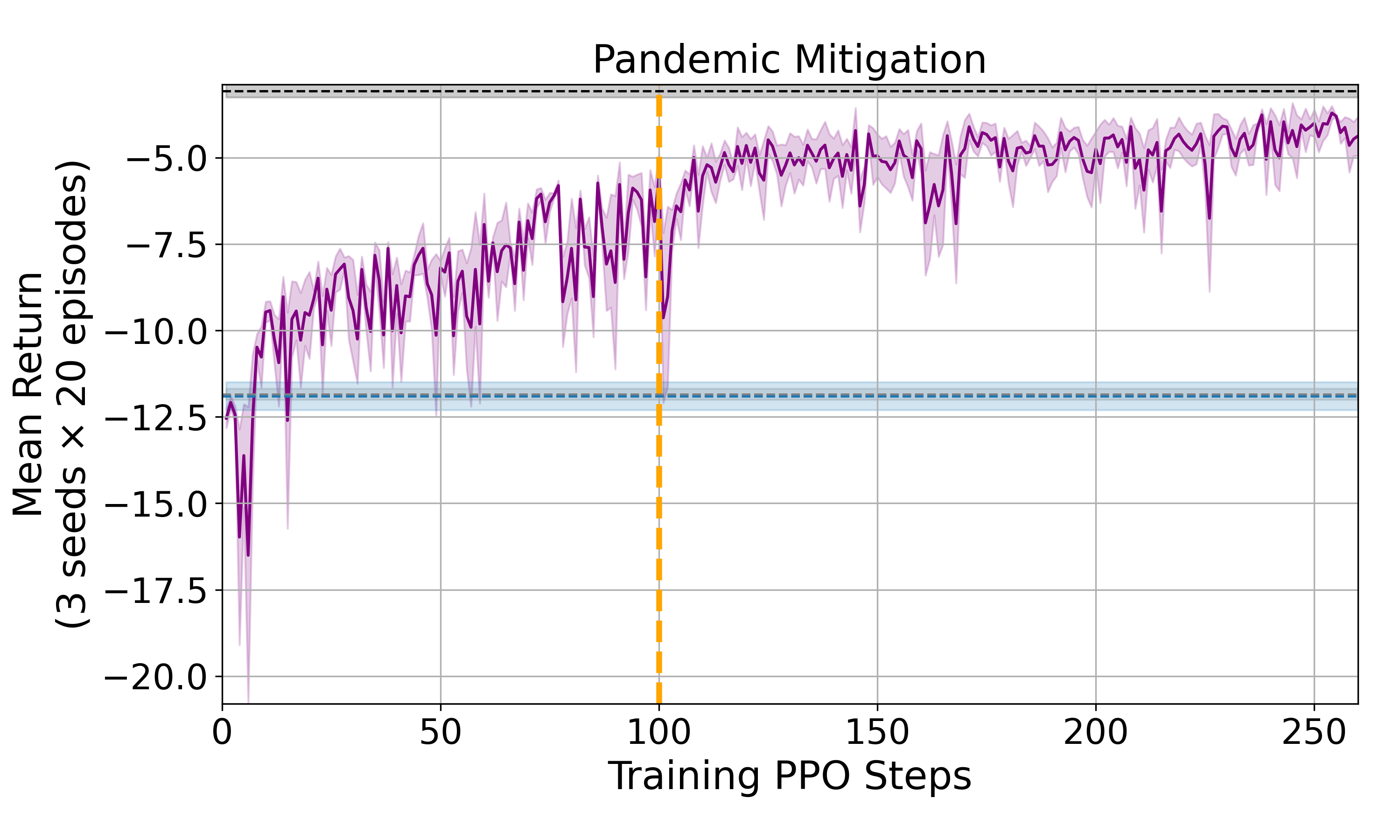}
    \includegraphics[width=0.32\textwidth]{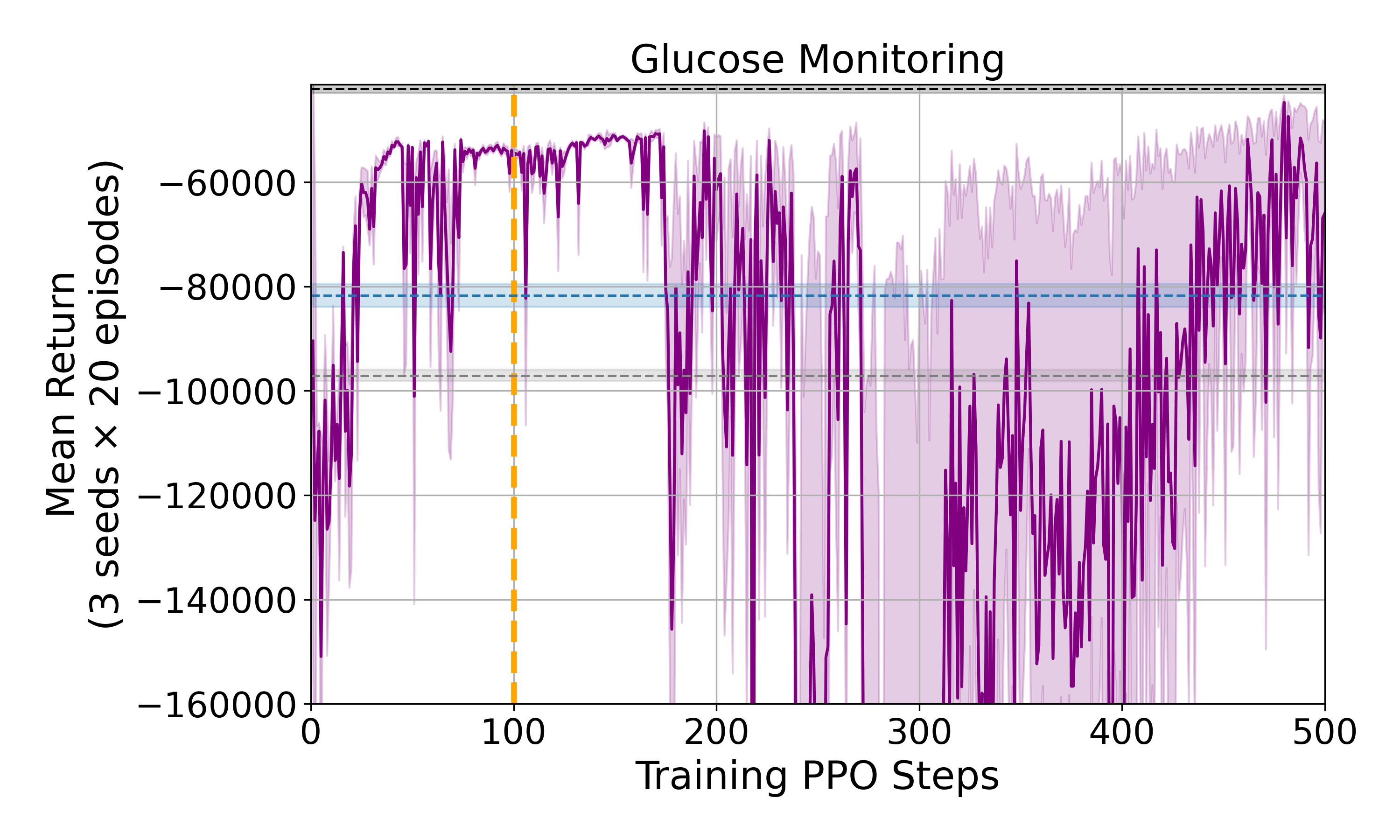}

    \includegraphics[width=0.85\textwidth]{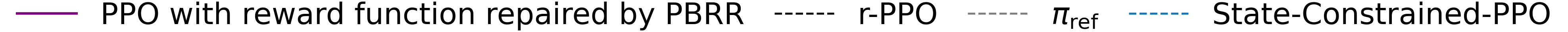}

    \caption{Mean return under the ground-truth reward function achieved by re-training a newly initialized policy with the repaired proxy reward function learned by PBRR. The vertical line marks the number of steps used to train the policies when updating the proxy reward function in Algorithm \ref{alg:exploration}; performance beyond this point illustrates the robustness of that updated proxy reward function to extended optimization. Results are averaged over trajectories sampled from policies trained with the updated proxy reward function across 3 random seeds. }
    \label{fig:retrain_ppo}
    \vspace{-3mm}
\end{figure}

\citet{mckinney2023fragility} show that reward functions learned online from human feedback can fail to re-train new policies initialized from scratch with a different random seed. To test whether PBRR’s updated proxy reward function exhibits similar fragility, we reinitialize new policies with different seeds and train the policies using the repaired reward functions learned in Section~\ref{sec:results}. We also extend PPO training beyond the number of PPO steps used when learning the additive correction term with PBRR, probing whether the updated proxy reward function could induce undesirable behavior if optimized for longer. Figure \ref{fig:retrain_ppo} presents these results: in the Traffic Control, Glucose Monitoring, and Pandemic environments, retraining with the repaired proxy reward function still yield policies that induce near-optimal performance. In the Pandemic Mitigation environment, training PPO for more steps even yields an improvement. In the Glucose Monitoring environment, however, the performance of the policy induced by the repaired proxy reward function deteriorates when trained with substantially more RL steps than were used during reward learning. These findings broadly suggest that PBRR’s repaired proxy reward function is robust to policy reinitialization and to longer RL optimization, but the number of RL steps used while updating the proxy reward function can play a critical role in determining whether the learned signal remains valid under extended training. In other words, when re-training a new policy with a repaired proxy reward function, it is likely best to train that policy with the same number of RL steps as used when repairing the proxy reward function.

\subsection{Comparisons against baselines that update $\pi_{\text{ref}}$}
\label{app:pi_ref_updating_exp_res}
Figure \ref{fig:main_results} compares PBRR against the Online-State-Constraint and RRM + State-Constraint baselines respectively, which each constrain the learned policy to $\pi_{\text{ref}}$. Section \ref{app:baselines} details these baseline implementations. Here we compare PBRR against those baselines, except we update $\pi_{\text{ref}}$ every iteration to be the policy constructed at the previous iteration as outlined in Section \ref{app:baselines}. We refer to these additional baselines as Online-RLHF + Moving-State-Constraint and RRM + Moving-State-Constraint. 

Our goal with these additional baselines is to attempt to overcome a fundamental limitation of the Online-State-Constraint and RRM + State-Constraint baselines, namely that they may fundamentally limit the performance of the induced policy by penalizing divergences from a fixed $\pi_{\text{ref}}$. Figure \ref{fig:move_pi_ref_abaltion} below shows that updating $\pi_{\text{ref}}$ as described in Section \ref{app:baselines} does not lead to a substantial improvement in performance; PBRR still remains the most performant approach in all environments. 
\begin{figure}[tb]
    \centering
    \includegraphics[width=0.48\textwidth]{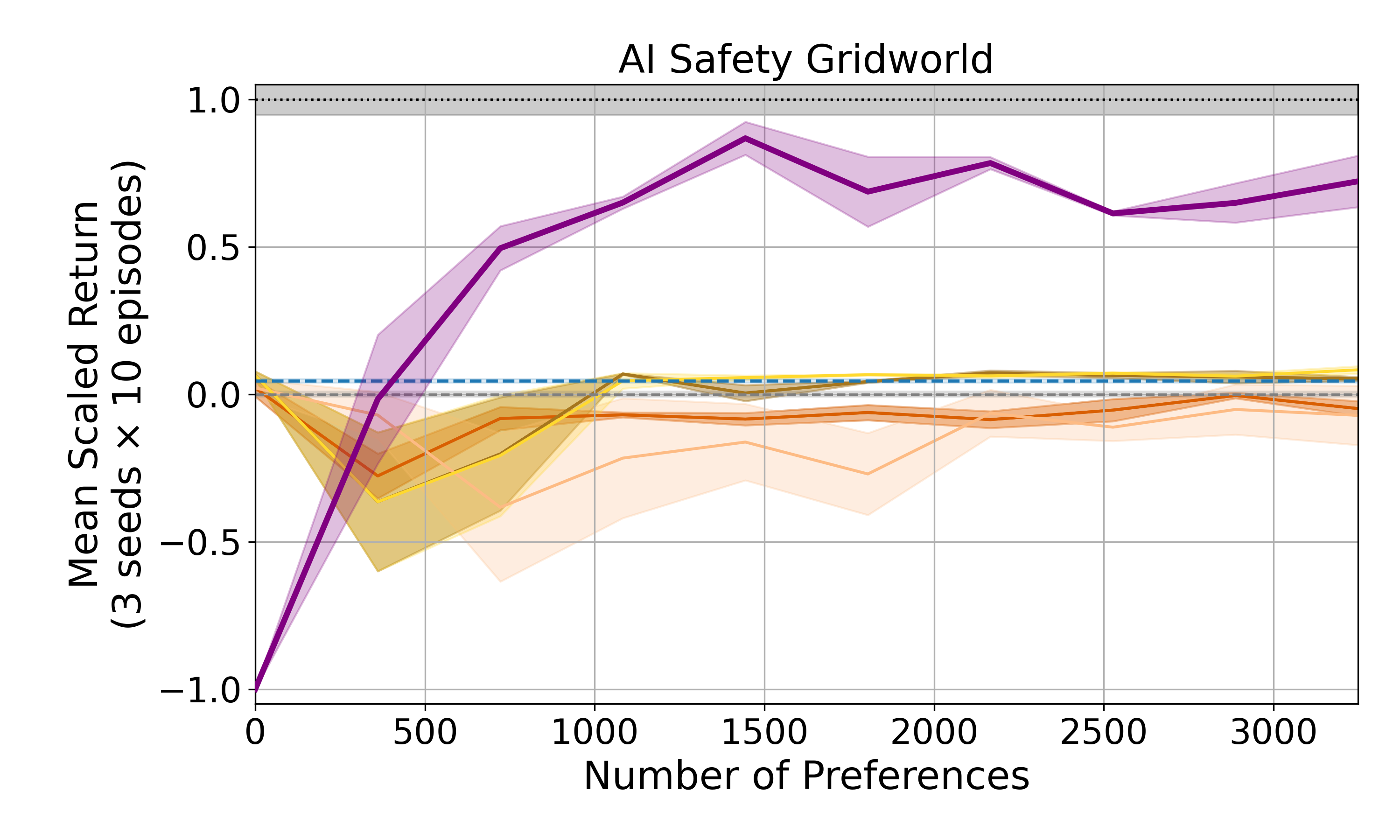}
    \includegraphics[width=0.48\textwidth]{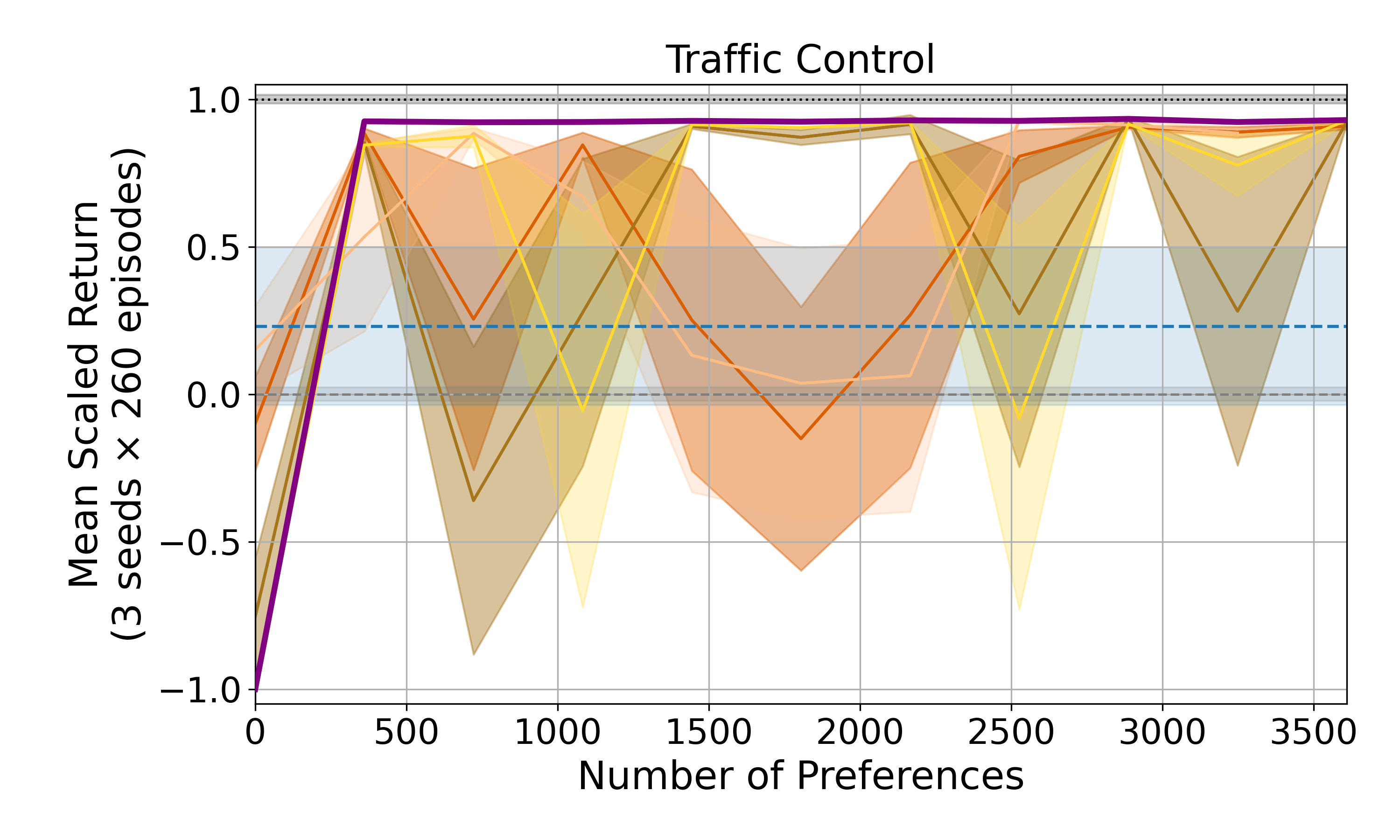}\\[0.5em]
    \includegraphics[width=0.48\textwidth]{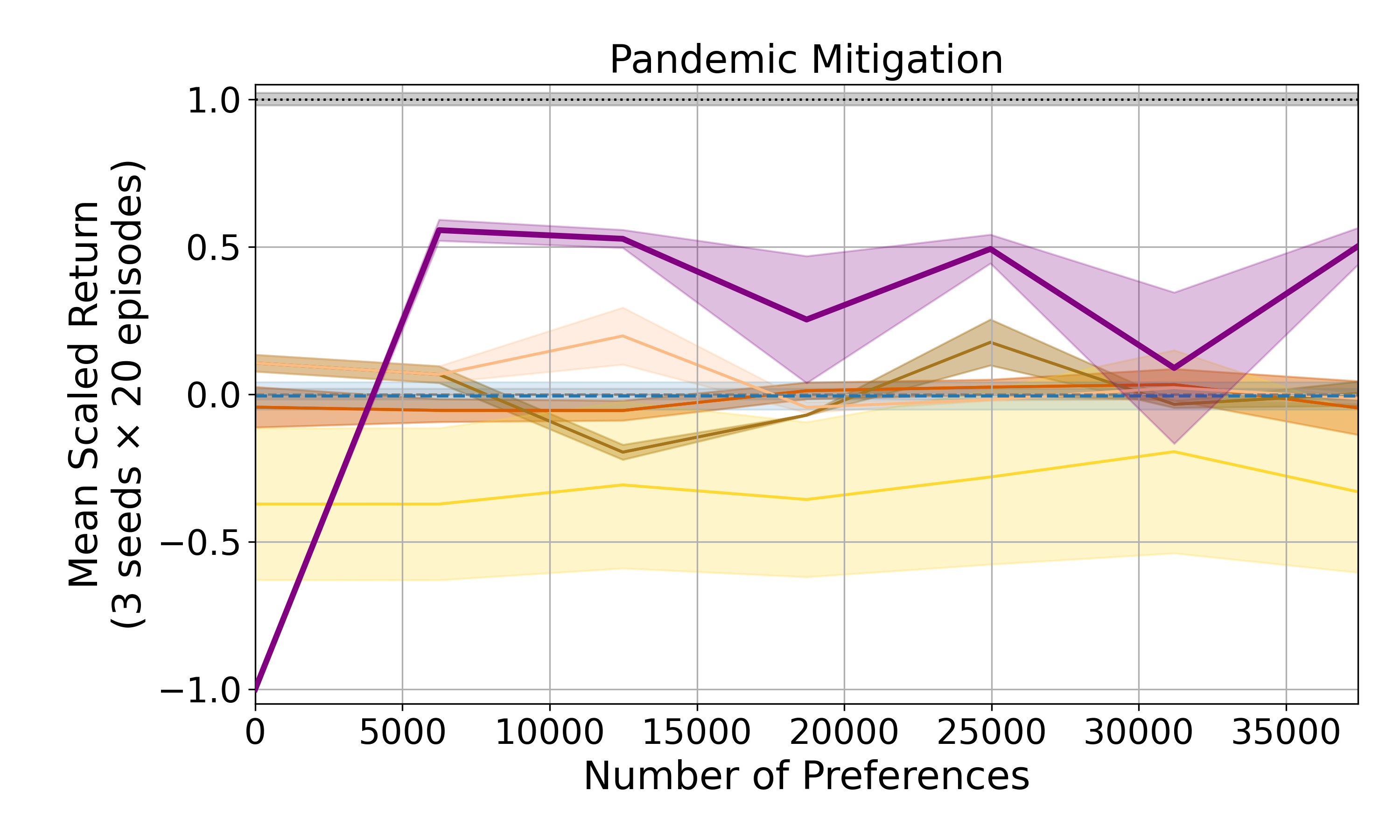}
    \includegraphics[width=0.48\textwidth]{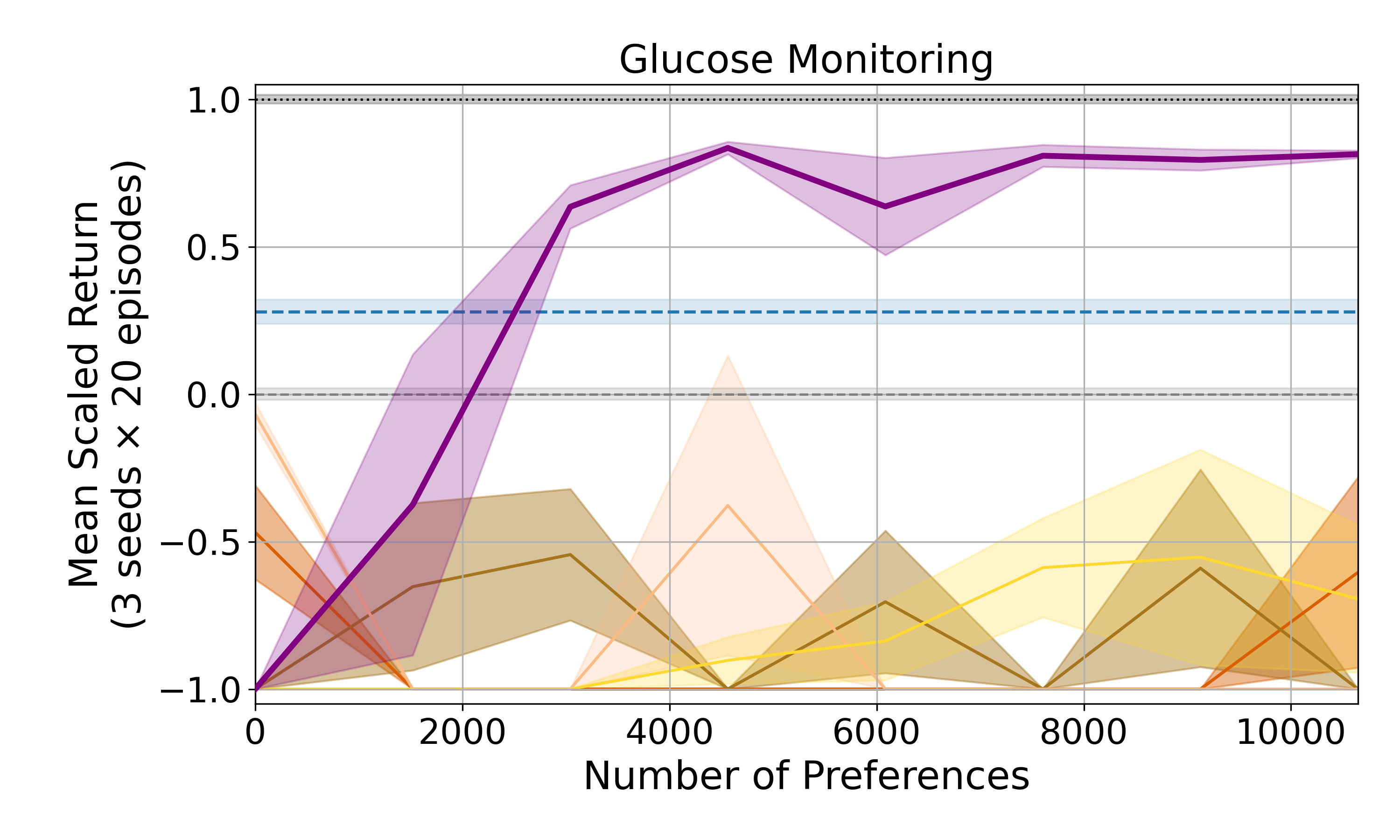}
    \includegraphics[width=0.99\textwidth]{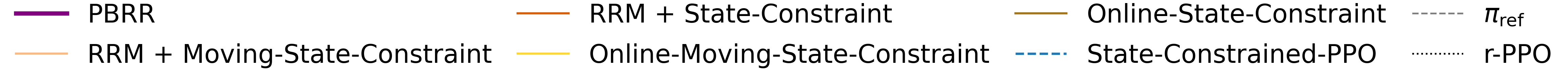}
    \caption{Mean return under the ground-truth reward function achieved by the Online-State-Constraint and RRM + State-Constraint baselines, and variants of those baselines that update $\pi_{\text{ref}}$ used for the state-constraint at each iteration as described in Section \ref{app:baselines}. See Figure \ref{fig:main_results} for plotting details.}
    \label{fig:move_pi_ref_abaltion}
\end{figure}

\subsection{Ablating PBRR's regularization terms}
\label{app:pbrr_reg_ablation}
\begin{figure}[tb]
    \centering
    \includegraphics[width=0.48\textwidth]{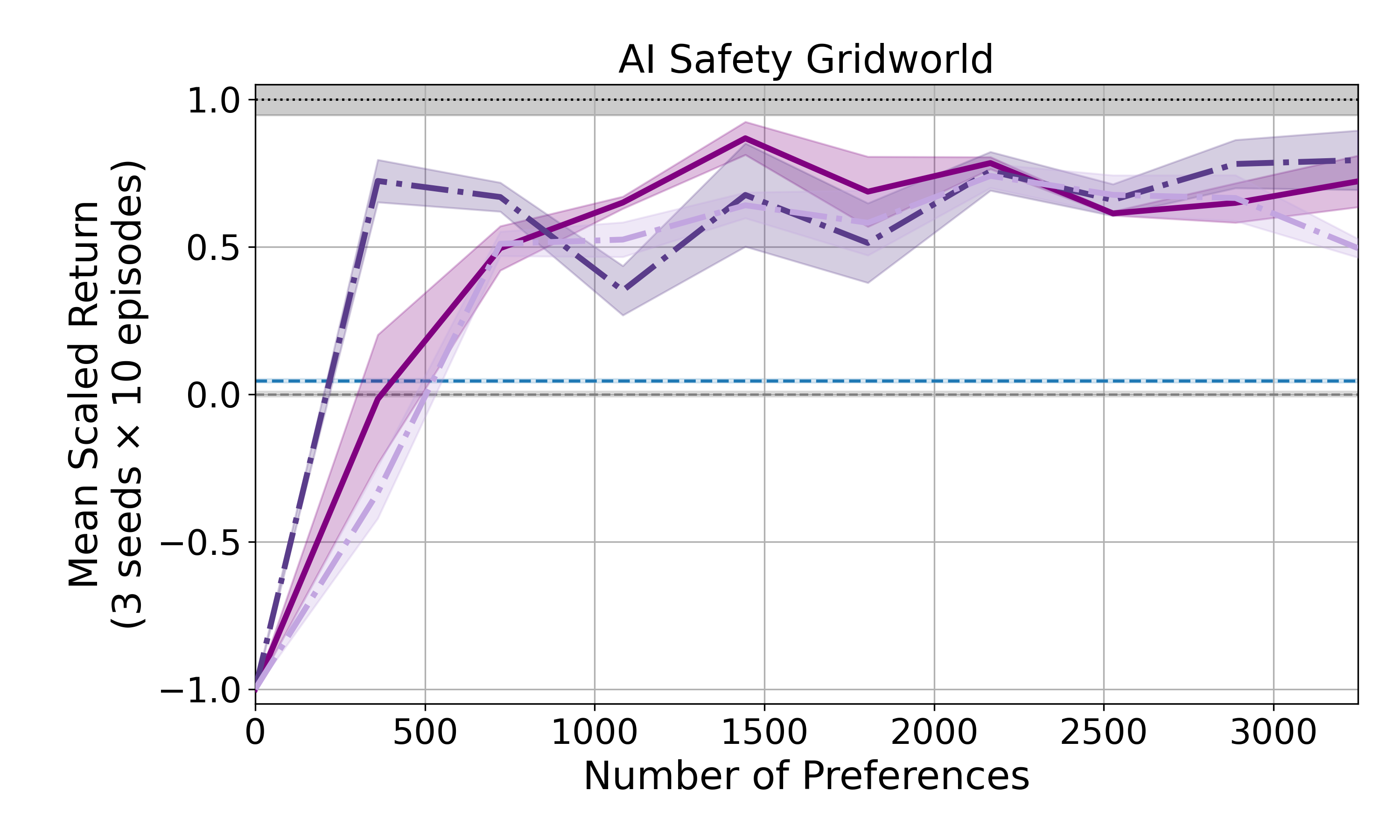}
    \includegraphics[width=0.48\textwidth]{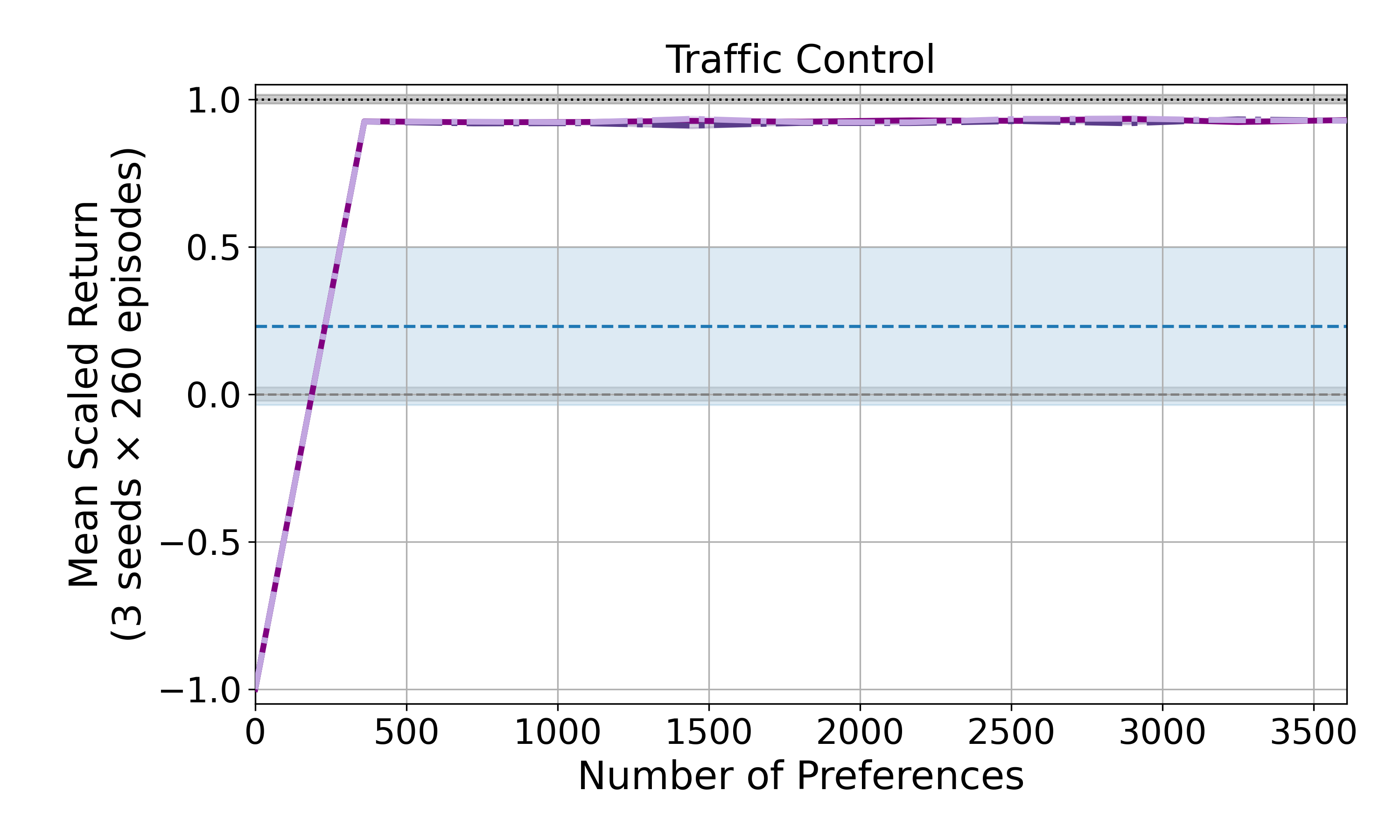}\\[0.5em]
    \includegraphics[width=0.48\textwidth]{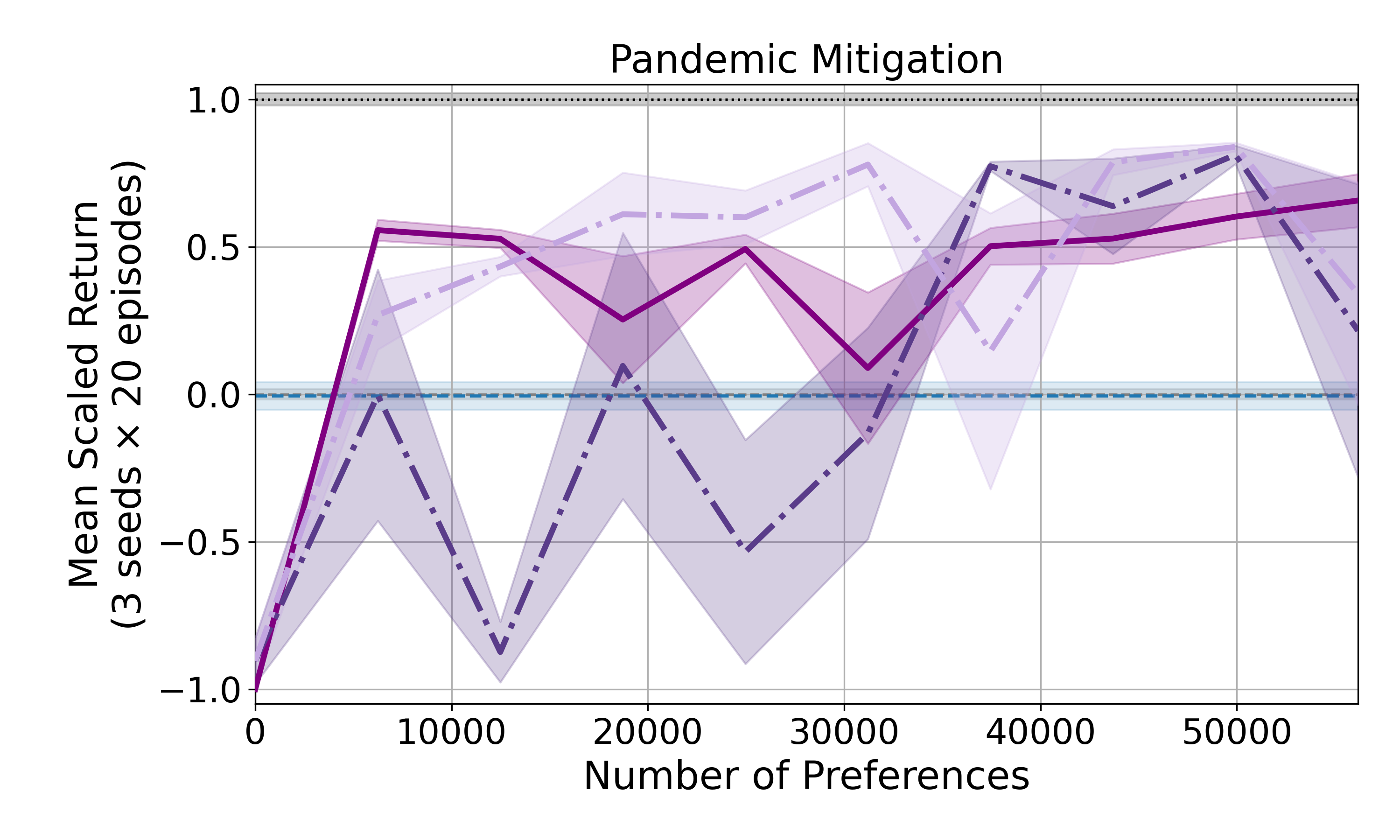}
    \includegraphics[width=0.48\textwidth]{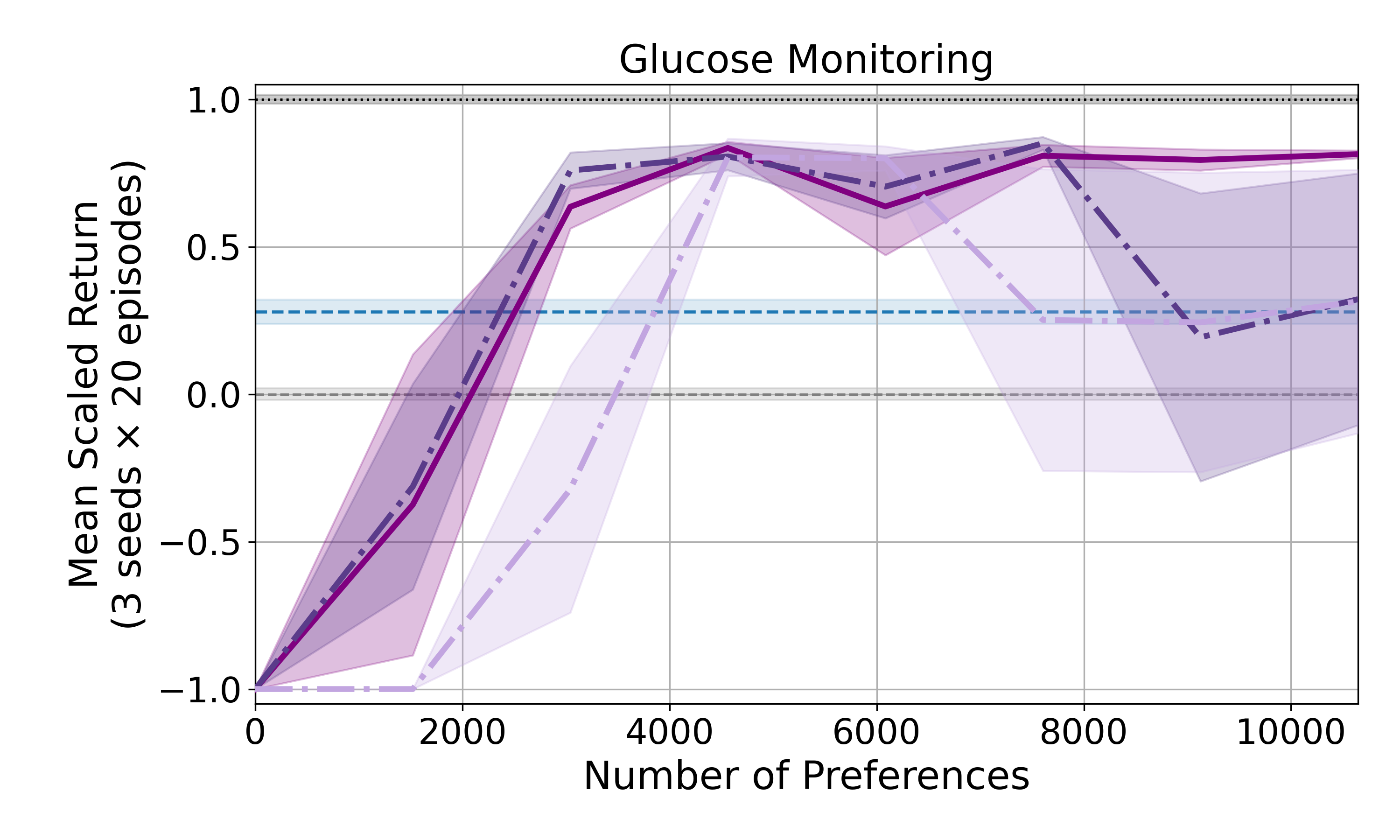}
    \includegraphics[width=0.85\textwidth]{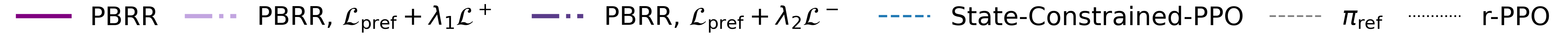}
    \caption{Mean return under the ground-truth reward function achieved by PBRR, compared to PBRR using the standard preference-learning objective $\mathcal{L}_{\text{pref}}$  (Eq.\ref{eq:loss}) with (i) only the $\mathcal{L}^-$ regularization term, and (ii) only the $\mathcal{L}^+$ regularization term. Note that PBRR uses the preference learning objective in Eq. \ref{eq:reg-pref-loss} with all three terms: $\mathcal{L}_{\text{pref}} + \lambda_1\mathcal{L}^+ + \lambda_2\mathcal{L}^-$. See Figure \ref{fig:main_results} for plotting details.}
    \label{fig:reg_further_abalation}
\end{figure}

PBRR's preference learning objective in Eq. \ref{eq:reg-pref-loss} consists of two regularization terms, $\mathcal{L}^-$ and $\mathcal{L}^+$. In Figure \ref{fig:abalation} we empirically show that, without these regularization terms, PBRR performs relatively poorly in both stability and achieved mean return. Here we investigate the individual impact of $\mathcal{L}^-$ and $\mathcal{L}^+$ respectively. Figure \ref{fig:reg_further_abalation} illustrates that PBRR with both regularization terms matches or outperforms all alternatives across environments. The relative effect of using only $\mathcal{L}^-$ or $\mathcal{L}^+$ is environment dependent. PBRR's performance is only degraded by removing a regularization term in the Pandemic Mitigation and Glucose Monitoring environments: removing $\mathcal{L}^+$ causes the larger drop in Pandemic, while removing $\mathcal{L}^-$ has the greater effect in Glucose.

\subsection{PBRR with intra-policy preferences}
\label{app:pbrr_tntra_pol_prefs}

\begin{figure}[tb]
    \centering    \includegraphics[width=0.48\textwidth]{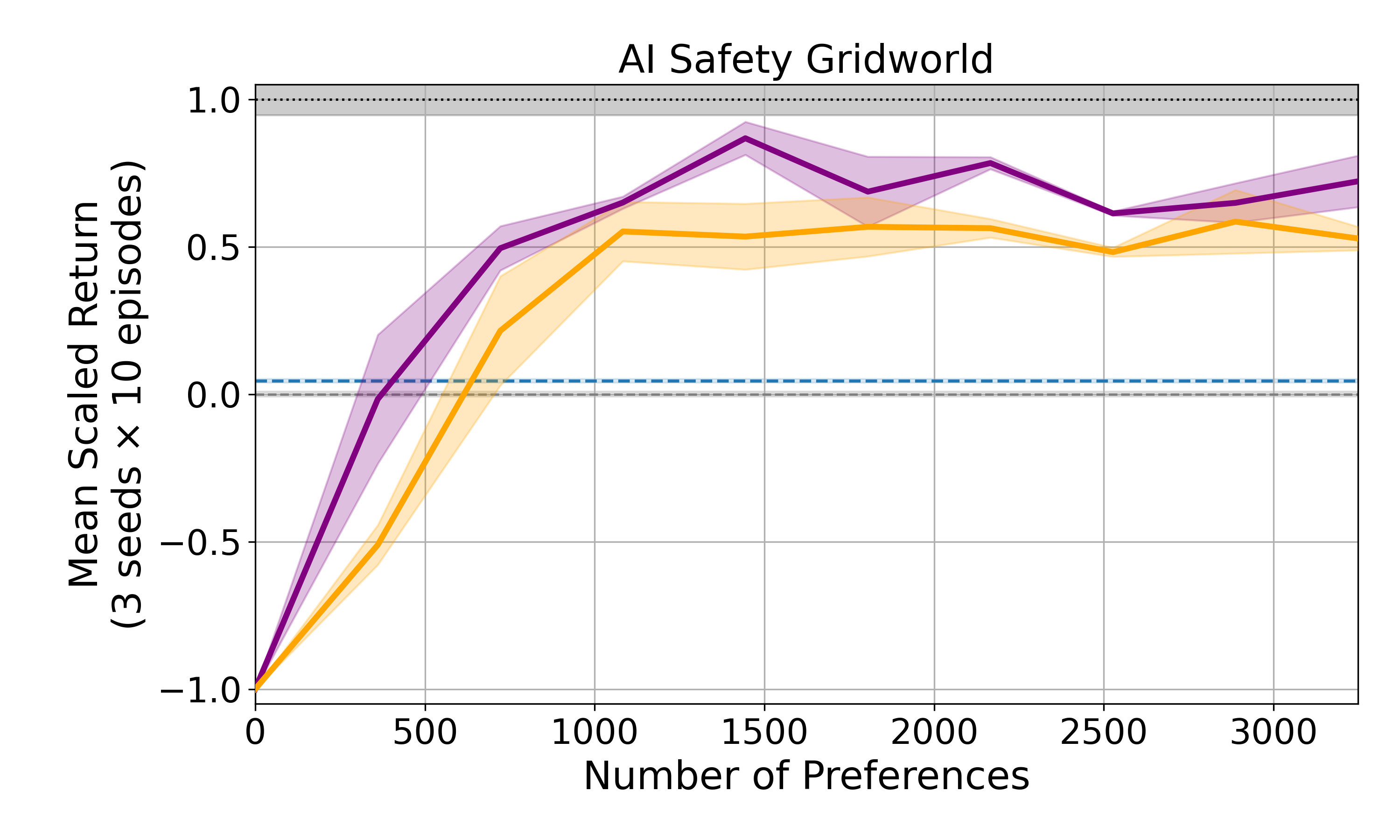}
    \includegraphics[width=0.48\textwidth]{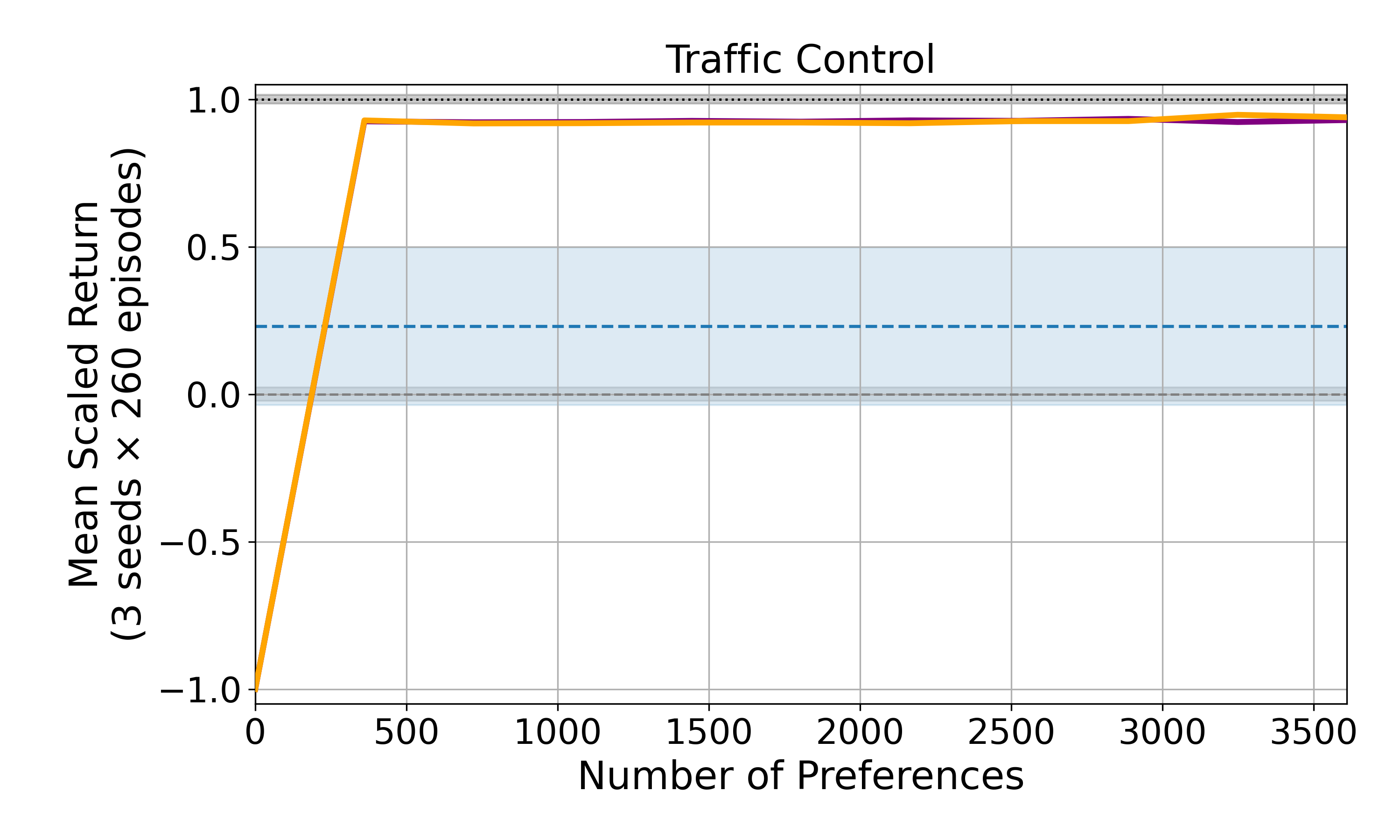}\\[0.5em]
    \includegraphics[width=0.48\textwidth]{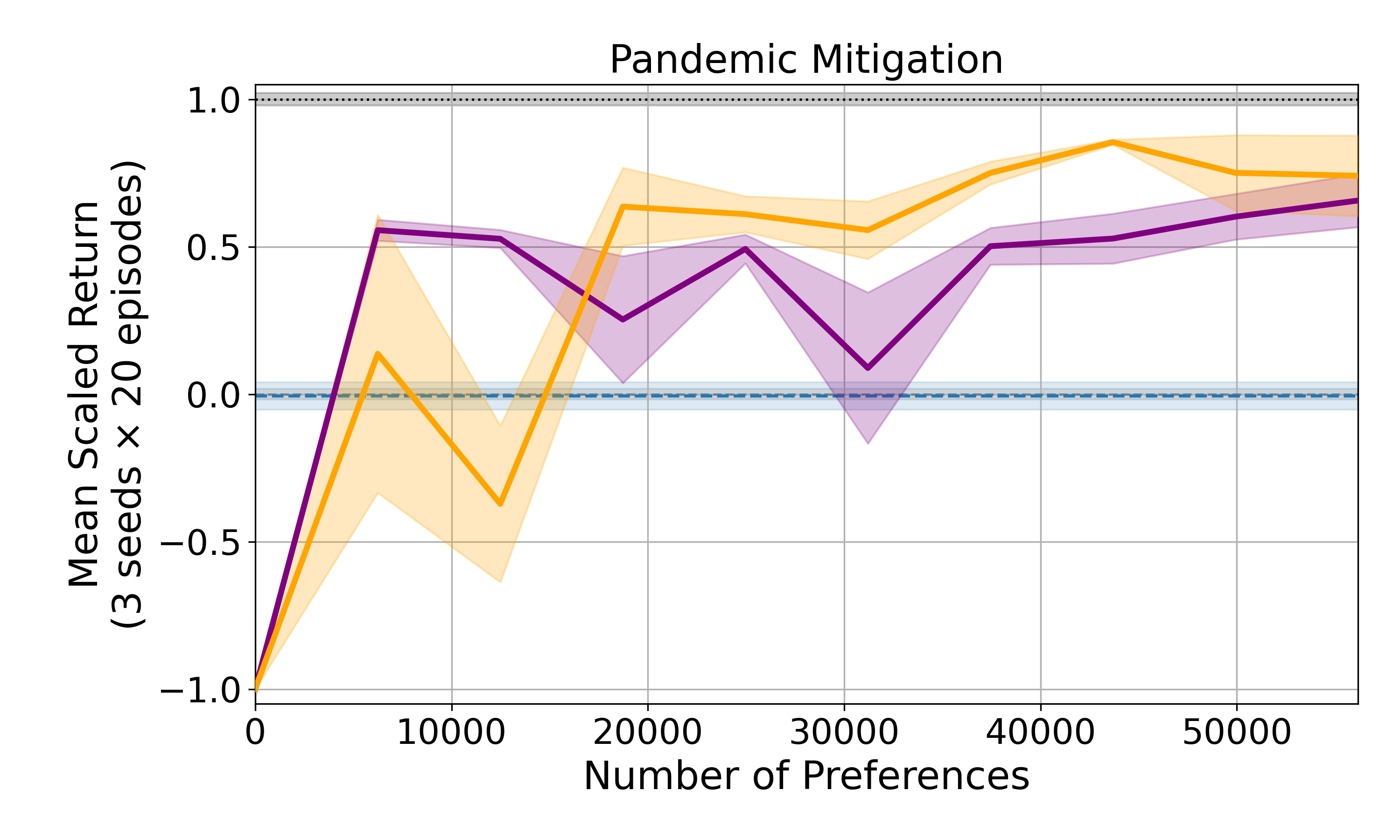}
    \includegraphics[width=0.48\textwidth]{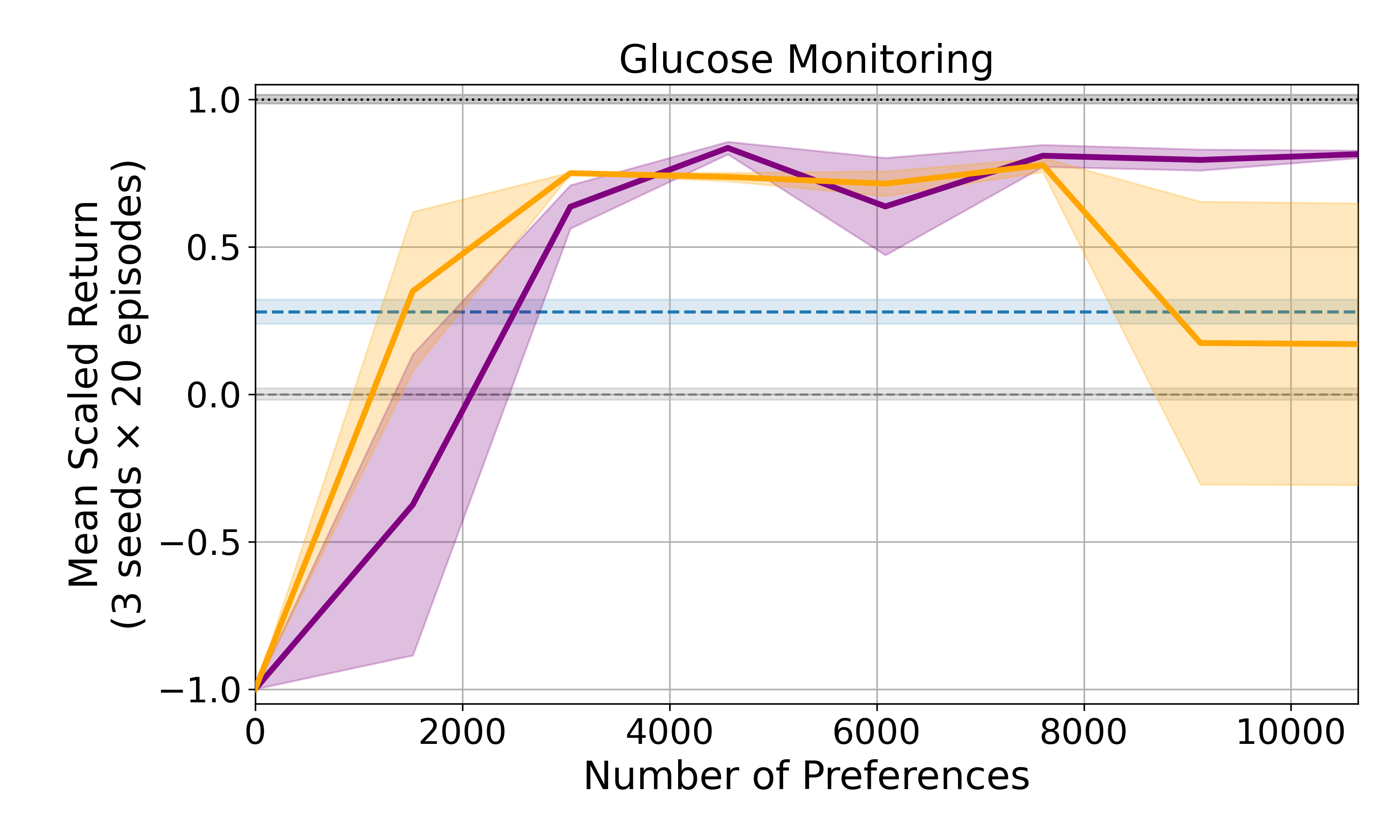}
    \includegraphics[width=0.85\textwidth]{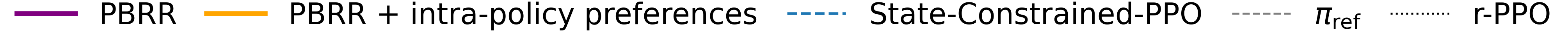}
    \caption{Mean return under the ground-truth reward function achieved by PBRR, compared to PBRR + intra-policy preferences. See Figure \ref{fig:main_results} for plotting details.}
    \label{fig:pbrr_tntra_pol_prefs}
\end{figure}

PBRR, as implemented for the results in Section \ref{sec:results}, elicits preferences over trajectory pairs where one trajectory is sampled from $\pi_{\hat{r}_t}$ and the other from $\pi_{\text{ref}}$. Here we investigate adding intra-policy preferences, i.e., preferences over trajectory pairs where both trajectories are sampled from $\pi_{\hat{r}_t}$. Figure \ref{fig:pbrr_tntra_pol_prefs} plots the results when half of the elicited batch is intra-policy preferences. We find that including these preferences degrades data-efficiency and stability. In particular, the performance in the early iterations of the PBRR + intra-policy preferences approach is worse than PBRR in the AI Safety Gridworld and Pandemic environment, and performance decreases in later iterations in the Glucose environment. For the Pandemic environment, PBRR + intra-policy preferences eventually outperforms PBRR, suggesting a potential benefit to including intra-policy preferences. Future work should explore how to leverage this additional data to improve PBRR's performance without reducing data-efficiency or learning stability in environments where PBRR already performs well. 

\subsection{Unscaled Main Results}
\label{app:unscaled_res}
\begin{figure}[!htbp]
  \centering
  \includegraphics[width=0.48\textwidth]{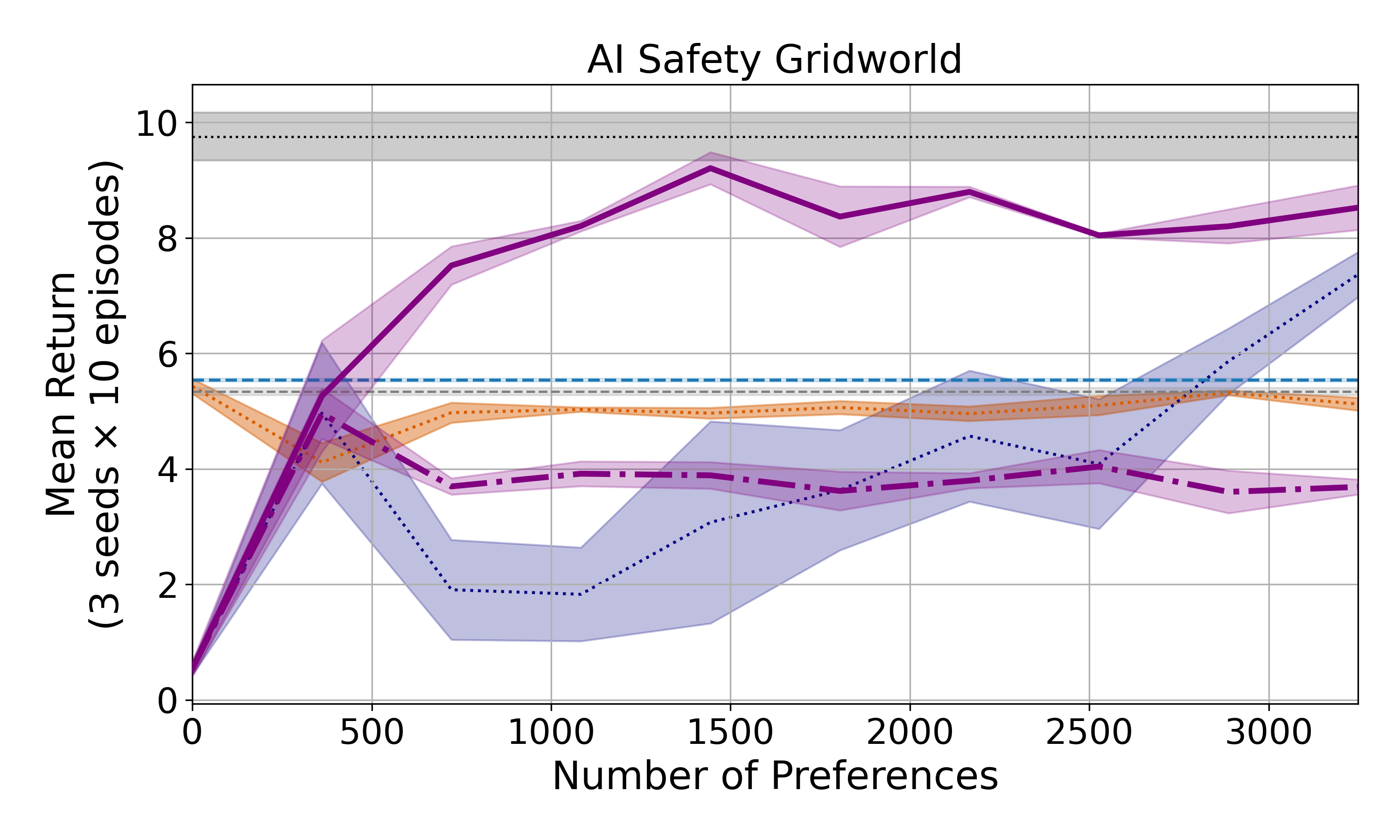}
    \includegraphics[width=0.48\textwidth]{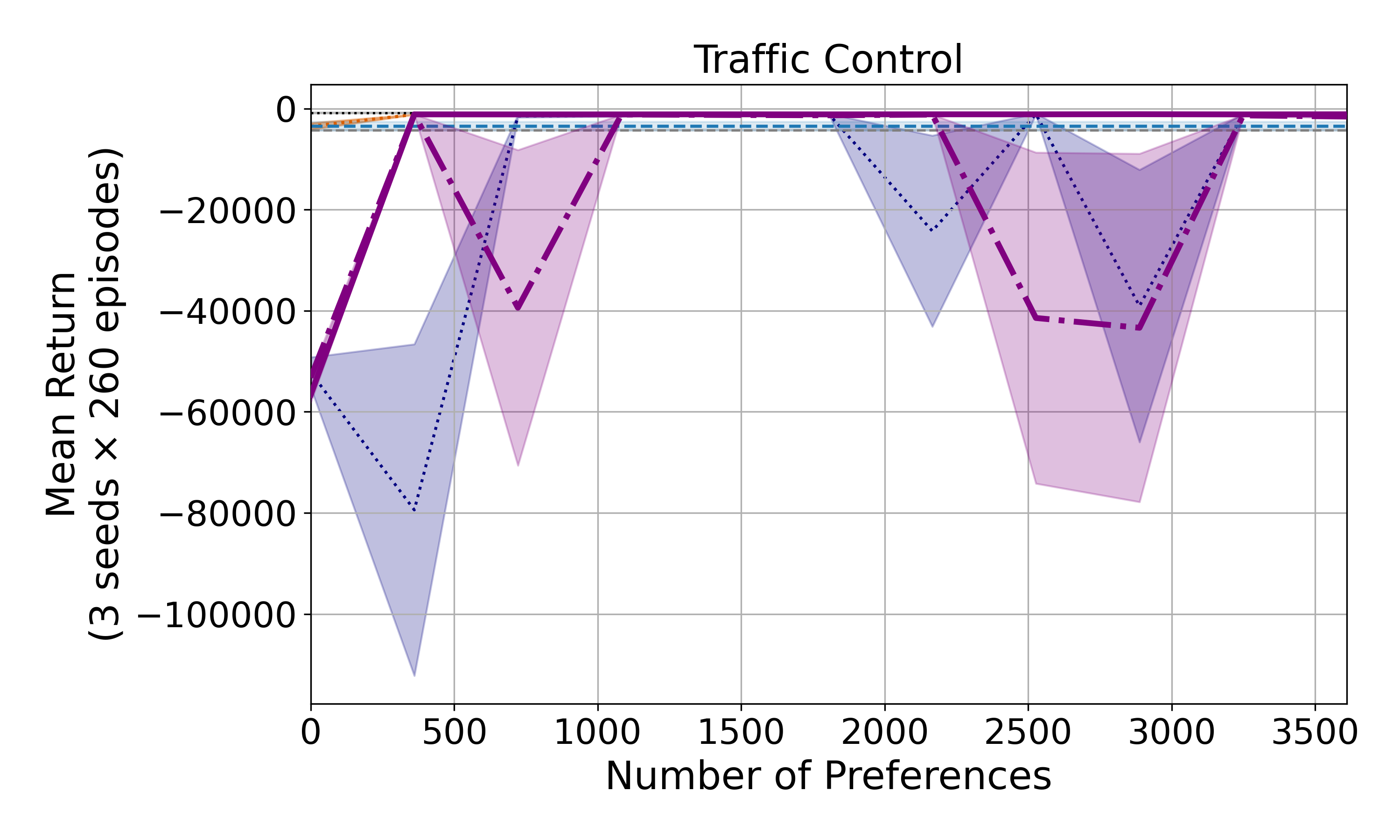}\\[0.6em]
    \includegraphics[width=0.48\textwidth]{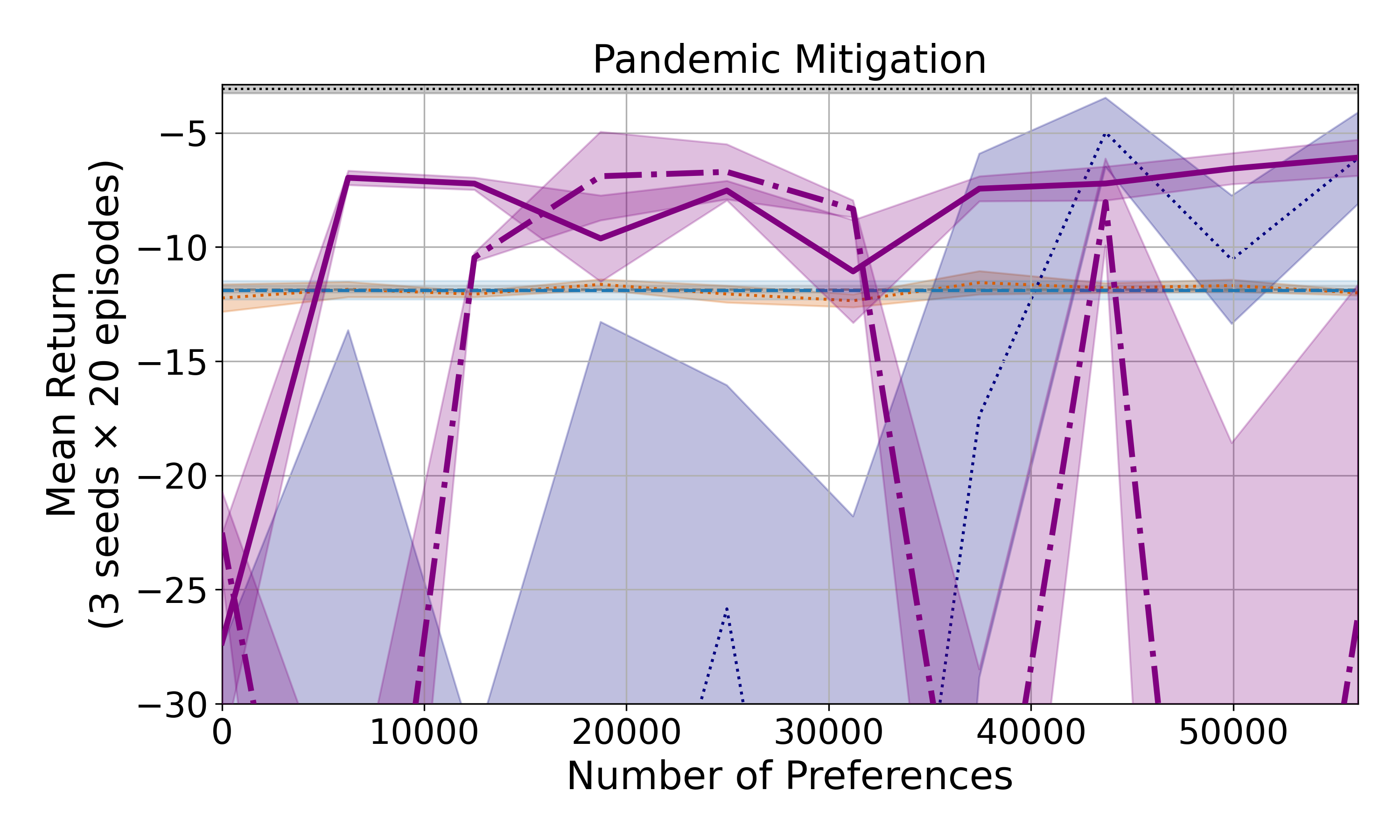}
    \includegraphics[width=0.48\textwidth]{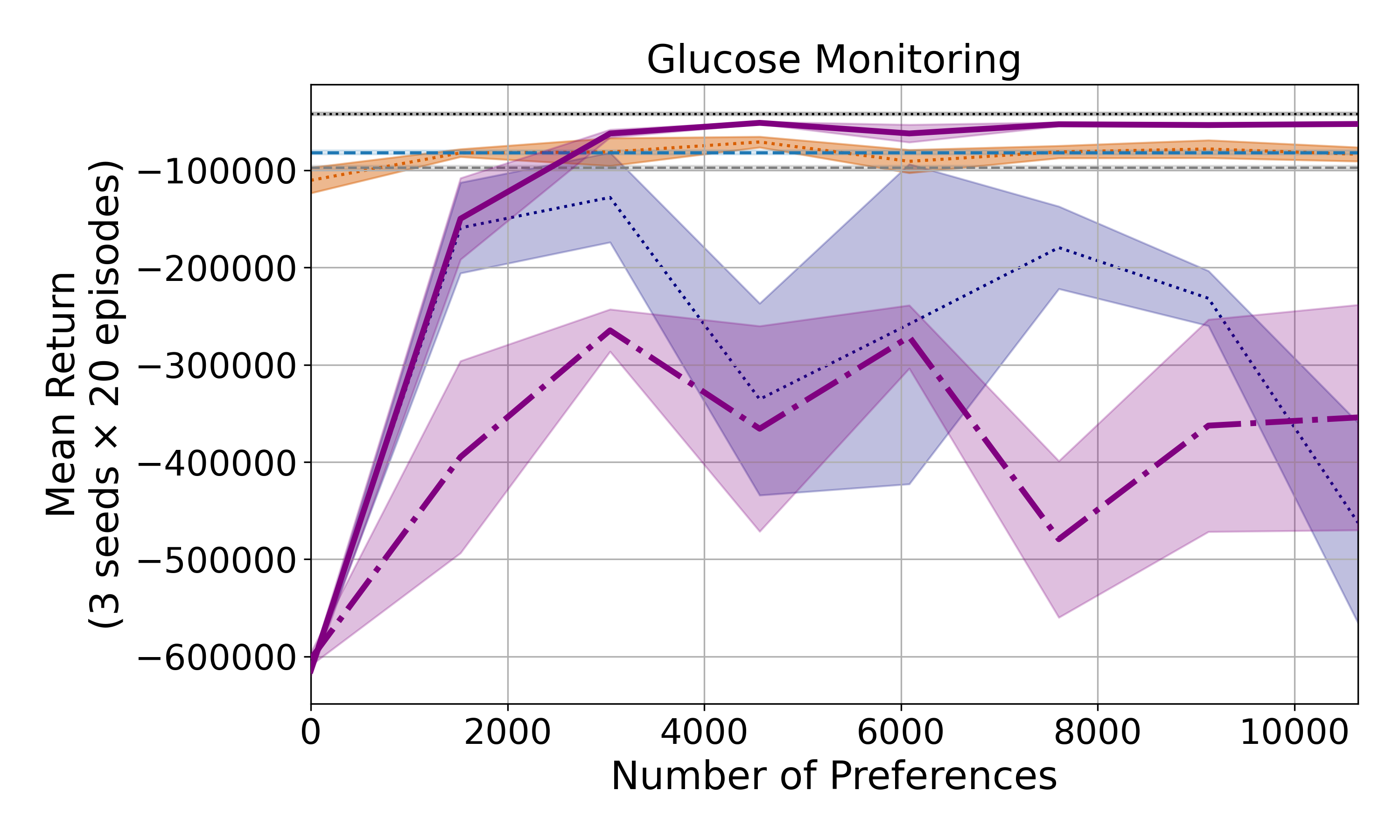}\\[0.6em]
    \includegraphics[width=0.85\textwidth]{images/frh_results_glucose_legend_scaled_abalation_scaled=True.png}
    \caption{Complementing Figure \ref{fig:abalation}: Mean return under the ground-truth reward function achieved by PBRR, compared to (i) PBRR using the standard preference-learning objective (Eq.\ref{eq:loss}) instead of our proposed objective (Eq.\ref{eq:reg-pref-loss}), and (ii) other methods that repair the proxy reward function equipped with our proposed objective. The mean return is averaged over trajectories sampled from policies trained with the updated proxy reward function across 3 random seeds. No scaling or clipping is applied to the plotted mean return values.}
    \label{fig:abalation_unscaled}
\end{figure}

In Figures \ref{fig:main_results} and  \ref{fig:abalation} we scale and clip the plotted mean return, as detailed in Appendix \ref{app:plot_details}, for clearer visual comparison. In Figure  \ref{fig:main_results_r_from_scratch_unscaled} we plot the unclipped, unscaled mean-return for PBRR and the baselines that learn a reward function \textit{ab initio}---Online-RLHF and Online-State-Constraint. In Figure \ref{fig:main_results_r_proxy_mod_baselines_unscaled} we plot those results for PBRR and the baselines that learn to repair a proxy reward function---RRM and RRM + State-Constraint. In Figure \ref{fig:abalation_unscaled} we plot those results for the ablations from Figure \ref{fig:abalation}.

\begingroup
\setlength{\textfloatsep}{8pt plus 2pt minus 2pt}
\setlength{\floatsep}{8pt plus 2pt minus 2pt}
\setlength{\intextsep}{8pt plus 2pt minus 2pt}

\begin{figure}[p]
  \centering

  \begin{subfigure}[t]{\textwidth}
    \centering
    \includegraphics[width=0.45\textwidth]{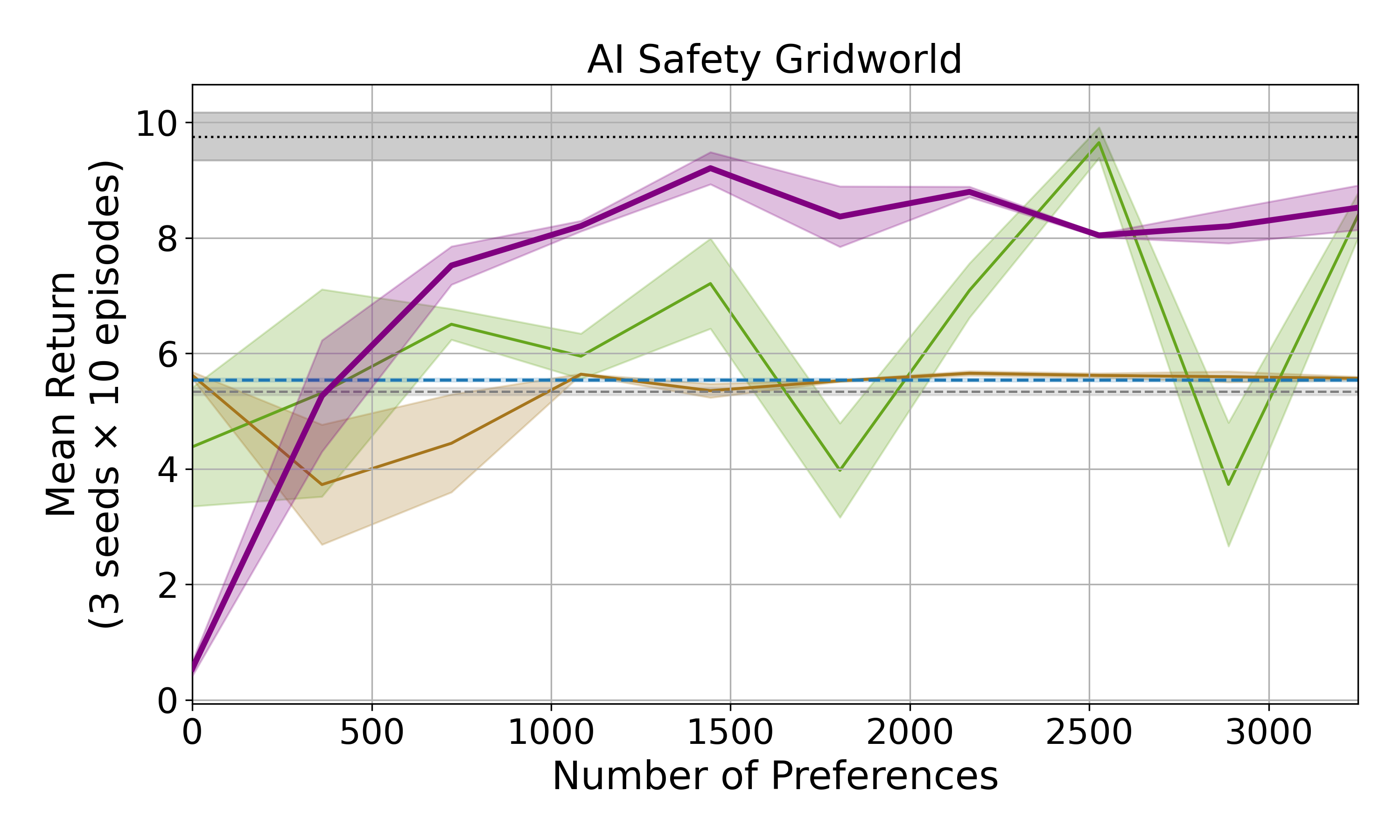}
    \includegraphics[width=0.45\textwidth]{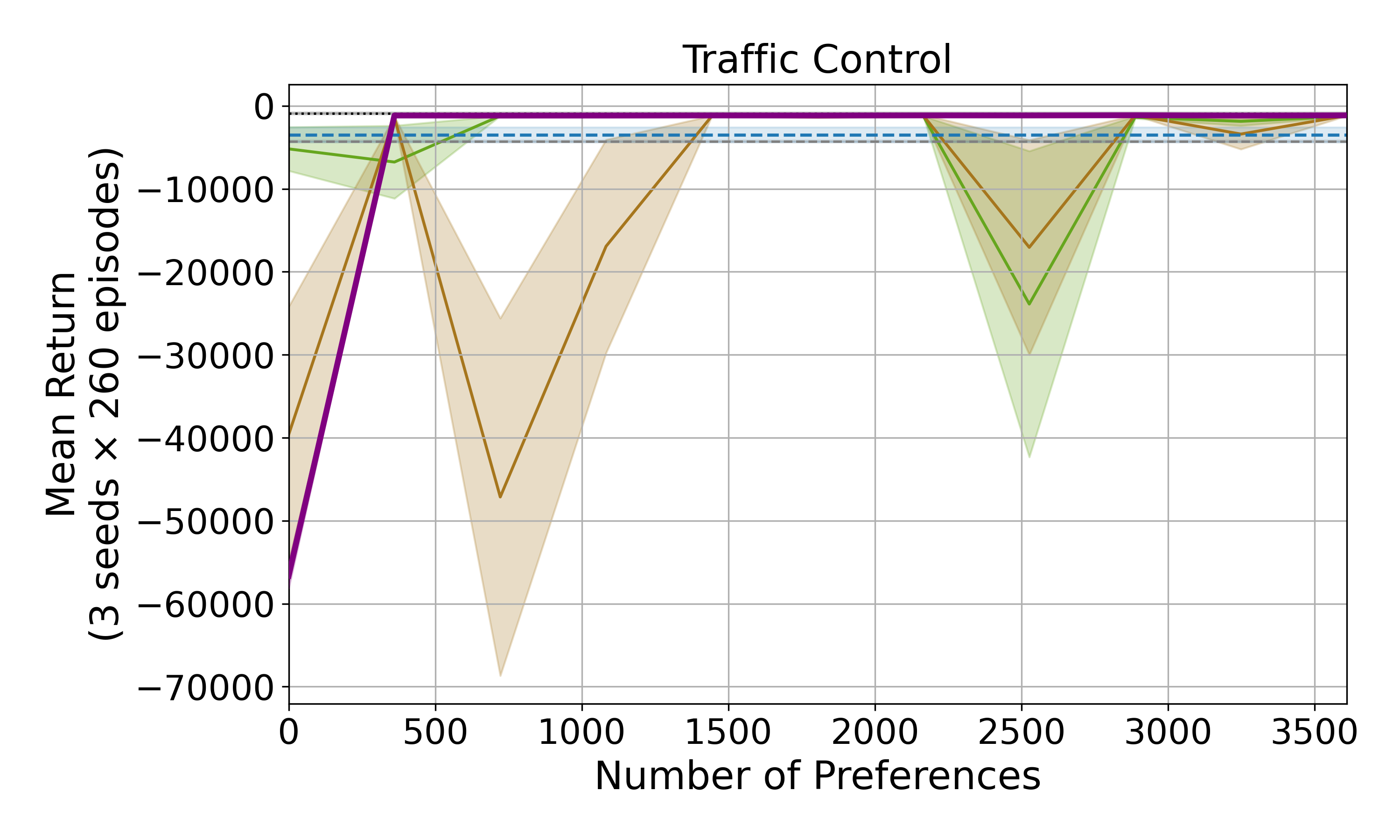}\\[0.4em]
    \includegraphics[width=0.45\textwidth]{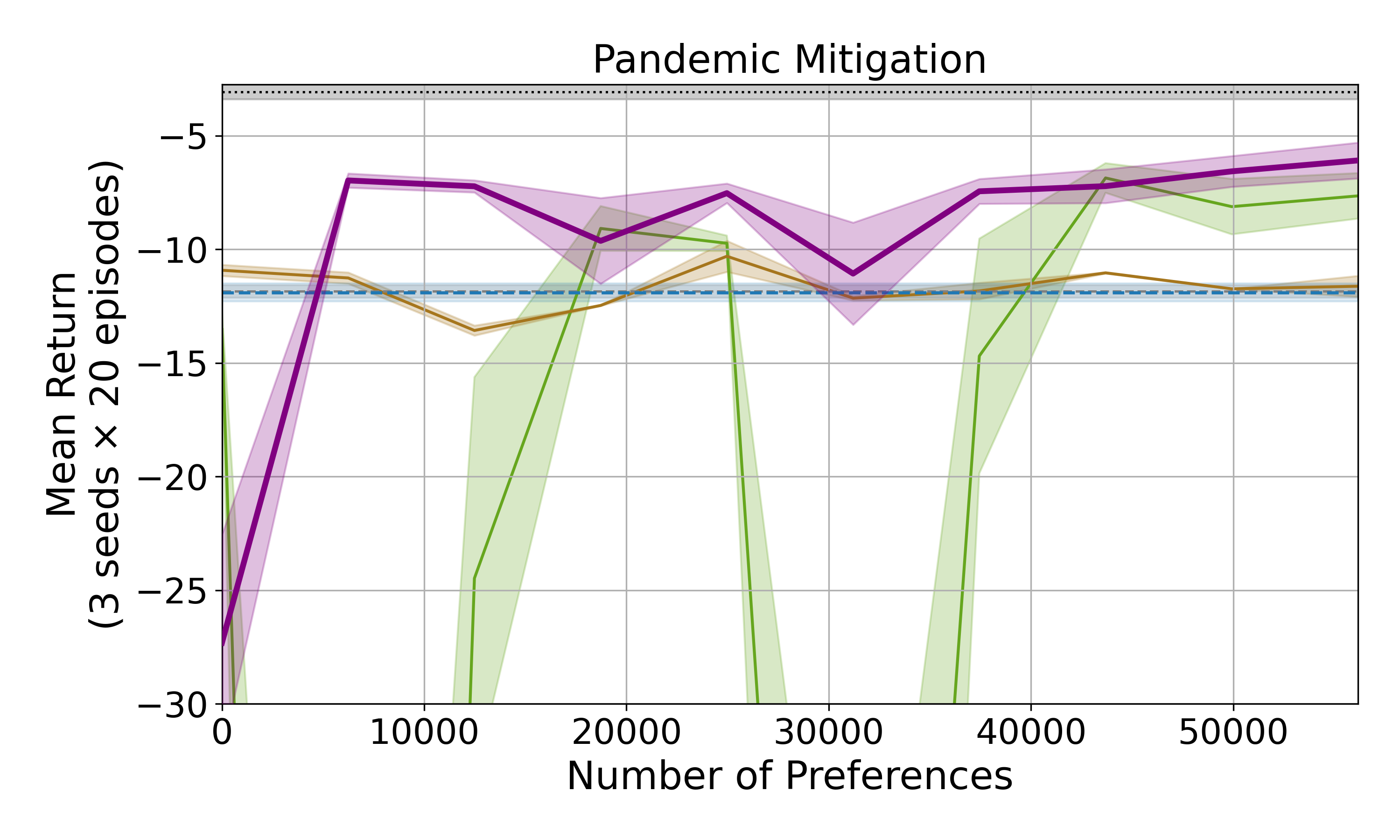}
    \includegraphics[width=0.45\textwidth]{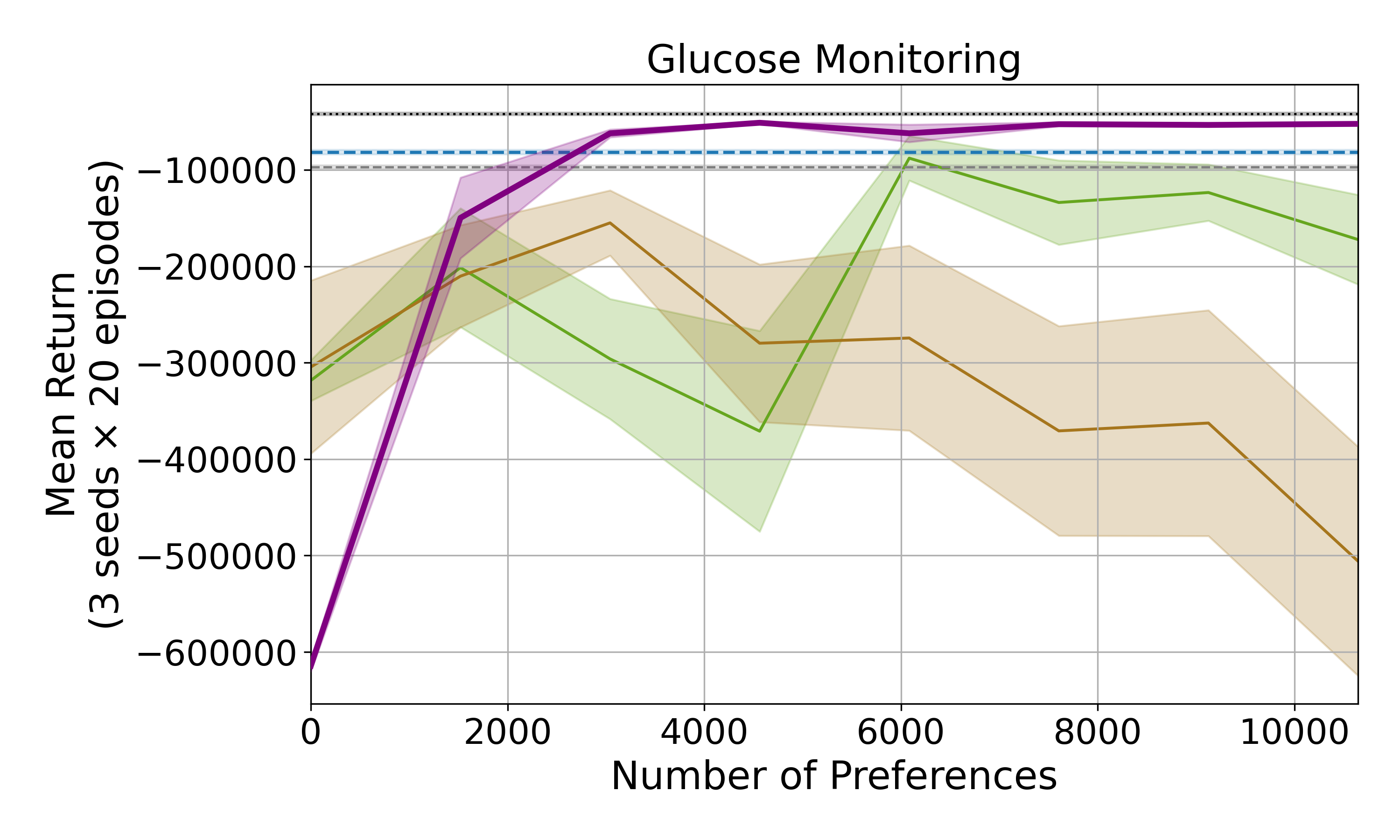}\\[0.4em]
    \includegraphics[width=0.78\textwidth]{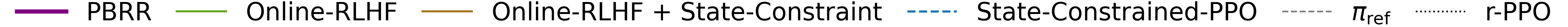}
    \subcaption{Complementing Figure \ref{fig:main_results}:Mean return under the ground-truth reward function achieved by PBRR compared to baselines that learn a reward function \textit{ab initio} from preferences, averaged over trajectories sampled from policies trained with the learned reward function across 3 random seeds. Shaded regions indicate the standard error. No scaling or clipping is applied to the plotted mean return values.}
    \label{fig:main_results_r_from_scratch_unscaled}
  \end{subfigure}

  \vspace{0.8em}

  \begin{subfigure}[t]{\textwidth}
    \centering
    \includegraphics[width=0.45\textwidth]{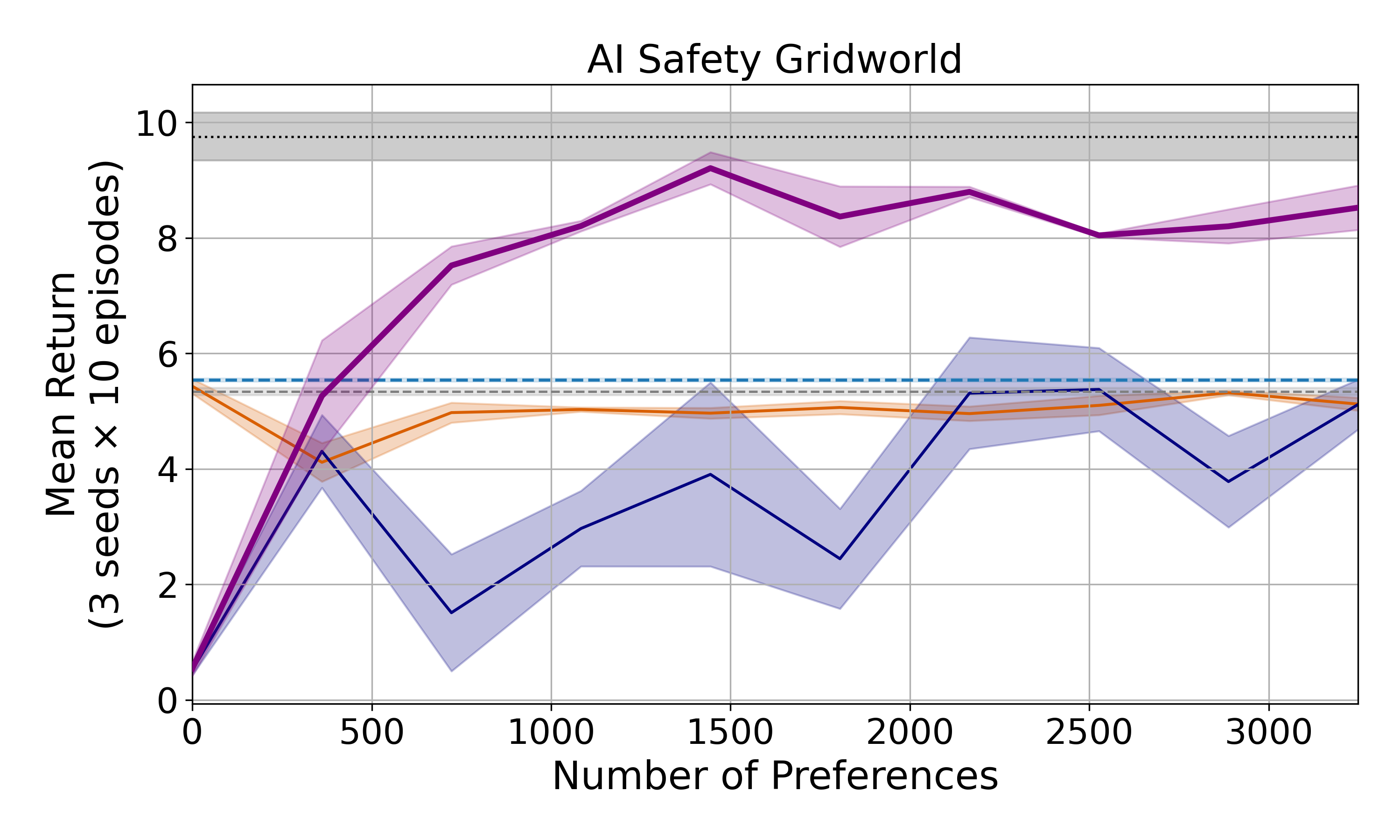}
    \includegraphics[width=0.45\textwidth]{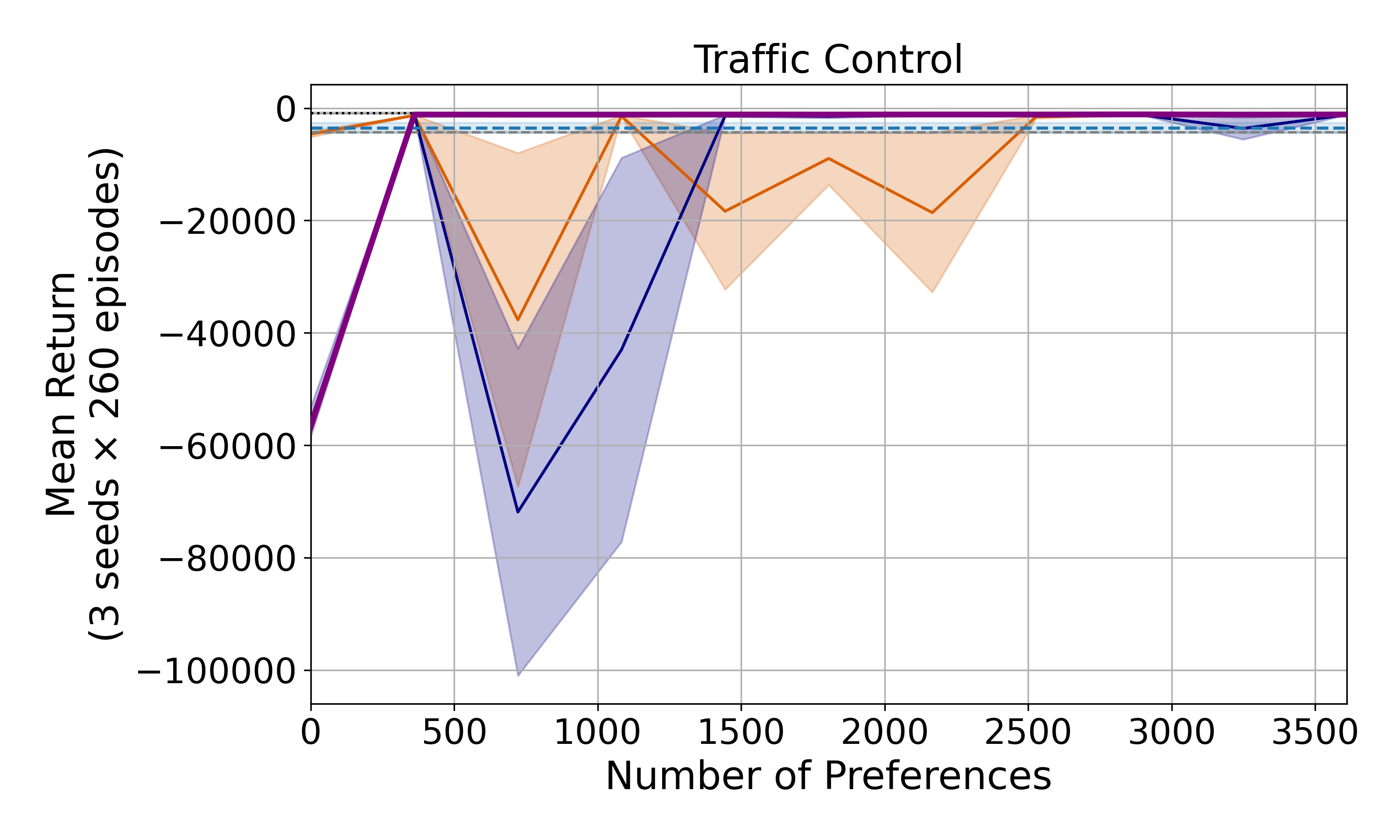}\\[0.4em]
    \includegraphics[width=0.45\textwidth]{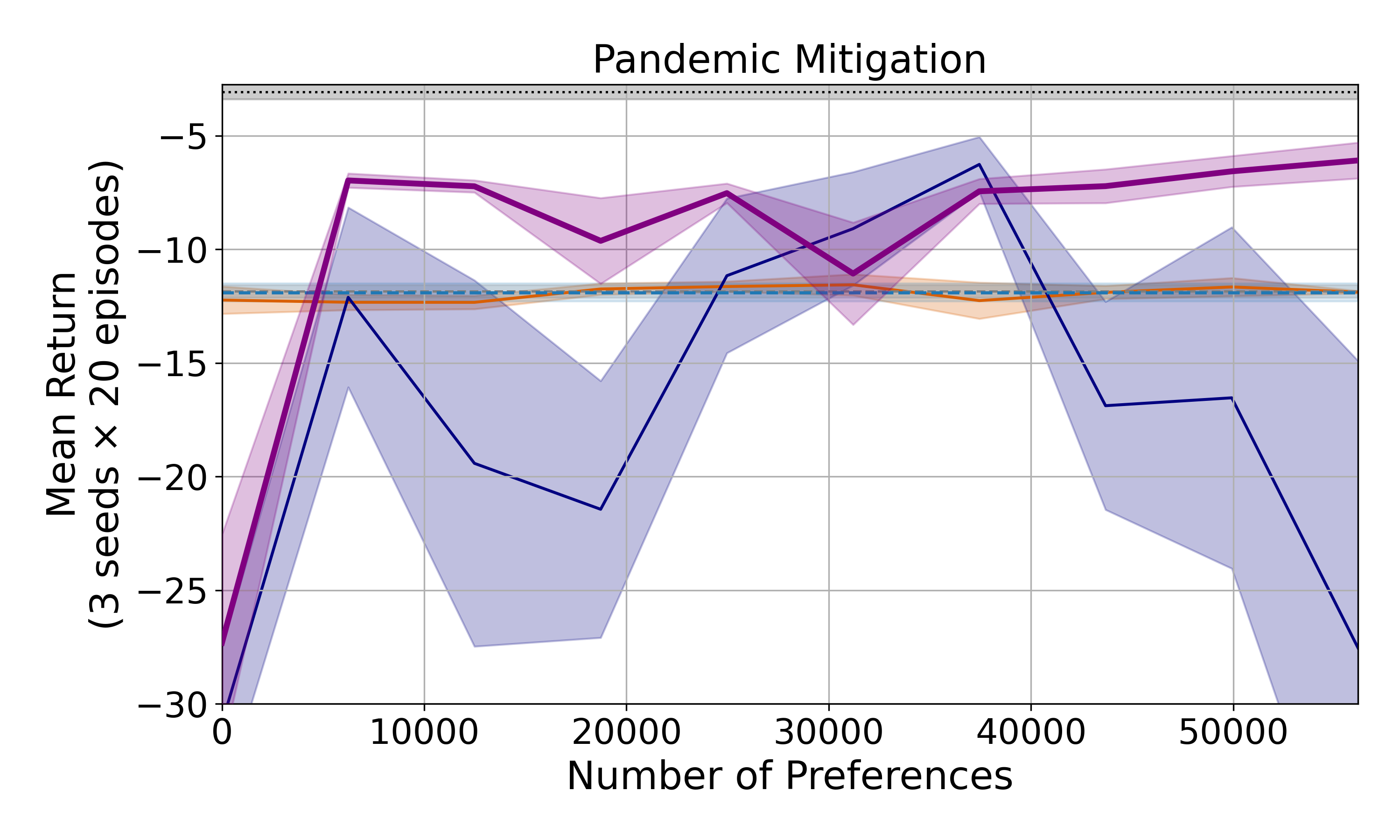}
    \includegraphics[width=0.45\textwidth]{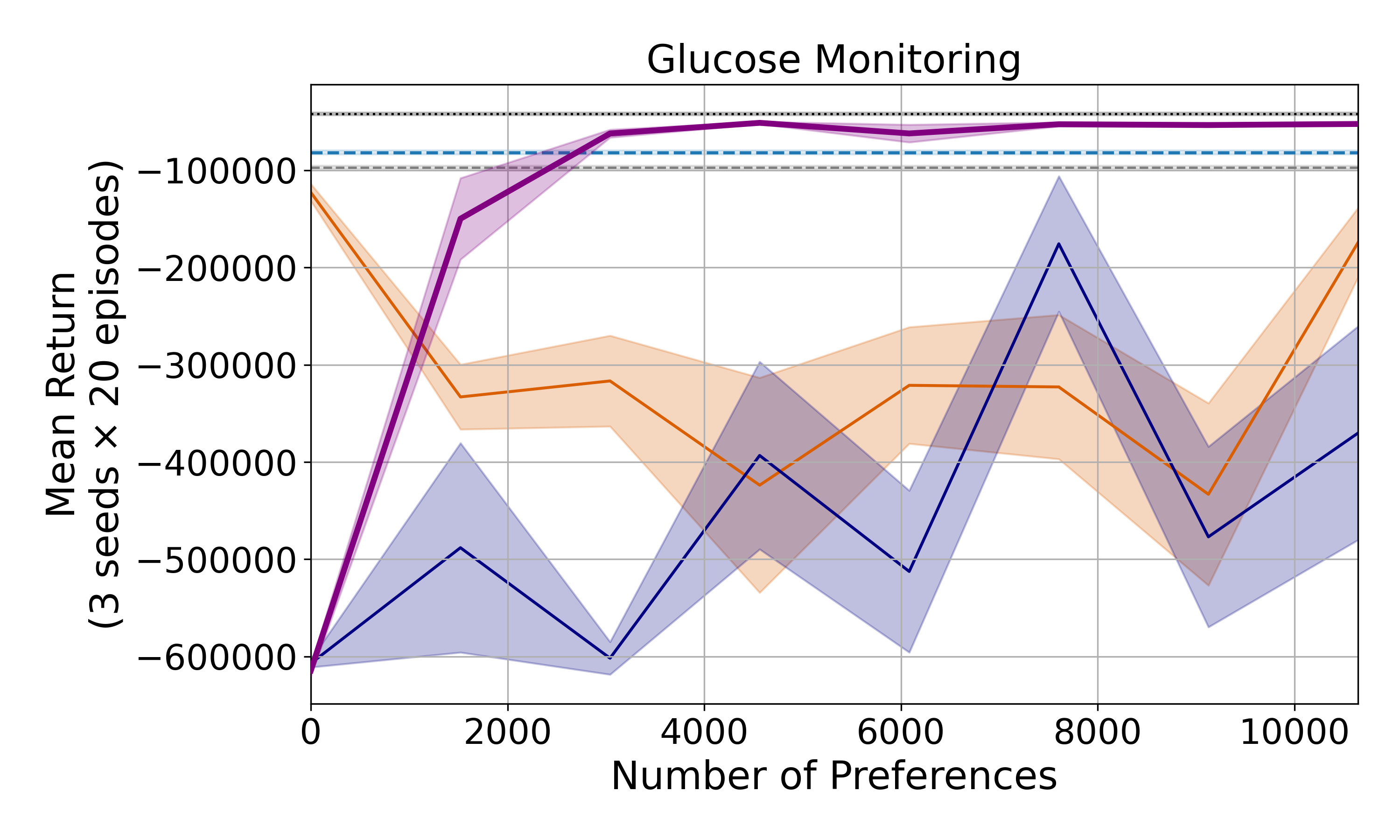}\\[0.4em]
    \includegraphics[width=0.78\textwidth]{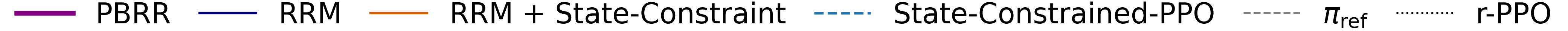}
    \subcaption{Complementing Figure \ref{fig:main_results}: Mean return under the ground-truth reward function achieved by PBRR compared to other approaches that repair the proxy reward function with preferences, averaged over trajectories sampled from policies trained with the updated proxy reward function across 3 random seeds. No scaling or clipping is applied to the plotted mean return values.}
    \label{fig:main_results_r_proxy_mod_baselines_unscaled}
  \end{subfigure}

  \caption{Unscaled mean returns, complementing the results in Figure \ref{fig:main_results}.}
\end{figure}

\FloatBarrier
\endgroup

\subsection{PBRR with a pessimistic proxy reward function}
\label{app:pessimistic_proxy}
\begin{figure}[!htbp]
    \centering
    \includegraphics[width=0.32\textwidth]{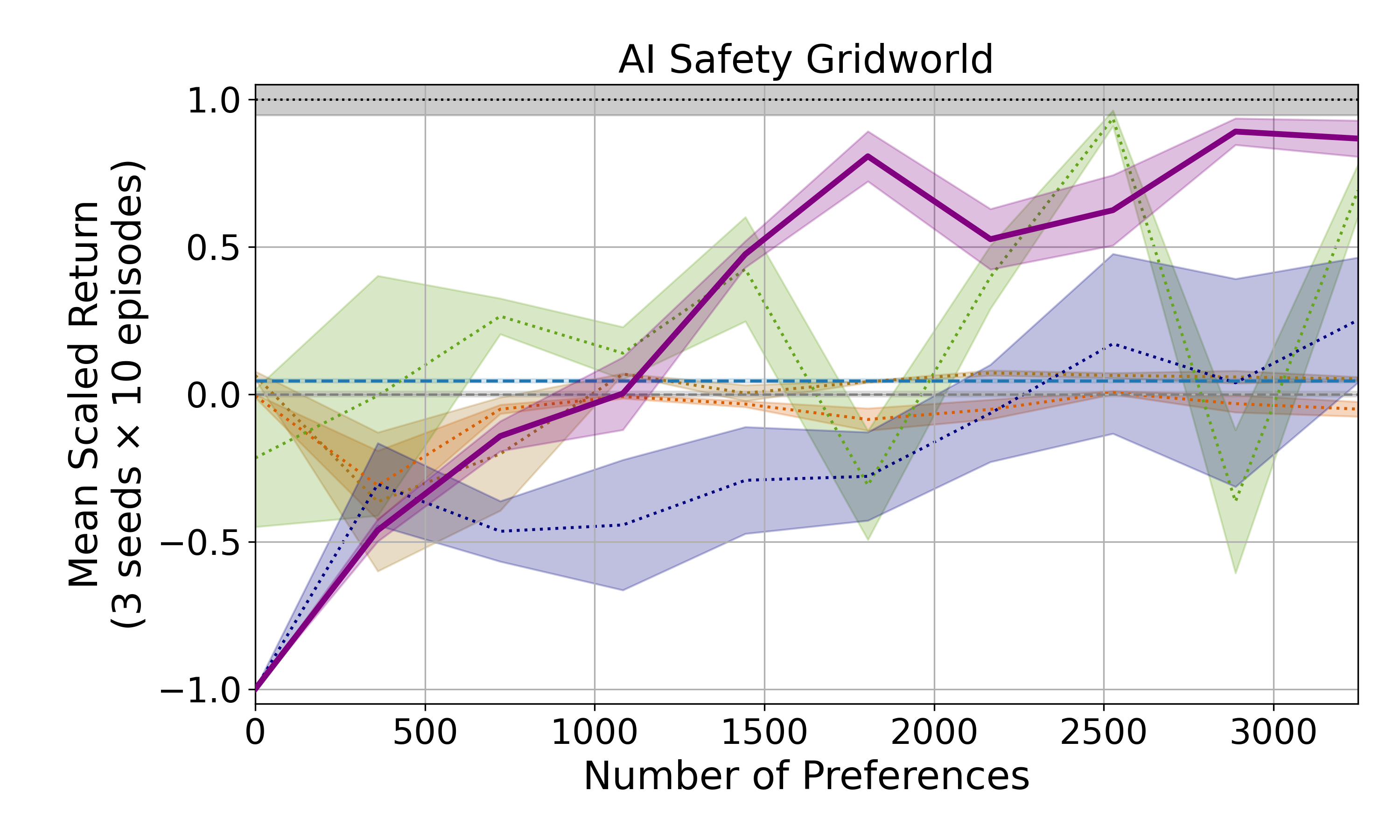}\hfill
    \includegraphics[width=0.32\textwidth]{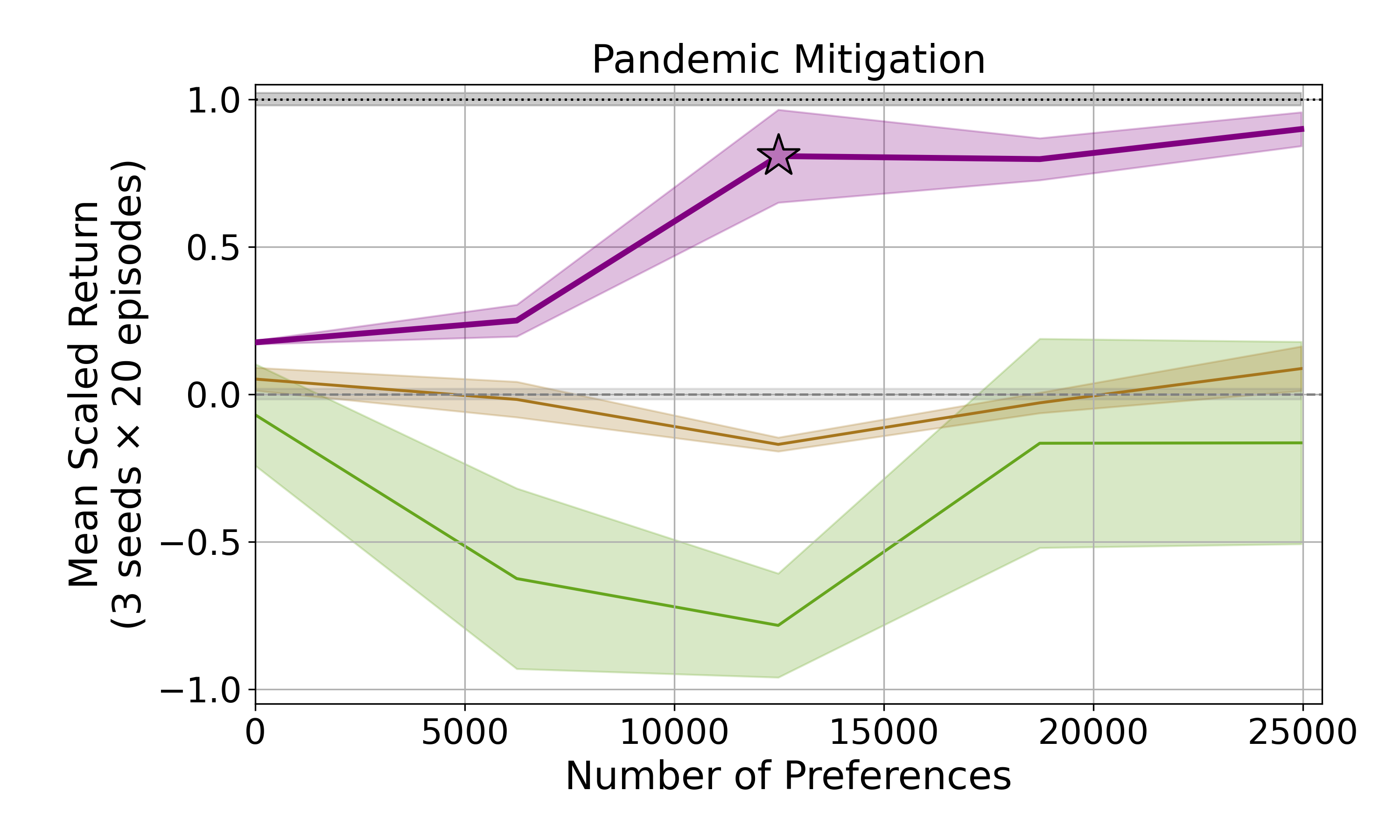}\hfill
    \includegraphics[width=0.32\textwidth]{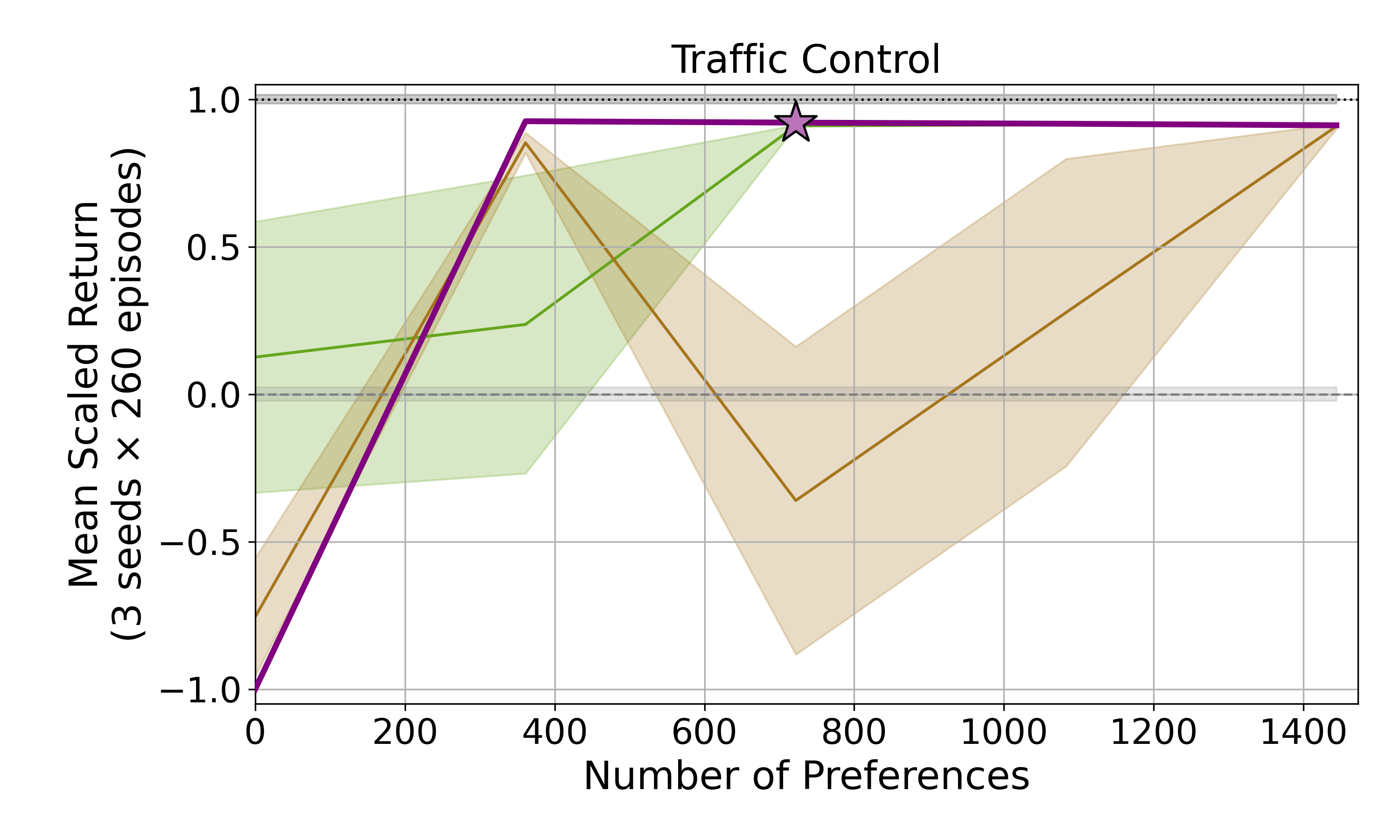}\\[4pt]
    \includegraphics[width=0.72\textwidth]{images/frh_results_glucose_legend_combined_scale=True.png}
    \caption{Mean return under the ground-truth reward function achieved by PBRR compared to the baselines from Section \ref{sec:baselines} for three environments. The proxy reward functions input to PBRR in this figure do not follow the optimism assumption leveraged by PBRR. See Figure \ref{fig:main_results} for plotting details.}
    \label{fig:pessimistic_rew_results}
\end{figure}
In Section \ref{sec:method} we discuss a key assumption behind PBRR's preference learning objective: the human-specified proxy reward function is either aligned or optimistic. Here we investigate PBRR's performance in the AI safety gridworld, Pandemic Mitigation, and Traffic Control when repairing a pessimistic proxy reward function. 

\textbf{AI Safety Gridworld} We construct a new proxy reward function that is the same as the proxy reward function used for the AI safety gridworld in Section \ref{sec:results}, except that it assigns a negative reward for visiting $\frac{3}{9}$ tomato-containing states---which the ground-truth reward function assigns a positive reward for visiting. As such, this proxy reward function breaks our optimism assumption. We plot the results in Figure \ref{fig:pessimistic_rew_results}. PBRR eventually attains near-optimal performance but requires more preferences than when repairing an optimistic proxy reward function, as assumed in Figure \ref{fig:main_results}. We note that even when the optimism assumption does not hold, PBRR still outperforms other methods for repairing the proxy reward function. 

\textbf{Pandemic Mitigation} We construct a new proxy reward function that is the same as the ground-truth reward function, except that the penalty for critical cases is $20x$ lower than in the ground-truth reward function. Therefore, this proxy reward function assigns substantially lower return to trajectories that reduce critical cases but, for example, have higher lockdown restrictions. We plot the results when using PBRR to repair this reward function in Figure \ref{fig:pessimistic_rew_results}, showing that PBRR continues to be an effective strategy to repair pessimistic reward functions and still is more data-efficient than learning a reward function from scratch.  

\textbf{Traffic Control} We construct a new proxy reward function that is the same as the ground-truth reward function except that we remove the reward component that assigns a higher reward for increased vehicle speed, i.e., the primary measure of traffic flow. This proxy reward function therefore assigns low return to trajectories where all vehicles drive quickly through the highway with no traffic. Figure \ref{fig:pessimistic_rew_results} shows that PBRR effectively repairs this reward function after a single round of preference elicitation. 

\subsection{PBRR with a randomly initialized reference policy}
\label{app:pbrr_random_ref_pol}

\begin{figure}[h]
\vspace{-2mm}
    \centering
    \includegraphics[width=0.4\textwidth]{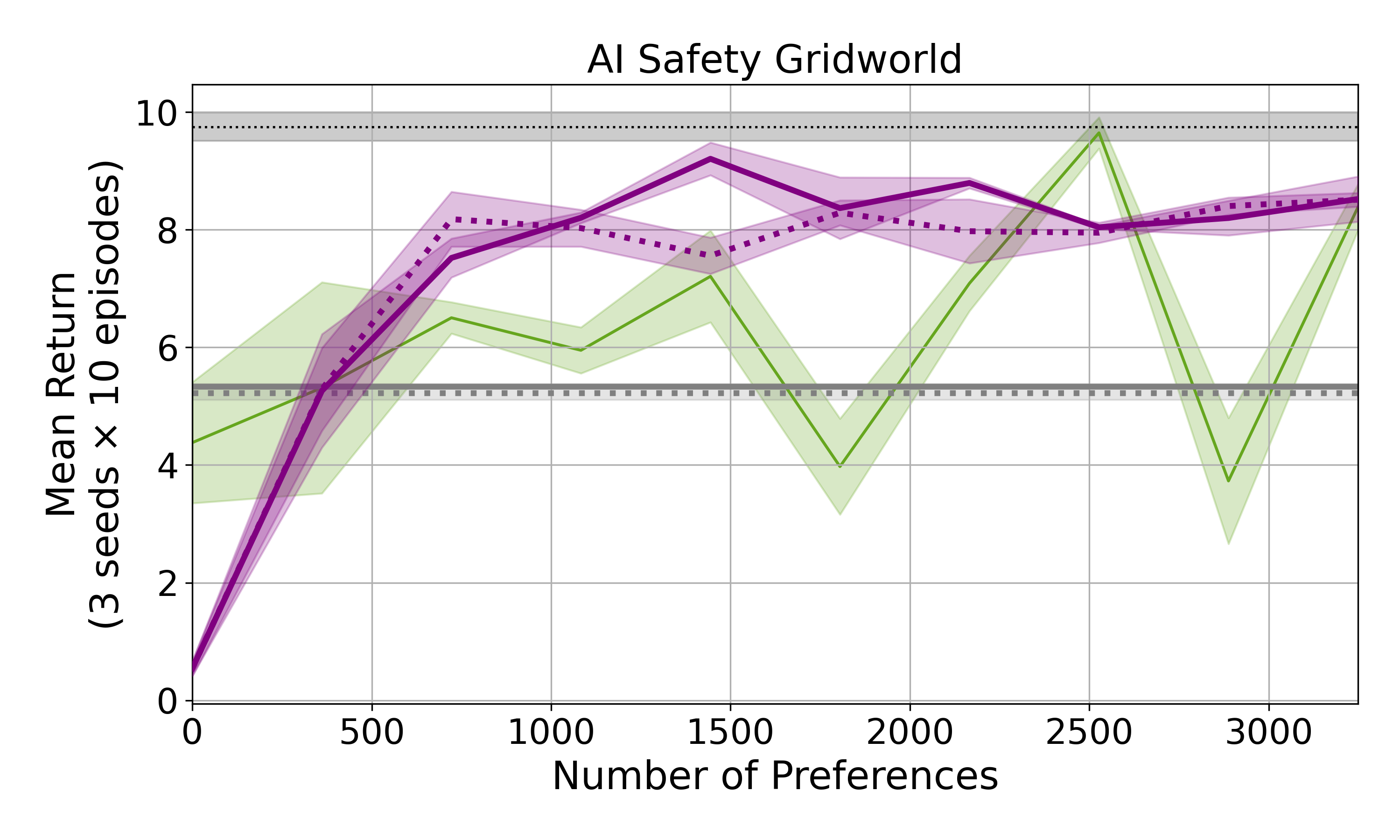}
    \includegraphics[width=0.4\textwidth]{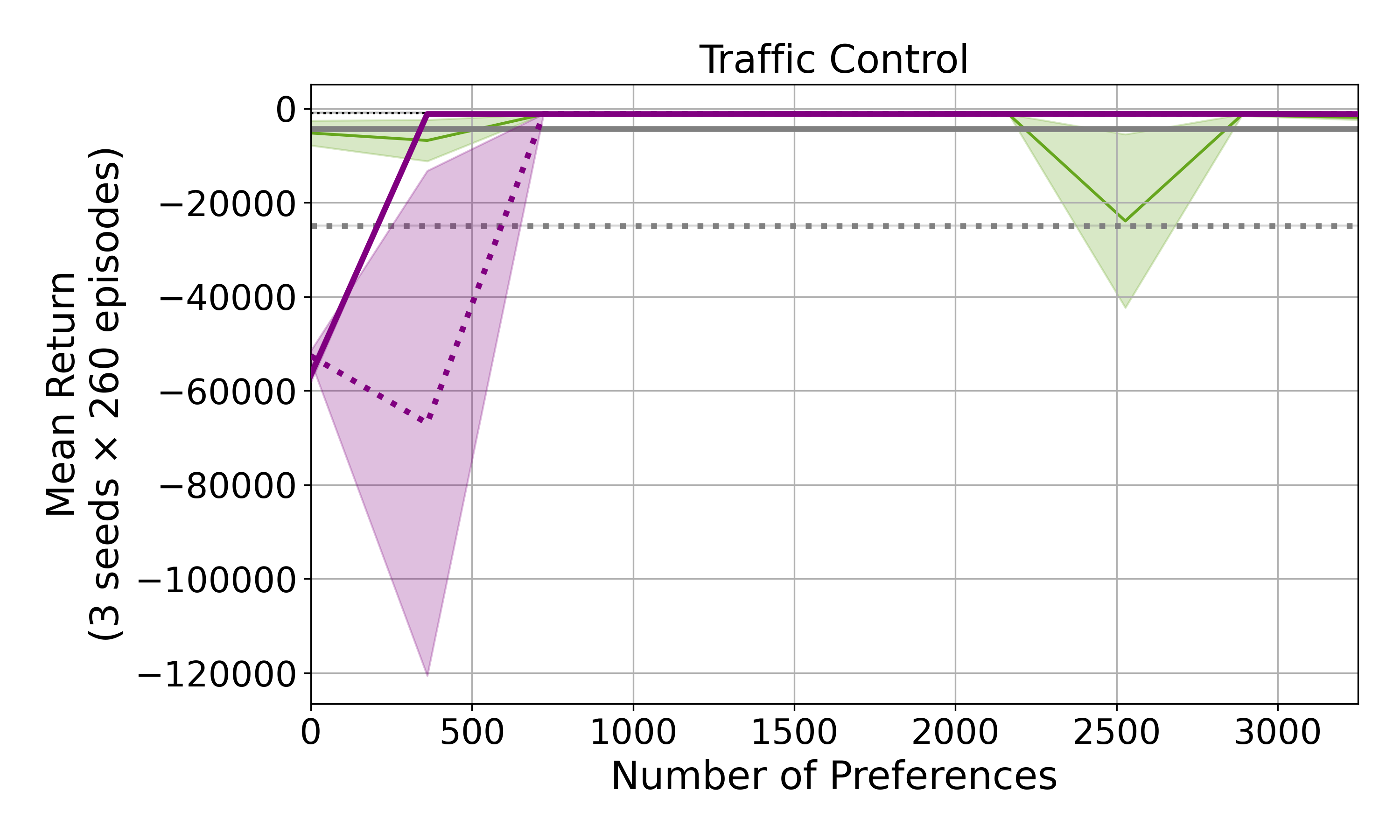}\\[0.5em]
    \includegraphics[width=0.4\textwidth]{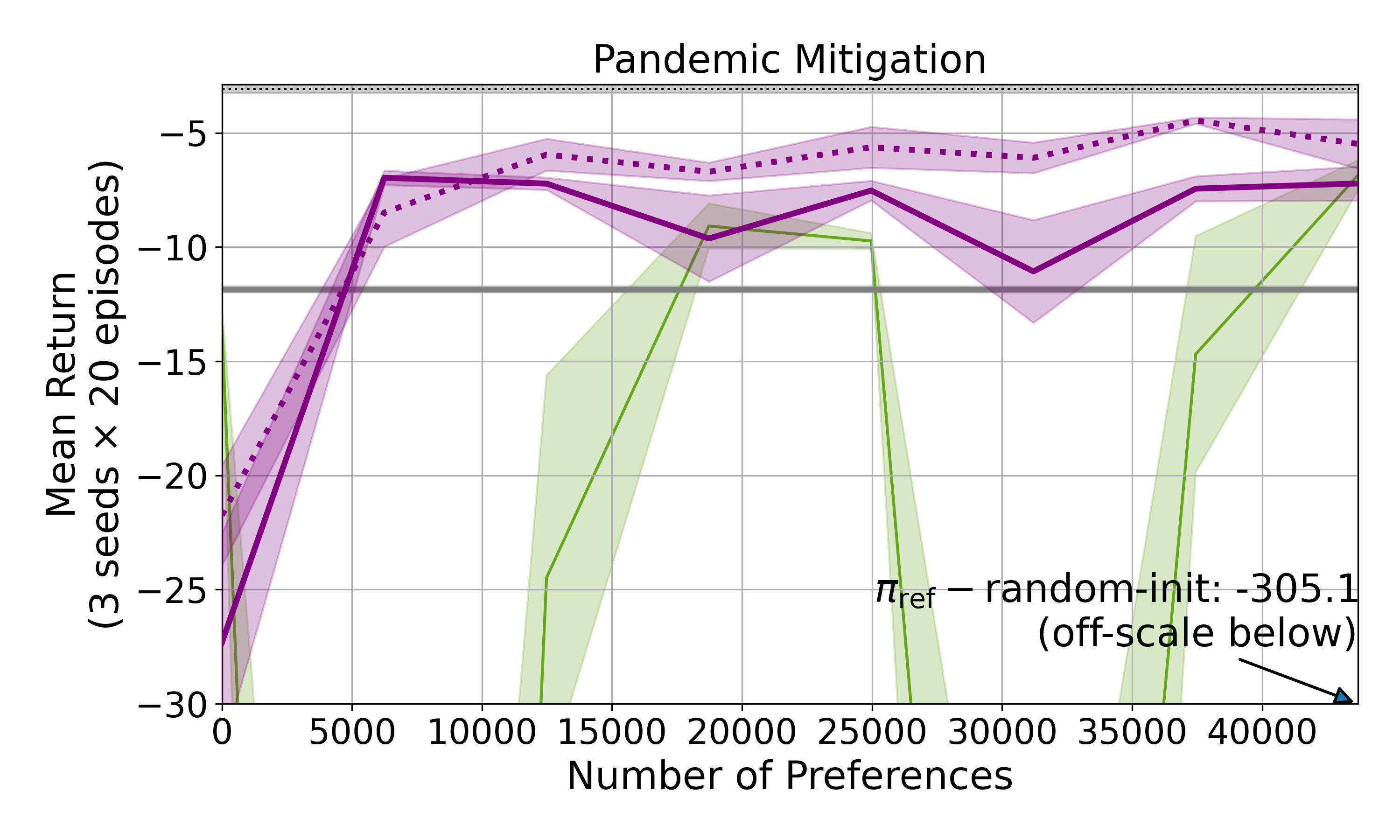}
    \includegraphics[width=0.4\textwidth]{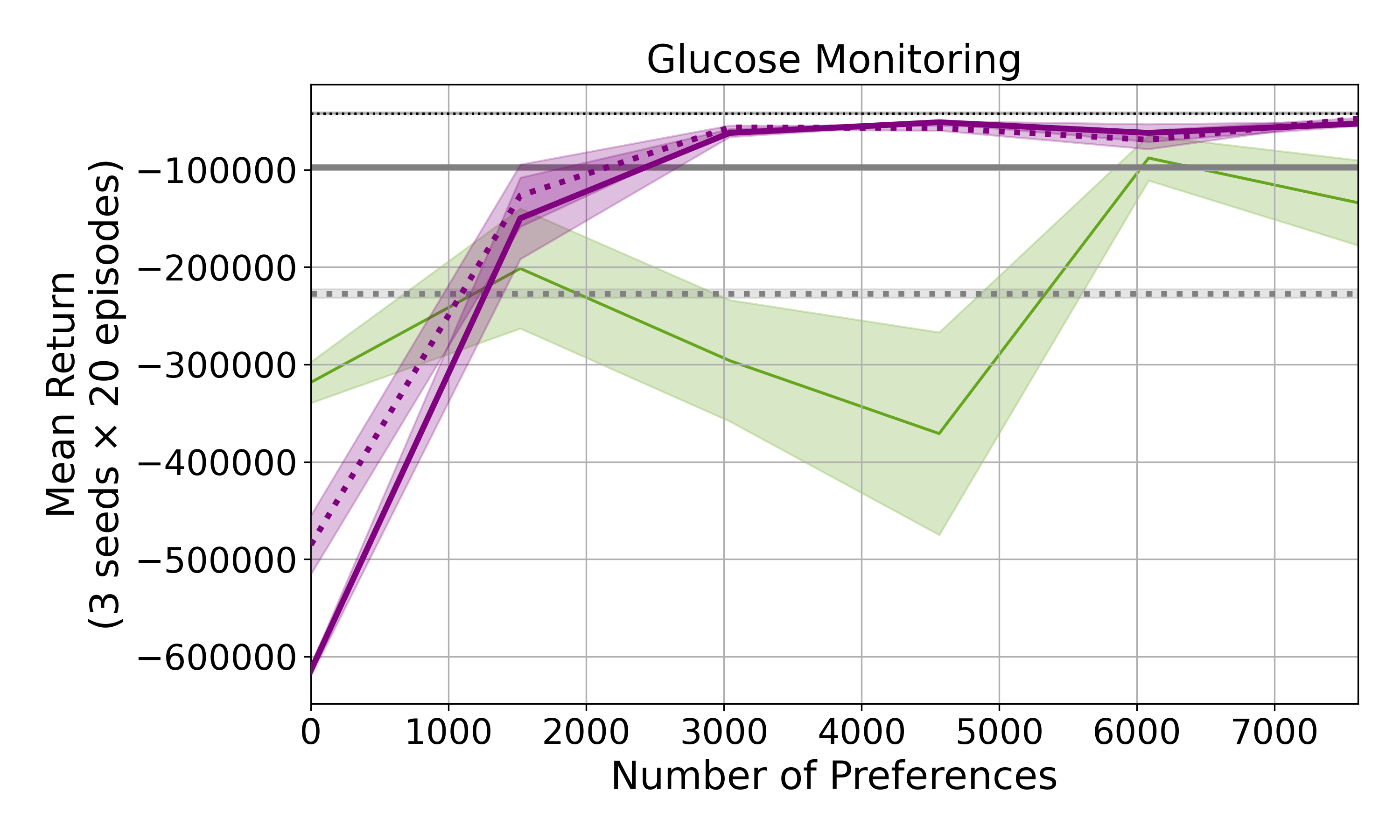}
    \includegraphics[width=0.8\textwidth]{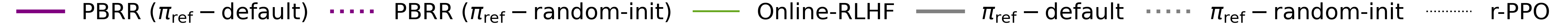}
    \caption{Mean return under the ground-truth reward function achieved by PBRR with the default reference policy used for all other experiments (PBRR ($\pi_{\text{ref}}-\text{default}$)), compared to (i) PBRR using a randomly initialized reference policy (PBRR ($\pi_{\text{ref}}-\text{random-init}$)), and (ii) the Online-RLHF method. The plot for the Pandemic Mitigation environment is truncated to make it easier to differentiate performance between various methods; the full plot is shown in Figure \ref{fig:pbrr_pandemc_unclipped_ref_abalation}.}
    \label{fig:pbrr_ref_abalation}
\vspace{-3mm}
\end{figure}

PBRR requires a reference policy for preference elicitation. For all prior experiments, unless otherwise stated, we use a realistic reference policy that a human could provide such as one trained with a handful of human demonstrations or from imperfect hand-written rules. Here, we investigate PBRR's performance when using a randomly initialized reference policy instead. Figure \ref{fig:pbrr_ref_abalation} shows that performance is largely unchanged; the quality of the reference policy used by PBRR is not judged by its performance, but rather its coverage of the state-action space relative to the policy induced by the proxy reward function. These results imply that for many tasks, a sufficient reference policy should always be available by simply using a randomly initialized policy. 

\begin{figure}[H]
\centering
\begin{minipage}{0.4\textwidth}
    \centering
    \includegraphics[width=\linewidth]{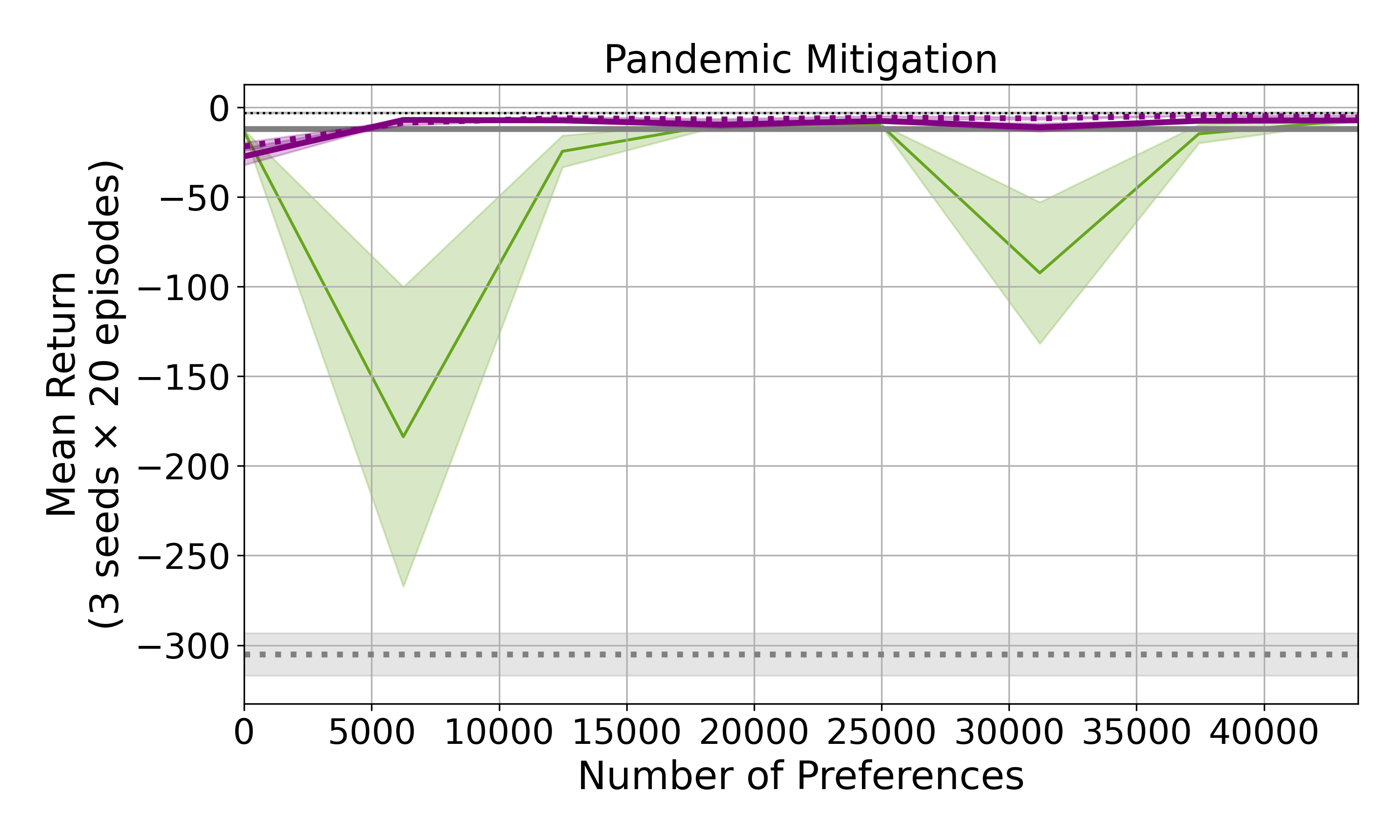}
\end{minipage}
\hfill
\begin{minipage}{0.54\textwidth}
    \caption{Mean return under the ground-truth reward function achieved by PBRR with the default reference policy used for all other experiments (PBRR ($\pi_{\text{ref}}-\text{default}$)), compared to (i) PBRR using a randomly initialized reference policy (PBRR ($\pi_{\text{ref}}-\text{random-init}$)), and (ii) the Online-RLHF method. For easier visual comparison, we truncate the plot for the Pandemic Mitigation environment in Figure~\ref{fig:pbrr_ref_abalation}.}
    \label{fig:pbrr_pandemc_unclipped_ref_abalation}
\end{minipage}
\end{figure}

\subsection{PBRR with preferences that vary in noise}
\label{app:pref_noise}
\begin{figure}[!htbp]
    \centering
    \includegraphics[width=0.54\textwidth]{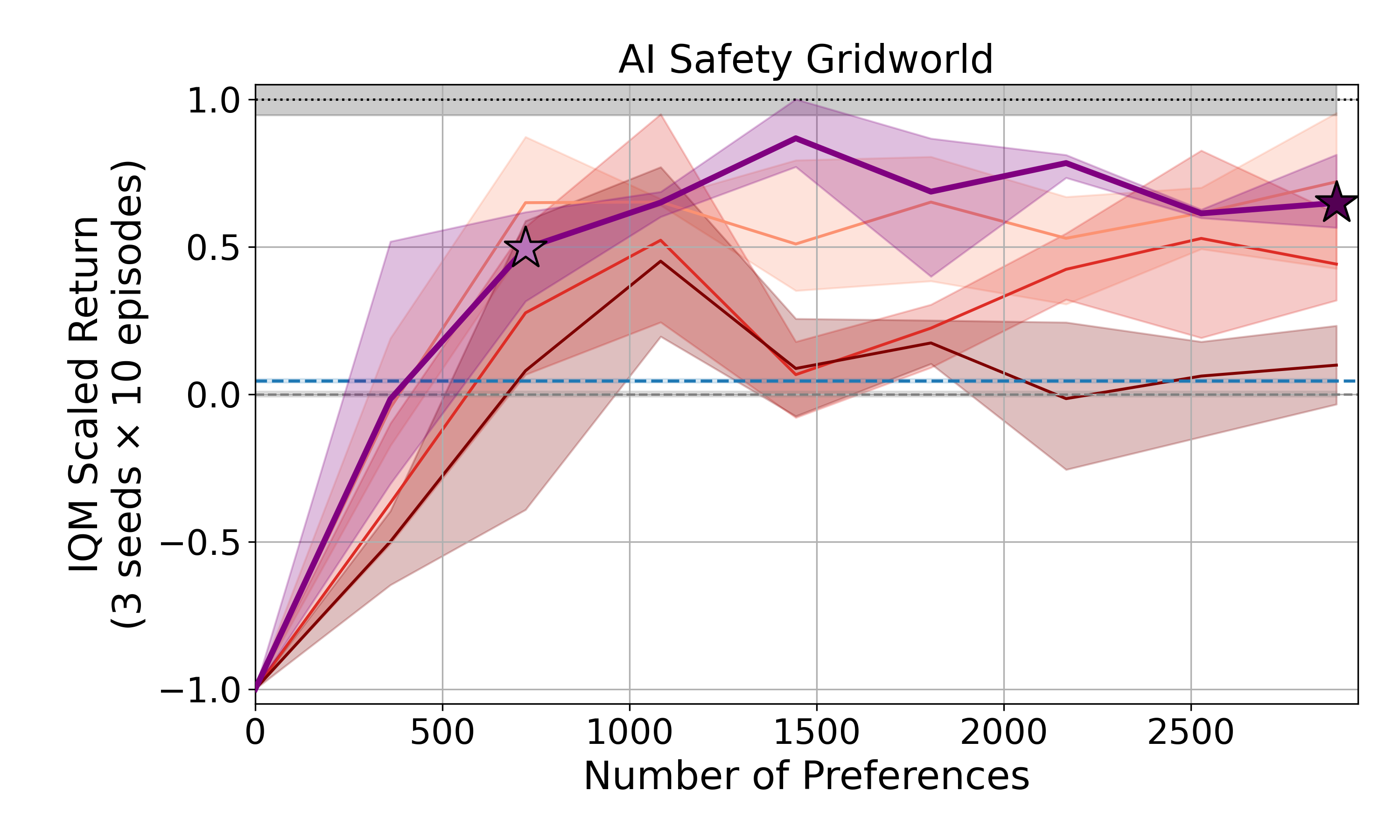}\hfill
    \includegraphics[width=0.72\textwidth]{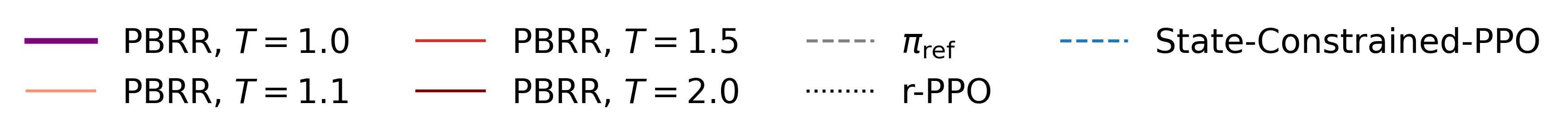}
    \caption{Mean return under the ground-truth reward function achieved by PBRR
    as the Bradley--Terry temperature $T$ used to label preferences increases in the AI Safety Gridworld
    See Figure~\ref{fig:main_results} for plotting details.}
    \label{fig:temperature}
\end{figure}

To simulate human preferences, we assume that preferences are generated by the
Bradley--Terry preference model of Section~\ref{sec:background} under the
ground-truth reward function. To vary the noisiness of the simulated labels, we
introduce a temperature parameter $T > 0$:
\begin{equation}
    P(\tau_1 \succ \tau_2 \mid r, T) = \sigma\!\left(\frac{r(\tau_1) - r(\tau_2)}{T}\right),
\end{equation}
which recovers the model used elsewhere in the paper at $T = 1$. As $T$
increases, preferences become less sensitive to the difference in return and
therefore noisier, approaching uniformly random labels as $T \to \infty$.
Doubling the temperature is equivalent to halving every return difference: the
labeler behaves as though the two trajectories were half as far apart in
quality, so a pair that was labeled correctly $90\%$ of the time at $T = 1$
is labeled correctly only $75\%$ of the time at $T = 2$. Here we analyze how
the performance of PBRR changes as $T$ increases in the AI Safety Gridworld. As illustrated in
Figure~\ref{fig:temperature}, PBRR's performance remains unaffected by a slight
increase in temperature (e.g., from $1.0$ to $1.1$) and degrades, as expected,
when the temperature is doubled.

\subsection{PBRR with a random reward function}

\begin{figure}[!htbp]
    \centering
    \includegraphics[width=0.54\textwidth]{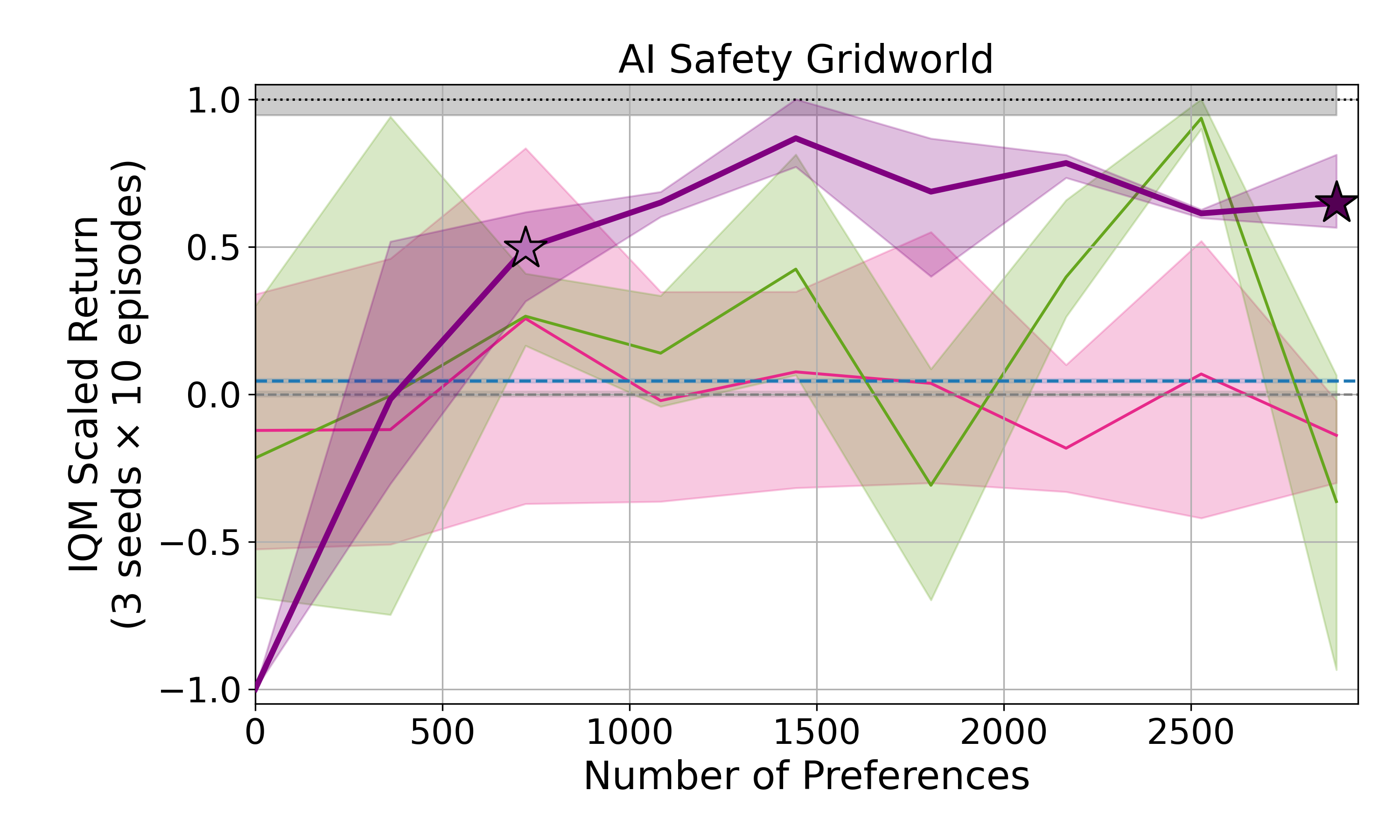}\hfill
    \includegraphics[width=0.72\textwidth]{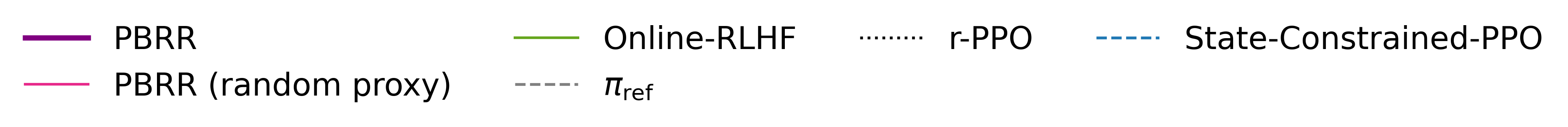}
    \caption{Mean return under the ground-truth reward function achieved by PBRR with a proxy reward function that is parameterized as a random neural network, compared to the proxy reward function assumed throughout this work from \cite{pan2022effects}. See Figure~\ref{fig:main_results} for plotting details.}
    \label{fig:random_proxy}
\end{figure}

Section~\ref{sec:main_results} shows that PBRR effectively repairs the proxy
reward functions from prior work across four environments, and
Appendix~\ref{app:pessimistic_proxy} shows that it also repairs proxy reward
functions for these environments that violate our optimism assumption. Both
settings share a common feature: the input reward function, however misspecified, encodes some information about the designer's objective. Here we probe the opposite extreme, repairing a reward function parameterized as a randomly initialized neural network, i.e., one that carries no information about the task at all. As shown in Figure~\ref{fig:random_proxy}, PBRR does not recover near-optimal performance in the AI Safety Gridworld under these conditions. This is consistent with the premise of our method: PBRR is designed for settings in which a human reward designer can specify an imperfect but informative reward function, which we argue is common in practice. When no such reward function is available, learning a reward function \emph{ab initio}---for instance with
Online-RLHF---is likely the better choice.

\subsection{Statistical significance of our results}
  \label{app:significance}

  The results in Figure~\ref{fig:main_results} were originally computed over $3$ random seeds. To
  establish that the comparisons we draw are not an artifact of seed noise, we re-ran each of the five
  methods compared in Figure~\ref{fig:main_results}---PBRR, Online-RLHF, Online-State-Constraint, RRM,
  and RRM~+~State-Constraint---for $6$ additional random seeds in each of the four environments. A seed indexes a
  completely independent execution of the corresponding algorithm: it determines policy and
  reward-model initialization, environment stochasticity, the trajectories sampled for preference
  elicitation, and the sampled preference labels. 
  
  We aggregate across seeds with the \emph{interquartile mean} (IQM): writing
  $x_{(1),t}\le\dots\le x_{(n),t}$ for the seed values at iteration $t$ sorted in ascending order and
  $k=\lfloor n/4 \rfloor$,
  \begin{equation}
    \mathrm{IQM}_t \;=\; \frac{1}{n-2k}\sum_{j=k+1}^{n-k} x_{(j),t},
  \end{equation}
  so with $n=9$ we discard the two best and two worst seeds and average the middle five. We report the
  IQM rather than the sample mean following \citet{agarwal2021deep}, because the across-seed distribution
  of returns is heavy-tailed or bimodal in several of our settings.

\begin{figure}[!htbp]
  \centering
  \includegraphics[width=0.48\textwidth]{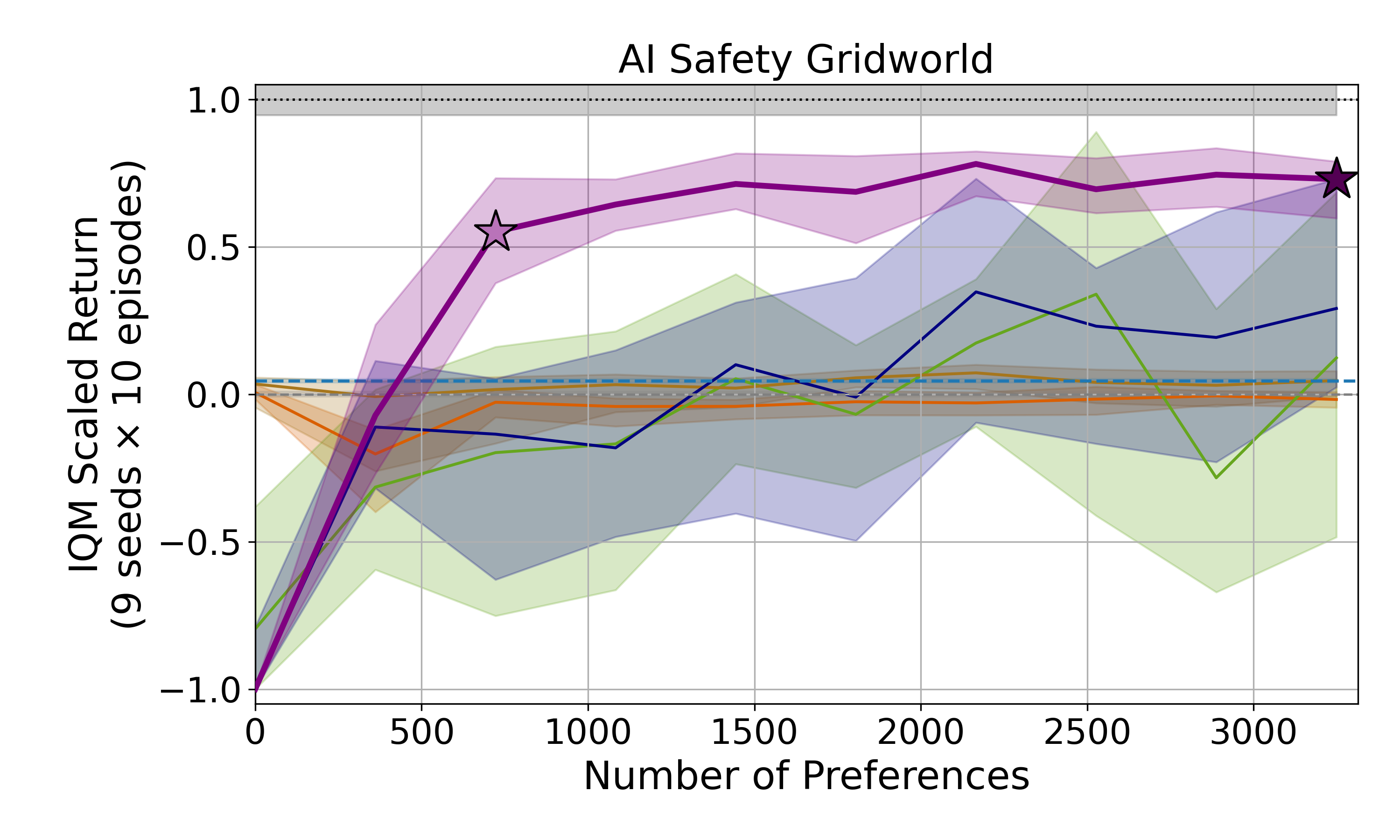}
    \includegraphics[width=0.48\textwidth]{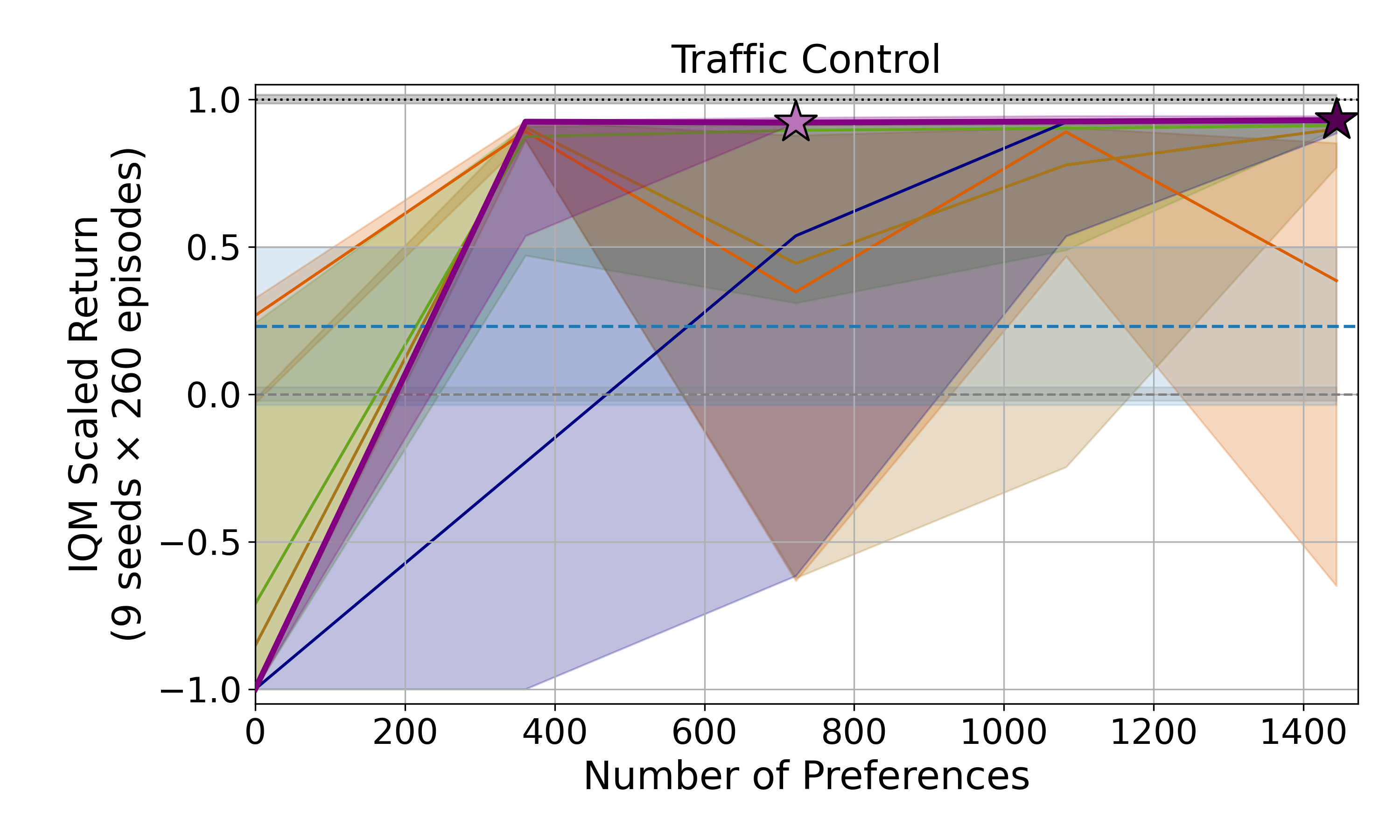}\\[0.6em]
    \includegraphics[width=0.48\textwidth]{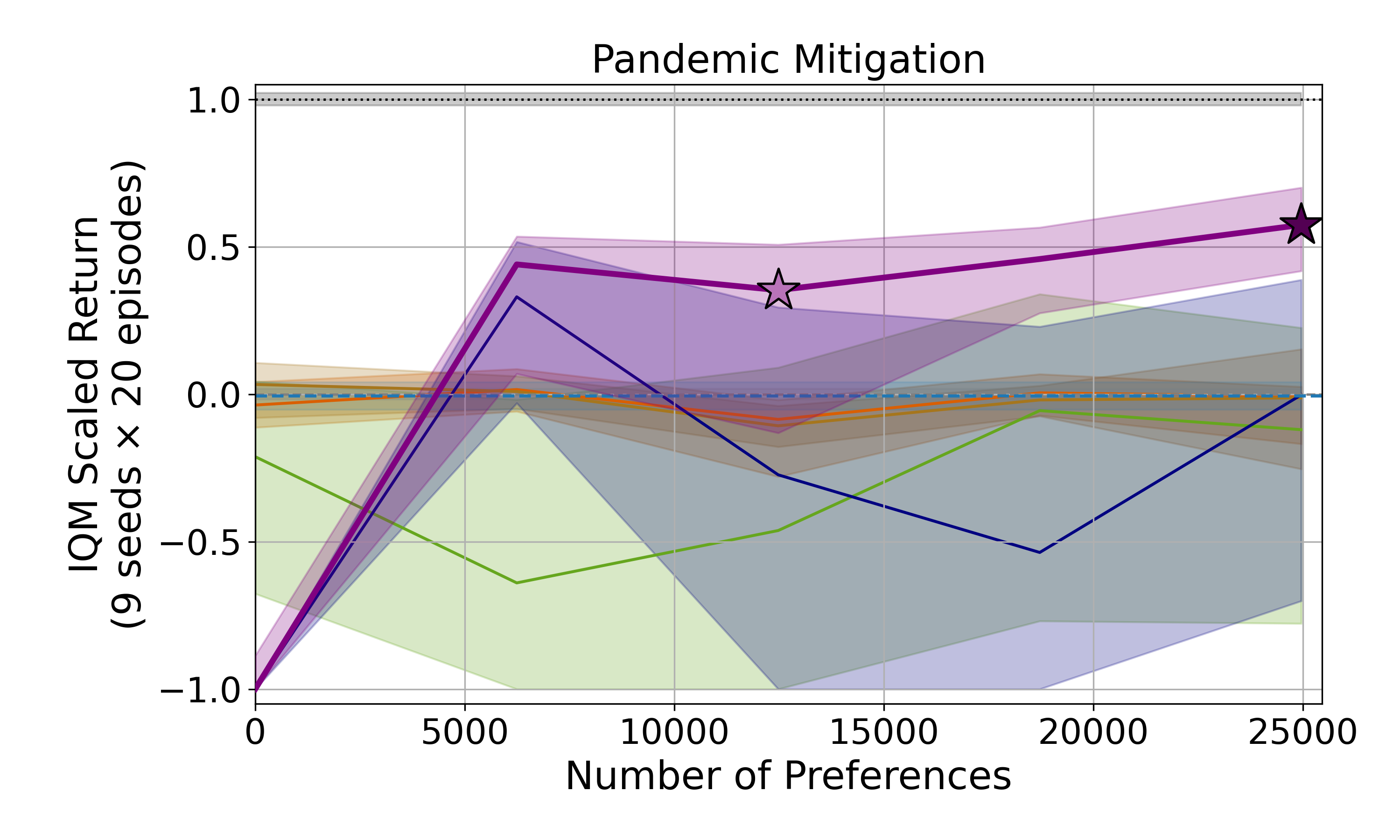}
    \includegraphics[width=0.48\textwidth]{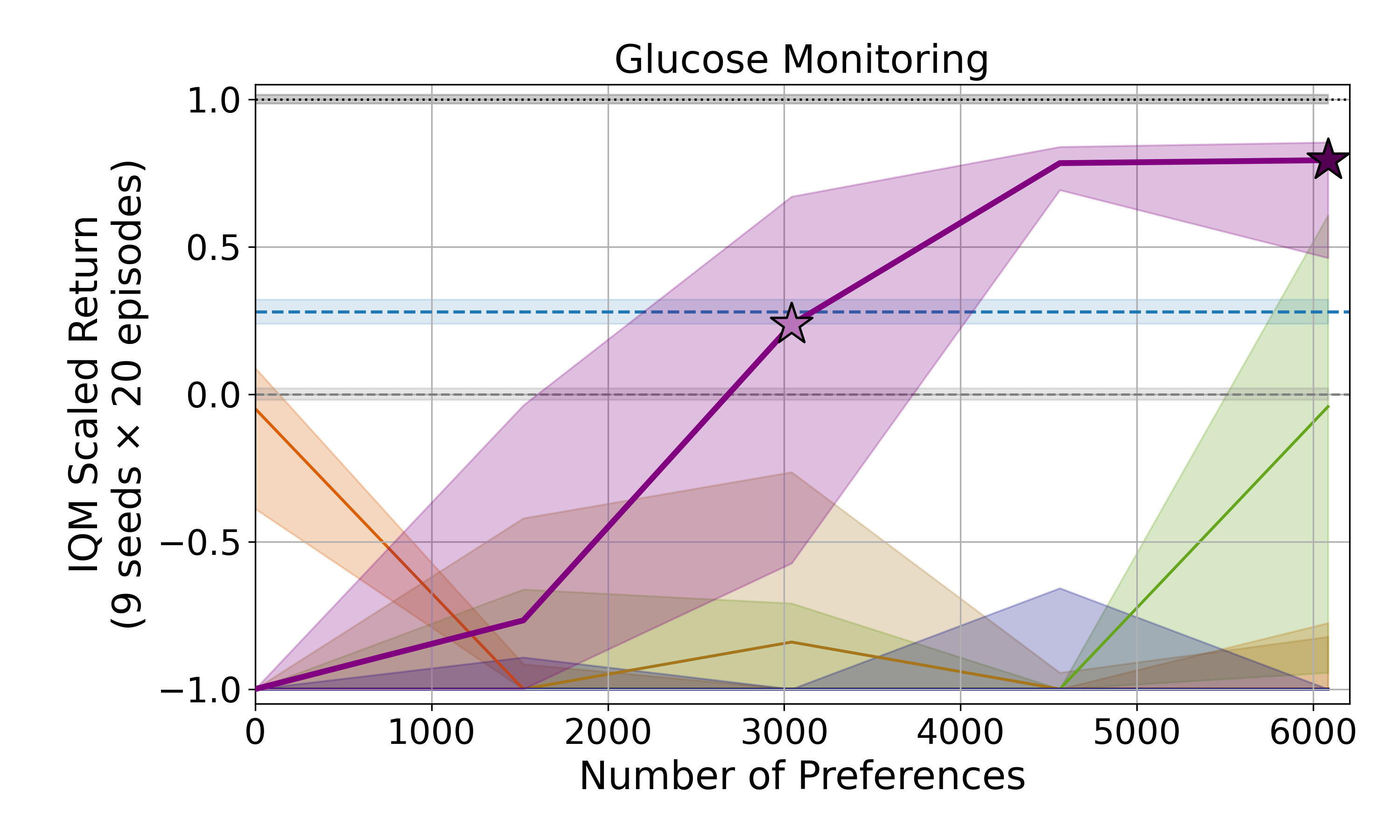}\\[0.6em]
    \includegraphics[width=0.85\textwidth]{images/frh_results_glucose_legend_scaled_abalation_scaled=True.png}
    \caption{Complementing Figure \ref{fig:main_results}: Mean return under the ground-truth reward function achieved by PBRR compared to the baselines from Section \ref{sec:baselines}, averaged over 9 random seeds. Shaded regions indicate the $95\%$ confidence-intervals, computed as described in Appendix \ref{app:significance}}
    \label{fig:main_fig_sig}
\end{figure}

Figure \ref{fig:main_fig_sig} plots results to complement Figure \ref{fig:main_results}. For all environments except for AI Safety Gridworld, due to computational limitations, we evaluate these methods with the additional seeds for fewer iterations than in Figure \ref{fig:main_results}. Shaded regions are $95\%$ percentile bootstrap confidence intervals
  for the IQM, where we draw $B=50{,}000$ bootstrap replicates. 

PBRR attains a significant improvement over every baseline in all environments except for Traffic Control in at least one iteration, and in no environment does any baseline significantly outperform PBRR once preference elicitation has begun. Moreover, the comparisons that do not reach significance still favor PBRR: from the second round of preference elicitation onward, PBRR's interquartile mean return is at least as high as that of every baseline at every iteration in all four environments, so the intervals that overlap do so around a higher central estimate for PBRR.

\section{Empirical check of the regret bound from Theorem \ref{thm:known_trans_model_regret_bound}}
  \label{app:regret-check}
  
  We empirically test the $\tilde O(\sqrt{T})$ regret guarantee of Theorem \ref{thm:known_trans_model_regret_bound} in a setting with known dynamics, a linear reward function, and preferences that follow the Bradley-Terry model. 

  \subsection{Environments}
  \label{app:regret-check:mdps}

  We randomly generate three tabular MDPs. Each has $|\mathcal S| = 6$ states, $|\mathcal A| = 3$ actions, horizon
  $H = 4$, a fixed initial state $s_0$, and no discounting. Each transition distribution
  $P(\cdot \mid s, a)$ is drawn independently from $\mathrm{Dirichlet}(\mathbf 1)$.
  The policy class $\Pi$ is the set of all deterministic stationary policies. All six
  states are reachable within the horizon, so $|\Pi| = 3^6 = 729$. Because the
  dynamics are known, the feature expectations $\phi(\pi)$ are computed exactly by
  forward propagation.

  \paragraph{Ground-truth reward function.} Each feature is a one-hot encoding of a state-action transition. The
  ground-truth reward function is $r^\star(s,a) = \phi(s,a)^\top w^\star$, with $w^\star$ drawn uniformly
  from the sphere of radius $W = \|w^\star\| = 1.5$. Trajectory features therefore
  satisfy $\|\phi(\tau)\| \le B = H = 4$ (Assumption \ref{asm:bounded_features}), which was checked on every
  sampled trajectory.

  \paragraph{Proxy reward function} We construct a proxy reward function
  $w_{\mathrm{proxy}} = w^\star + b\,\mathbf 1_K$, which adds a positive bonus $b$ to a set
  $K$ of three state-action transitions. To choose them, we sample a target policy such that its return under the ground-truth reward function (i.e., $J(\pi) = \phi(\pi)^\top w^\star$) lies in the bottom quartile of returns. $K$ determines the three
  transitions that this policy visits most, and so we set $b$ such that it makes the optimal policy with respect to the proxy reward function fall in the bottom
  quartile of return under the ground-truth reward function. In other words, the proxy reward functions are constructed so they induce poor performance under the ground-truth reward function. The resulting three proxy reward functions induce optimal policies that lie at the 19th, 12th
  and 17th percentile of true return, respectively. 

  \paragraph{Reference policy}
  The reference policy
  $\pi_{\mathrm{ref}}$ is selected to be a policy at the 40th percentile of true return.

  \subsection{Algorithm and protocol}
  \label{app:regret-check:alg}

  We run PBRR (Algorithm~1), which we more specifically detail below for this empirical evaluation. We update the proxy reward function by learning a linear correction term, $\hat w_t = w_{\mathrm{proxy}} + \Delta_t$.
  At each iteration $t = 1, \dots, T$:
  \begin{enumerate}
    \item \textbf{Roll out policy.} $\pi^1_t = \arg\max_{\pi \in \Pi} \phi(\pi)^\top \hat w_t$
          (exact enumeration over $\Pi$), and $\pi^2_t = \pi_{\mathrm{ref}}$.
    \item \textbf{Compute confidence set.} Let
          $V_t = \sum_{l<t} x_l x_l^\top + \kappa\lambda I$, where
          $x_l = \phi(\tau^1_l) - \phi(\tau^2_l)$, and let
          $f(\pi, \pi') = \|\phi(\pi) - \phi(\pi')\|_{V_t^{-1}}$. $\Pi_t$ is the set of
          policies not dominated at width $\gamma_t$ (Eq.~15) around the projected
          center of Eq.~9.
    \item \textbf{Check policy pair (Lines 6--10).} If $\pi^1_t \notin \Pi_t$, or
          $\pi_{\mathrm{ref}} \notin \Pi_t$, or
          $C_1 f(\pi^1_t, \pi_{\mathrm{ref}}) \le \max_{\pi,\pi' \in \Pi_t} f(\pi,\pi')$,
          the pair is replaced by the maximally uncertain pair in $\Pi_t$.
    \item \textbf{Update proxy reward function} Sample one trajectory from each policy, then
          draw a label $o_t \sim \mathrm{Bernoulli}\big(\sigma(x_t^\top w^\star)\big)$.
          Refit $\Delta_t$ by minimizing the Bradley--Terry cross-entropy of
          $\sigma\big(x^\top (w_{\mathrm{proxy}} + \Delta)\big)$ via Newton's method.
  \end{enumerate}
  In our main empirical analysis, we set $C_1 = 0$. Here we evaluate Algorithm \ref{alg:exploration} with $C_1 > 0$. All constants follow Theorem~\ref{thm:known_trans_model_regret_bound}: $\kappa = 1/\sigma'(BW) \approx 405$, the
  smallest admissible regulariser $\lambda = B/\kappa$, and $\delta = 0.05$.

  \paragraph{Confidence width.} We set  $\gamma_t$ of Eq.~14, which determines the undominated policy set, to $10^{-3}$. If $\gamma_t$ is too large, no
  policy is ever eliminated within any feasible horizon: $\Pi_t = \Pi$ throughout. We chose this factor before evaluation on a held out set of random seeds, using coverage alone and never regret: it is the smallest width in a sweep for
  which $\pi^\star \in \Pi_t$ in at least 99\% of iterations.

  \paragraph{Evaluation.} We use $C_1 \in \{1, 2, 5\}$, $T = 50{,}000$ and 20 seeds per configuration and MDP. The dueling regret at $t$ is
  $R_t = \sum_{s \le t} J(\pi^\star) - \tfrac12\big(J(\pi^1_s) + J(\pi^2_s)\big)$. 95\% CIs are computed from 2000
  bootstrap resamples of seeds. We plot the following:
  \begin{itemize}
    \item \textbf{Empirical:} PBRR's actual cumulative regret $R_t$, averaged over 20 seeds, with a shaded 95\% band.
    \item \textbf{Proof bound:} the running sum of the per-step regret bound from the proof of Theorem~\ref{thm:known_trans_model_regret_bound},
          computed from each run's own confidence width and uncertainty; the empirical line
          staying below it shows the proof's main inequality held.
    \item \textbf{Trivial bound:} the regret incurred by always playing the
          worst policy, $t\,(J(\pi^\star) - \min_\pi J(\pi))$, shown for scale.
    \item \textbf{$\propto \sqrt{t}$:} a $\sqrt{t}$ curve pinned to meet the
          empirical line at $t = T$; when the solid line is
          flatter than it, regret is growing slower than $\sqrt{t}$, as the theorem allows.
  \end{itemize}

  \subsection{Results}
  \label{app:regret-check:results}

\begin{figure}[t]
  \centering
  \includegraphics[width=0.6\linewidth]{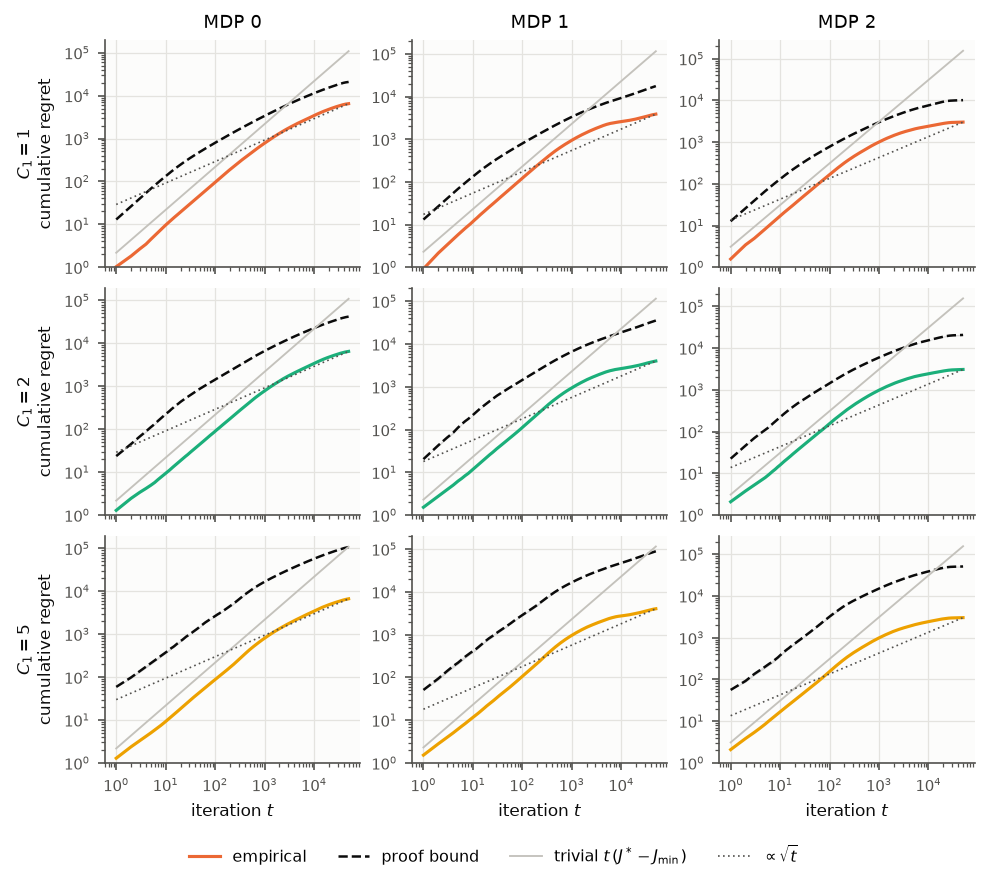}
  \caption{Cumulative dueling regret of PBRR on three random MDPs for
  $C_1 \in \{1, 2, 5\}$, $T = 50{,}000$ and across
  20 seeds; both axes are logarithmic.}
  \label{fig:regret-check}
  \end{figure}

  Figure~\ref{fig:regret-check} shows that, for every $C_1 > 0$ and on all three MDPs, cumulative regret grows more slowly than $\sqrt{t}$ over
  the second half of training as
  Theorem~\ref{thm:known_trans_model_regret_bound} predicts. The bound on the per-step regret at the core of the proof, $r_t \le 2C_1\gamma_t\,\|\phi(\pi^1_t) - \phi(\pi^2_t)\|_{V_t^{-1}}$, held at
  every iteration of every run, and its running sum is itself sublinear. Together, these results show that PBRR attains the $\tilde O(\sqrt{T})$
  regret rate of Theorem~\ref{thm:known_trans_model_regret_bound}, and does so through the mechanism used in its proof, with $\pi^\star$ remaining in the confidence set in all iterations.

  \paragraph{Limitation.} Because $\gamma_t$ is scaled down from the value in
  Theorem~4.1, coverage of $\pi^\star$ at this width is observed rather than guaranteed.
  The experiment therefore shows that the algorithm and the proof's per-step inequality
  behave as the analysis predicts. It does not verify the theorem's constants.

\end{document}